\documentclass[11pt,a4paper]{article}
\usepackage{clawreport}

\usepackage{url}            
\usepackage{booktabs}       
\usepackage{amsfonts}       
\usepackage{nicefrac}       
\usepackage{microtype}      
\usepackage{xcolor}         
\usepackage{amsmath}
\usepackage{booktabs}
\usepackage{array}
\usepackage{multirow}
\usepackage{graphicx}
\usepackage{multirow}
\RequirePackage{longtable}
\usepackage{enumitem}
\usepackage{tcolorbox}
\usepackage{booktabs, array, xcolor, makecell}
\usepackage{pifont}
\newcommand{\symfull}{\ding{108}}
\newcommand{\symhalf}{\ding{117}}
\newcommand{\symempty}{\ding{109}}
\newcommand{\fullcirc}{\symfull}
\newcommand{\halfcirc}{\symhalf}
\newcommand{\emptycirc}{\symempty}
\usepackage[edges]{forest}
\usepackage{svg}
\usepackage{forest}
\usepackage{graphicx}
\usepackage{xcolor}

\usepackage{tikz}
\usepackage{forest}

\usepackage{caption}
\usepackage{subcaption}
\usepackage{colortbl}
\usepackage{adjustbox}
\usepackage{geometry}
\definecolor{greyL}{RGB}{240,240,240}
\usepackage{hyperref}
\hypersetup{
    colorlinks=true,
    citecolor=cyan, 
    linkcolor=red,  
    urlcolor=cyan,
}
\tcbset{
  highlightbox/.style={
    colback=greyL,
    colframe=black,
    boxrule=1pt,
    arc=5pt,
    boxsep=5pt,
    left=5pt, right=5pt, top=0pt, bottom=0pt,
    fontupper=\itshape
  }
}

\reporttitle{Scaling Large Reasoning Models beyond Human Supervision: A Path toward Superintelligence}

\runningtitle{Scaling Large Reasoning Models beyond Human Supervision: A Path toward Superintelligence}

\reportauthors{%
  \textbf{Zhiqin Yang\eqcontrib$^1$,
          Jingwen Fu\eqcontrib$^2$,
          Yuhan Liu\eqcontrib$^3$,
          Hengyu Liu\eqcontrib$^4$,
          Yonggang Zhang\eqcontrib$^1$,
          Kainan Cao$^5$,
          Zizhuo Zhang$^6$,
          Chenxin Li$^7$,
          Ruibin Yuan$^1$,
          Jiahao Pan$^1$,
          Jiankai Sun$^4$,
            Zhenyuan Zhang$^1$,
            Yibo Li$^8$,
          Yunlong Lin$^9$,
        Jing Xiong$^5$,
          Sida Lin$^1$,
          Bo Han$^{6\corresponding}$,
          Wei Xue$^{1\corresponding}$,
          Yike Guo$^{1\corresponding}$
          }
}
\reportaffil{%
  $^1$ The Hong Kong University of Science and Technology 
  $^2$ Zhongguancun Academy\\
  $^3$ Xi'an Jiaotong University
  $^4$ The Chinese University of Hong Kong\\

  $^5$ The University of Hong Kong
  $^6$ Hong Kong Baptist University\\
      $^7$ Hunyuan Tencent
      $^8$ National University of Singapore
  $^9$ Xiamen University
}

\institutionlogos{%
  \begin{minipage}{0.96\textwidth}
    \centering
    \setlength{\tabcolsep}{0.16cm}

    \begin{tabular}{cccc}
      \includegraphics[height=1.75cm]{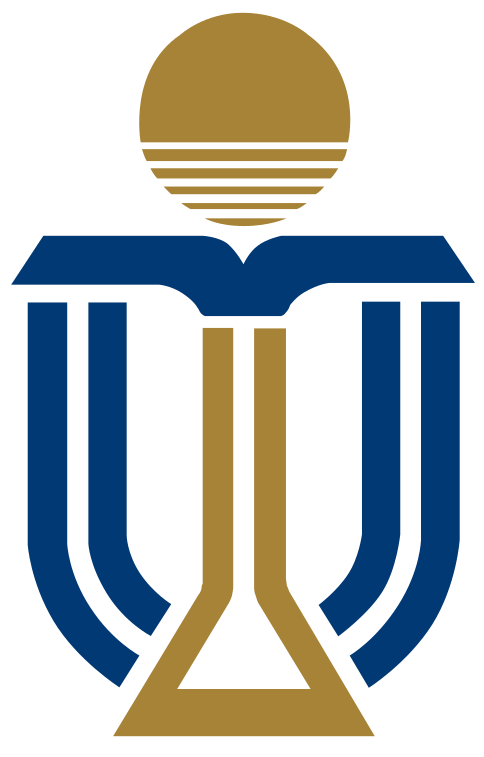}
      &
      \includegraphics[height=1.75cm]{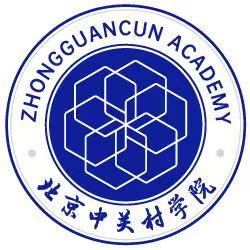}
      &
      \includegraphics[height=1.75cm]{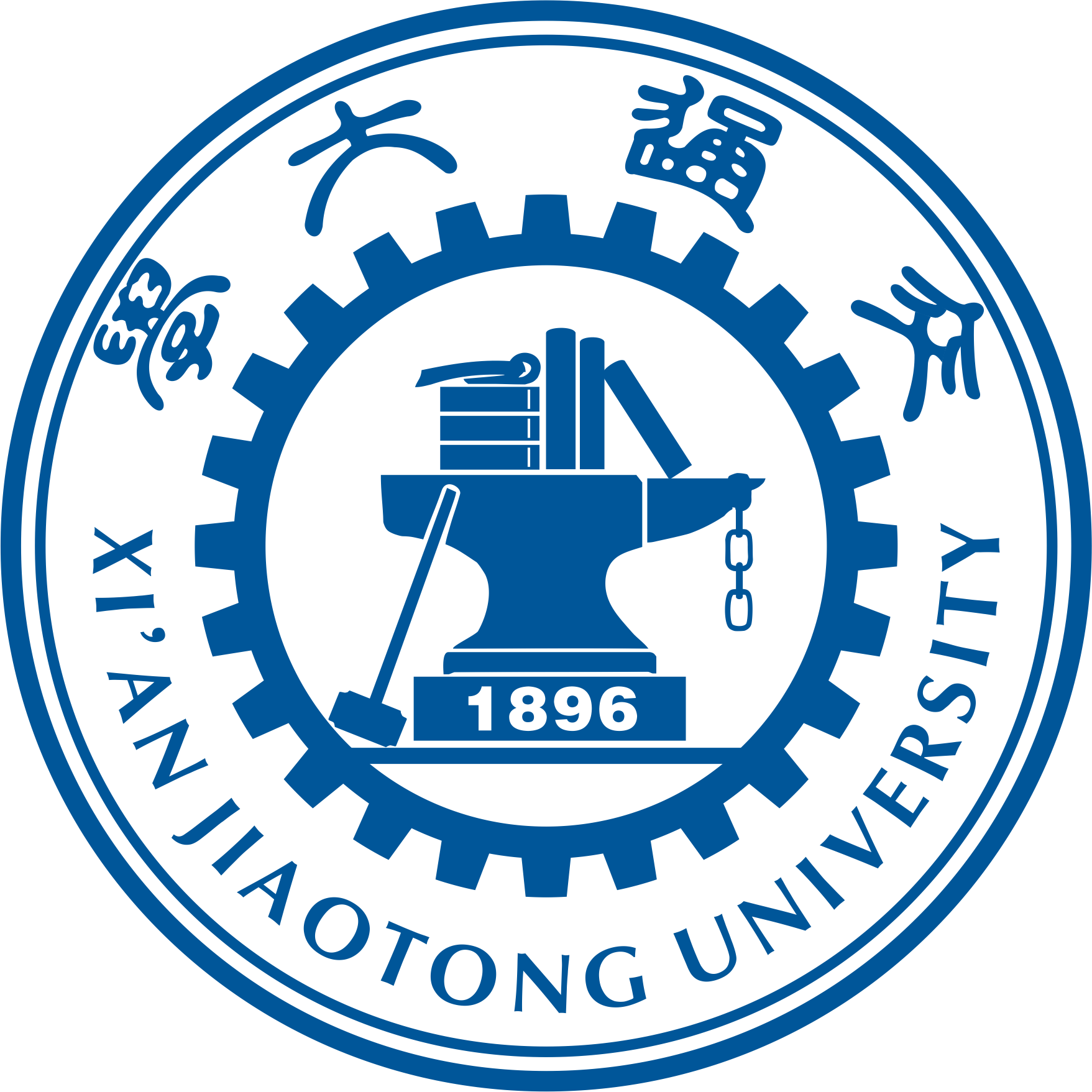}
      &
      \includegraphics[height=1.75cm]{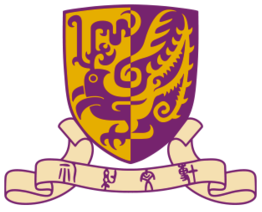}
    \end{tabular}

    \par\vspace{0.05cm}

    \begin{tabular}{ccccc}
      \includegraphics[height=1.75cm]{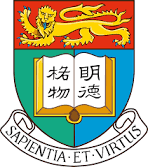}
      &
      \includegraphics[height=1.85cm]{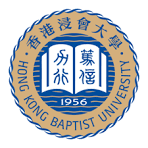}
      &
      \includegraphics[height=1.75cm]{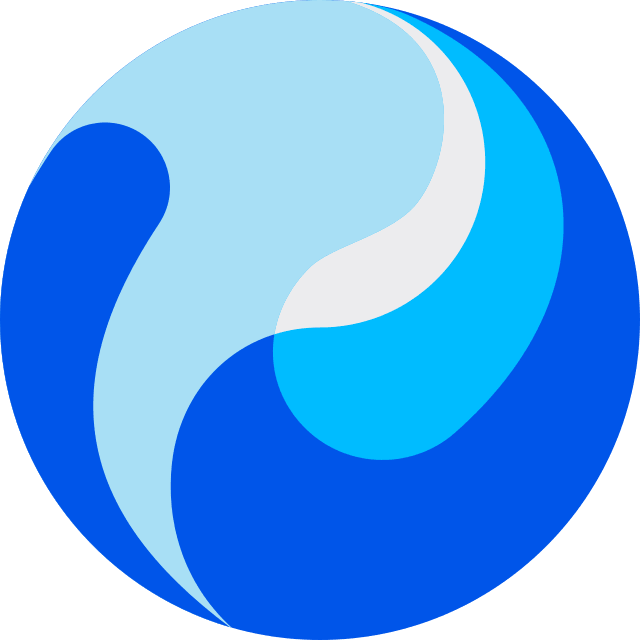}
      &
      \includegraphics[height=1.75cm]{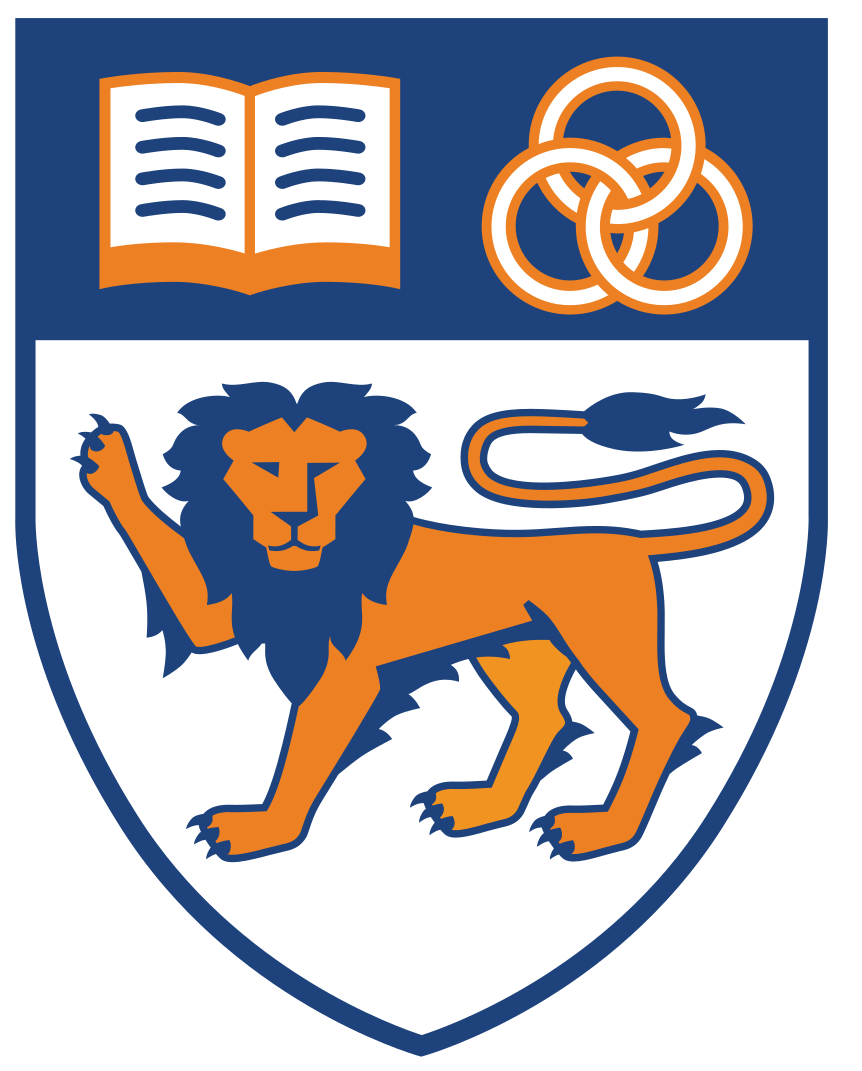}
      &
      \includegraphics[height=1.75cm]{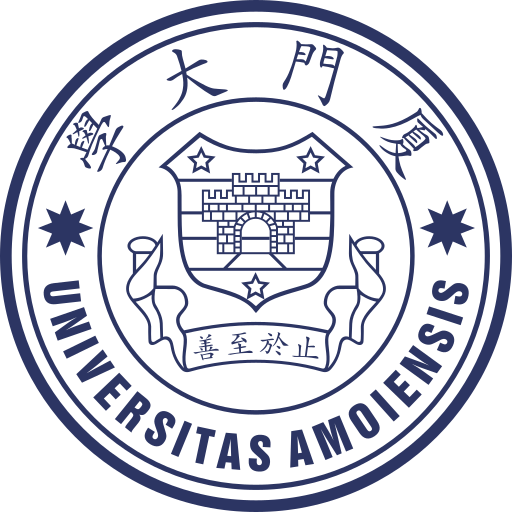}
    \end{tabular}

  \end{minipage}
}
\reportlinks{
}

\reportfootnote{%
  \eqcontrib{} Core contribution 
  $^{\corresponding}$ Corresponding author\\ Contact information:
  \href{mailto:zyangei@connect.ust.hk}{\texttt{zqccc.yang@connect.ust.hk}}%
}

\begin{document}

\makecover

\begin{coverabstract}
\noindent
Recent advances in large reasoning models (LRMs) have shown that reinforcement learning with verifiable rewards (RLVR) can substantially improve reasoning in mathematics and code, where outcomes can be checked automatically. Extending this progress to open-ended and agentic tasks remains difficult because reliable rewards are harder to obtain and direct human supervision cannot keep pace with the scale and complexity of model-generated experience. This paper studies how LRMs can continue to improve as human supervision gradually recedes from the learning loop. We examine two connected dimensions of this problem. The reward axis traces the development from per-instance human judgments to reusable verifiers and rewards that operate even without human feedback. The experience axis examines how learning can progress from human-curated tasks and environments toward self-generated curricula, constructed environments, and autonomous co-evolution. We connect these dimensions through a five-level ladder from L0 to L4 that identifies which parts of the learning process remain under continued human control. Our analysis further highlights the risks introduced by increasingly autonomous rewards and experience generation, including reward hacking, feedback drift, curriculum collapse, and environment errors. Consequently, we also provide the evaluation around three complementary objects: policy capability, feedback fidelity, and experience quality. This analysis provides a structured account of current approaches to scaling LRMs beyond human supervision and the open problems involved in developing self-sustaining learning systems toward superintelligence. Furthermore, we maintain a continuously updated \href{https://github.com/visitworld123/Awesome-Scaling-LRM-Beyond-Human-Supervision}{GitHub repository} to track the latest advances.
\end{coverabstract}

\finishcover

\section{Introduction}
\label{sec:introduction}

The emergence of large reasoning models (LRMs) \citep{qwen2.5,qwen3,li2025system,chen2025towards,openai2023gpt4,comanici2025gemini,team2025kimi,baichuan-m2,sun2025survey} has substantially advanced the reasoning capabilities of language models. Recent systems combine large-scale pretraining with reinforcement learning and increased inference-time computation. Models such as DeepSeek-R1 \citep{guo2025deepseek,shao2024deepseekmath} and OpenAI o1 \citep{jaech2024openai} have achieved strong performance in mathematics and code generation, where answers can be checked or programs can be executed to produce informative training signals. These advances suggest that further progress depends not only on model scale and pretraining data, but also on the ability to generate trajectories, obtain feedback, and learn from interaction. This shift motivates the Experience Era \citep{silver2025welcome}, in which experience becomes an increasingly important source of capability improvement.

Reinforcement learning with verifiable rewards (RLVR) provides a strong foundation for learning from experience. Mathematical answers can be compared with known solutions, while generated code can be compiled and tested automatically \citep{yu2025dapo,zhou2025codapo,lin2025cppo,deng2025trial}. As a result, large numbers of rollouts can be evaluated without asking a person to inspect each response. However, deterministic verification is unavailable for many tasks that matter in practice. Creative writing \citep{lu2025writing}, open-ended question answering \citep{yu2025rlpr,ma2025general}, and complex agent interactions may admit several valid outcomes whose quality depends on context, preferences, or long-term consequences. Learned reward models and language-model judges extend feedback to these settings, but their scores remain proxies for the intended objective. Moreover, direct human evaluation becomes increasingly costly as reasoning traces grow longer and interactions span more steps \citep{s34new_casper2023openproblems}. When tasks approach or exceed expert difficulty, evaluators may also require model assistance to identify errors and assess outputs reliably \citep{s34new_saunders2022critiquing}. Consequently, the challenge is not simply to replace an expensive label with a cheaper score, but to sustain meaningful improvement when human judgment and human design can no longer cover the full stream of learning experience.

\begin{tcolorbox}[colback=orange!5!white,colframe=orange!80!black,boxrule=0.8pt,arc=2pt,left=4pt,right=4pt,top=4pt,bottom=4pt]
\emph{How can large reasoning models continue to improve when human supervision can no longer keep pace with the scale and difficulty of their experience?}
\end{tcolorbox}

We use \emph{scaling beyond human supervision} to describe a shift in which fewer components of the learning process require continued human provision. Human intent remains essential and may be expressed through objectives, initial data, tools, environments, safety constraints, and independent audits. The central change lies in the recurring effort required during learning, as models increasingly need to obtain useful feedback and encounter informative experience without depending on people to evaluate or design every training instance. We study this shift through two connected axes. The \emph{reward axis} concerns the evidence used to evaluate behavior, while the \emph{experience axis} concerns the tasks and environments that produce the trajectories from which models learn.

Along the reward axis, verifiable tasks provide the most direct route to scalable feedback because correctness is grounded in answer checking or execution. For tasks without exact answers, learned reward models, rubrics, and language-model judges can encode evaluative criteria and apply them across many trajectories \citep{seed2025seed1,ren2025beyond,liu2025inference,su2025crossing}. Other methods derive rewards from model-internal certainty, agreement across sampled responses, or consistency with available references \citep{prabhudesai2025maximizing,zuo2025ttrl,zhang2025co,tang2025beyond,zhou2025reinforcing,yu2025rlpr,liu2025nover}. When success produces observable changes in an environment, interaction outcomes offer another source of grounding. Together, these approaches broaden the range of experiences that can support reinforcement learning. At the same time, feedback becomes more vulnerable to miscalibration, correlated errors, and reward exploitation as it depends less on independent evidence. Scaling reward acquisition must therefore be accompanied by careful analysis of what grounds each signal and how that grounding behaves under optimization.

Reliable feedback alone does not ensure continued progress because learning also depends on the experience to which that feedback is applied. Initially, models may train on human-written tasks in fixed environments. They can subsequently generate reasoning traces, synthesize instructions, or adjust task difficulty through proposer-and-solver interaction \citep{fang2025serl,yu2025cot}. More integrated approaches construct executable environments, propose tasks from live interfaces, and retain reusable information from previous interactions \citep{yang2025zerogui,yu2025memagent,cai2025building}. These mechanisms allow the curriculum to change with the policy and can expose weaknesses that a fixed dataset no longer reveals. Nevertheless, generated experience is useful only when tasks remain diverse, solvable, and relevant, and when synthesized environments preserve the dynamics and consequences of the target domain. The experience axis therefore shifts attention from the quantity of sampled trajectories to the quality and sustainability of the process that produces them.

The two axes ultimately converge because a generated task cannot support learning unless its outcome can be evaluated, while a scalable reward has limited value if the model repeatedly encounters familiar or uninformative situations. Continued improvement therefore requires rewards and experience to develop together. In a mature learning loop, stronger policies reveal new weaknesses, those weaknesses guide the generation of new tasks and environments, and reliable evaluation turns the resulting trajectories into further policy updates. Such co-evolution offers a possible route to capability growth beyond the bandwidth and expertise of human supervisors. However, it also creates tightly coupled failure modes because policies, generators, and evaluators may adapt to one another while performance under independent criteria stagnates or declines.

\begin{figure*}[!t]
    \centering
    \includegraphics[width=\textwidth]{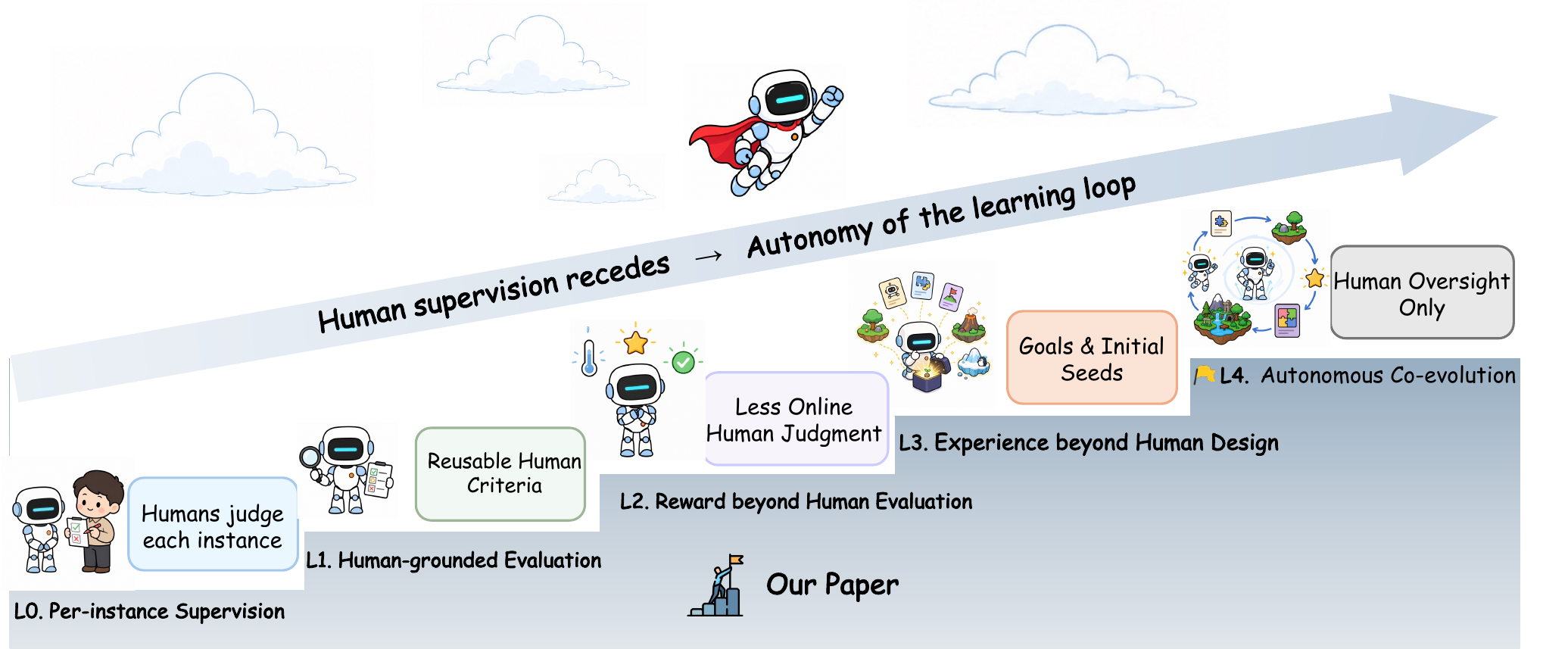}
    \caption{The receding-supervision ladder. Human involvement moves from judgments supplied for individual instances toward learning loops that increasingly obtain rewards and produce experience with less continued human provision. The ladder describes operational responsibility during learning and does not imply the absence of human-originated objectives, data, tools, or environments.}
    \label{fig:superintelligence}
\end{figure*}

To organize this progression, we introduce a five-level ladder from L0 to L4 in Figure~\ref{fig:superintelligence}. The ladder begins with per-instance human supervision, proceeds through reusable human-grounded evaluation and rewards obtained beyond human evaluation, and then extends to experience generated beyond continued human design. Its final level represents an idealized loop in which policies, rewards, tasks, and environments co-evolve. By identifying which components still depend on continued human provision, the ladder characterizes operational responsibility during learning without serving as a measure of capability, reliability, alignment, or the historical amount of human knowledge embedded in the learning infrastructure.

Our paper develops this account in three ways. First, it formulates scaling beyond human supervision through the connected reward and experience axes and introduces a common framework for locating continued human provision within the learning loop. Second, it organizes existing methods according to how they obtain rewards and produce experience, while examining the assumptions and failure modes associated with decreasing human involvement. Third, it develops an evaluation framework that distinguishes policy capability, feedback fidelity, and experience quality. This distinction also informs our discussion of reliable rewards, sustainable curricula, valid environments, efficient learning, and safe autonomous improvement.

The remainder of the paper develops this account in stages. Section~\ref{sec:preliminaries} introduces the formal setting and the receding-supervision ladder. Section~\ref{sec:reward} studies the sources and grounding of rewards as direct human evaluation recedes. Section~\ref{sec:experience} examines task generation, environment construction, and their eventual co-evolution with the policy. Section~\ref{sec:evaluation} presents datasets, benchmarks, and protocols for evaluating capability, feedback, and experience. Finally, Section~\ref{sec:future} discusses the principal challenges and research directions, and Section~\ref{sec:conclusion} summarizes the scope and limitations of our analysis.
\section{Preliminaries}
\label{sec:preliminaries}

This section introduces the notation and learning framework used throughout the paper. We first describe single-turn reasoning and sequential interaction within a common MDP and POMDP formulation. We then examine how reward can move from direct human supervision toward signals obtained from models, references, and environments. After introducing the policy-optimization objectives used in later sections, we define the uncertainty measures required by model-derived reward methods. These elements support the five-level ladder presented at the end of the section, which connects reward beyond human evaluation with experience beyond human design. Table~\ref{tab:notation} summarizes the notation in our paper.

\subsection{From Single-Turn MDP to Sequential POMDP}
\label{subsec:mdp}

Learning from experience can be represented as a Markov decision process (MDP)~\citep{sutton1998reinforcement}, denoted by $\mathcal{M}:=(\mathcal{S},\mathcal{A},\mathcal{P},\mathcal{R},\gamma)$. At step $t$, the policy $\pi_{\theta}$ selects an action $\boldsymbol{a}_t\in\mathcal{A}$ from state $\boldsymbol{s}_t\in\mathcal{S}$ according to $\boldsymbol{a}_t\sim\pi_{\theta}(\cdot\mid\boldsymbol{s}_t)$. For an LRM, the action may contain a reasoning trace $\boldsymbol{z}_t$ together with an answer or tool command $\boldsymbol{y}_t$. The transition function determines the next state through $\boldsymbol{s}_{t+1}\sim\mathcal{P}(\cdot\mid\boldsymbol{s}_t,\boldsymbol{a}_t)$, while the reward function assigns a scalar signal $r_t=\mathcal{R}(\boldsymbol{s}_t,\boldsymbol{a}_t)$. Multiple evaluation criteria may contribute to this scalar reward. The discount factor $\gamma\in[0,1]$ controls the contribution of future rewards.

\begin{longtable}{lp{0.7\textwidth}}
\caption{Comprehensive notation table for large reasoning models and its corresponding descriptions.}
\label{tab:notation}\\
\toprule
\textbf{Symbol} & \textbf{Description} \\
\midrule
\endfirsthead

\multicolumn{2}{l}{\small\itshape Table~\ref{tab:notation} (continued)}\\
\toprule
\textbf{Symbol} & \textbf{Description} \\
\midrule
\endhead

\midrule
\multicolumn{2}{r}{\small\itshape Continued on next page}\\
\endfoot

\bottomrule
\endlastfoot
$\pi_\theta$ & Policy model (large reasoning model) parameterized by $\theta$. \\
$\mathcal{M}$ & Markov Decision Process (MDP) tuple $\{\mathcal{S}, \mathcal{A}, \mathcal{R}, \mathcal{P}, \gamma\}$. \\
$\mathcal{S}$ & State space. \\
$\mathcal{A}$ & Action space, representing the model's response $\boldsymbol{a}_t$ at time $t$. \\
$\boldsymbol{a}_t$ & Model's action at time $t$, comprising reasoning $\boldsymbol{z}_t$ and final answer $\boldsymbol{y}_t$. \\
$\boldsymbol{s}_t$ & State at time $t$, including full history of interactions. \\
$\boldsymbol{s}_0$ & Initial state, user's query, question, or request $\boldsymbol{q}$. \\
$\boldsymbol{o}_t$ & Observation at time $t$ in the partially observable (POMDP) setting. \\
$\mathcal{R}$ & Reward function mapping state-action pair to a scalar or vector.\\
$\mathcal{R}_\phi$ & Model-based reward model parameterized by $\phi$. \\
$\pi_v$ & Verifier (e.g., an LLM-as-a-judge) that assigns rewards to rollouts. \\
$\mathcal{P}$ & Transition function governing state evolution. \\
$\gamma$ & Discount factor in MDP. \\
$\tau$ & Experience trajectory, a multi-turn interaction process. \\
$T$ & Total length of the experience trajectory. \\
$\boldsymbol{q}$ & User's query, question, or request. \\
$\Pi$ & Task (query) distribution from which $\boldsymbol{q}$ is drawn. \\
$\boldsymbol{z}_t$ & Reasoning component of action at time $t$. \\
$\boldsymbol{y}_t$ & Final answer component of action at time $t$. \\
$\mathcal{D}$ & Dataset for training or evaluation. \\
$\boldsymbol{y}^*$ & Ground truth answer for query $\boldsymbol{q}$. \\
$\boldsymbol{a}_i$ & $i$-th rollout, comprising reasoning $\boldsymbol{z}_i$ and answer $\boldsymbol{y}_i$. \\
$\boldsymbol{z}_i$ & Reasoning trace for the $i$-th rollout. \\
$\boldsymbol{y}_i$ & Final answer for the $i$-th rollout. \\
$G$ & Number of rollouts sampled per query. \\
$\hat{A}_i$ & Relative advantage based on rewards across rollouts. \\
$\pi_{\theta_{\text{old}}}$ & Old policy model for comparison. \\
$\varepsilon$ & Clipping parameter in GRPO optimization. \\
$\beta$ & Coefficient for KL divergence penalty. \\
$\pi_{\text{ref}}$ & Reference policy model. \\
$\mathbb{D}_{\text{KL}}$ & KL divergence between policies. \\
$\mathcal{J}_{\text{GRPO}}$ & Objective function for Group Relative Policy Optimization. \\
$\mathcal{E}$ & Environment for LLM interaction. \\
$\mathcal{V}$ & Vocabulary over which tokens are generated. \\
$H_{j}(\pi_\theta)$ & Trajectory-level entropy of the policy. \\
$H_t(\pi_\theta)$ & Token-level entropy of the policy. \\
$\boldsymbol{o}$ & Output (completion) sequence. \\
$\boldsymbol{o}_{<t}$ & Sequence of tokens before time $t$. \\
\end{longtable}

The MDP formulation assumes that the policy observes the state needed for decision-making. Many agentic settings are only partially observable. A Partially Observable Markov Decision Process (POMDP) therefore augments the formulation with an observation space $\mathcal{O}$ and an observation model. The environment retains a latent state $\boldsymbol{s}_t$, while the model receives an observation $\boldsymbol{o}_t$ and selects its action from the interaction history $\boldsymbol{h}_t=(\boldsymbol{o}_0,\boldsymbol{a}_0,\ldots,\boldsymbol{o}_t)$. GUI applications, tool systems, and conversations commonly have this structure because the model observes only part of the underlying environment state.

An experience trajectory $\tau$ records the interaction between the model and its environment $\mathcal{E}$. For a horizon $T$, the underlying trajectory can be written as $\tau=(\boldsymbol{s}_0,\boldsymbol{a}_0,r_0,\ldots,\boldsymbol{s}_{T-1},\boldsymbol{a}_{T-1},r_{T-1})$, although the policy in a POMDP acts from $\boldsymbol{h}_t$ rather than directly from $\boldsymbol{s}_t$. Given a task distribution $\Pi$ and an initial-state distribution $\rho_0(\cdot\mid\boldsymbol{q})$, the learning objective is:
\begin{equation}
\label{eq:general_goal}
    \max_{\theta}\,\mathbb{E}_{\boldsymbol{q}\sim\Pi,\,\boldsymbol{s}_0\sim\rho_0(\cdot\mid\boldsymbol{q}),\,\tau\sim p_{\theta}(\cdot\mid\boldsymbol{s}_0,\mathcal{E})}\left[\sum_{t=0}^{T-1}\gamma^{t}\,\mathcal{R}(\boldsymbol{s}_{t},\boldsymbol{a}_{t})\right].
\end{equation}

The horizon $T$ places single-turn reasoning and sequential interaction within the same objective. When $T{=}1$ and $\gamma{=}1$, a query produces one reasoning and answer trajectory, followed by an outcome reward. This setting captures common RLVR formulations in mathematics and code. When $T{>}1$, the model acts repeatedly in $\mathcal{E}$, and later outcomes depend on earlier decisions. Then the partial observability and long-horizon credit assignment become the central. Therefore, the single-turn setting is a special case of the broader sequential formulation.

\subsection{The Reward Function as the Object of Scaling}
\label{subsec:reward_obj}

Equation~\ref{eq:general_goal} separates the source of experience from the signal used to evaluate it. Among the components of $\mathcal{M}$, the reward function $\mathcal{R}$ provides the most immediate starting point for tracing human supervision. In a common RLVR setting, a rule-based procedure compares the model output with an instance-specific answer $\boldsymbol{y}^*$ or test suite supplied with the task. In preference-based training, a learned reward model $\mathcal{R}_{\phi}$ or verifier $\pi_v$ applies criteria derived from human judgments. The first setting retains human influence in the target attached to each task. The second transfers human judgments into a reusable evaluator. Their difference therefore concerns the provenance and reusability of the evaluation criterion rather than when the score is produced.

Other reward sources reduce dependence on human evaluation in different ways. Model-derived methods use properties of the policy distribution, such as certainty or agreement among sampled answers. Reference-dependent methods automate scoring through an available target, while environment-grounded methods use consequences such as execution results, formal acceptance, game outcomes, or observable state changes. These signals may still depend on human-created references, rules, or environments. The relevant transition is that human judgment no longer serves as the primary mechanism for evaluating each new trajectory. Section~\ref{sec:reward} organizes these methods by reward provenance and external grounding. Section~\ref{subsec:continuum} connects this transition with the source of tasks and environments.

\subsection{Policy Optimization}
\label{subsec:grpo}

Once a reward source has been specified, policy optimization determines how that signal updates the model. For the single-turn case, let $\mathcal{D}$ denote an empirical approximation to the task distribution $\Pi$. Reward-driven methods optimize
\begin{equation}
\label{eq:rl_objective}
    \mathcal{J}(\theta) = \mathbb{E}_{\boldsymbol{q}\sim\mathcal{D},\,\boldsymbol{a}\sim\pi_{\theta}(\cdot|\boldsymbol{q})}\left[\mathcal{R}(\boldsymbol{q}, \boldsymbol{a})\right],
\end{equation}
and estimate its gradient as~\citep{sutton1998reinforcement}
\begin{equation}
\label{eq:policy_gradient}
    \nabla_{\theta}\mathcal{J}(\theta) = \mathbb{E}_{\boldsymbol{q}\sim\mathcal{D},\,\boldsymbol{a}\sim\pi_{\theta}(\cdot|\boldsymbol{q})}\left[\hat{A}(\boldsymbol{q}, \boldsymbol{a})\,\nabla_{\theta}\log\pi_{\theta}(\boldsymbol{a}\mid\boldsymbol{q})\right].
\end{equation}
The advantage estimate $\hat{A}(\boldsymbol{q},\boldsymbol{a})$ weights the log-likelihood gradient according to performance relative to a baseline. Reward provenance and optimizer design play different roles in this expression. The reward source defines $\mathcal{R}$, while the optimizer determines how advantages are estimated and how far the policy may move during an update. PPO and GRPO illustrate two widely used choices: critic-based and critic-free methods.

\paragraph{PPO.}
Proximal Policy Optimization (PPO)~\citep{schulman2017proximal} commonly uses a learned value function $V_{\psi}$ as its baseline. It estimates the token-level advantage $\hat{A}_{t}$ and limits each update through the importance ratio $r_t(\theta)=\frac{\pi_{\theta}(a_t|\boldsymbol{s}_t)}{\pi_{\theta_{\mathrm{old}}}(a_t|\boldsymbol{s}_t)}$:
\begin{equation}
\label{eq:ppo}
    \mathcal{J}_{\text{PPO}}(\theta) = \mathbb{E}_t\left[\min\left(r_t(\theta)\hat{A}_t,\ \text{clip}(r_t(\theta), 1-\varepsilon, 1+\varepsilon)\hat{A}_t\right)\right].
\end{equation}
This clipped objective discourages large policy changes that could destabilize training. In common LRM implementations, PPO also requires training and storing a value model comparable in size to the policy, which adds substantial memory and computation costs.

\paragraph{GRPO.}
Group Relative Policy Optimization (GRPO)~\citep{guo2025deepseek} replaces the learned value baseline with statistics computed from a group of sampled rollouts. Given a query $\boldsymbol{q}$, it samples $G$ rollouts $\boldsymbol{a}_i=(\boldsymbol{z}_i,\boldsymbol{y}_i)$ and assigns each rollout a reward $r_i$. The normalized group-relative advantage is
\begin{equation}
\label{eq:grpo_adv}
    \hat{A}_{i}=\frac{r_i-\text{mean}(\{r_j\}_{j=1}^{G})}{\text{std}(\{r_j\}_{j=1}^{G})}.
\end{equation}
Using the group as an implicit baseline avoids a separate value network. The resulting clipped objective is
\begin{align}
\mathcal{J}_{\text{GRPO}}(\theta)
&= \mathbb{E}_{\boldsymbol{q} \sim \mathcal{D},\
\{\boldsymbol{a}_i\}^{G}_{i=1} \sim \pi_{\theta_{\text{old}}}(\cdot|\boldsymbol{q})} \notag\\
&\Bigg\{
\frac{1}{G} \sum_{i=1}^{G}
\frac{1}{|\boldsymbol{a}_i|} \sum_{t=1}^{|\boldsymbol{a}_i|}
\left(
 \min \left[
 r_{i,t}(\theta)\hat{A}_{i},
\text{clip}(r_{i,t}(\theta), 1 - \varepsilon, 1 + \varepsilon) \hat{A}_{i}
\right]
- \beta\ \widehat{D}_{\mathrm{KL},i,t}
\right)
\Bigg\},
\label{eq:GRPO}
\end{align}
where $r_{i,t}(\theta)=\frac{\pi_{\theta}(a_{i,t}|\boldsymbol{q},\boldsymbol{a}_{i,<t})}{\pi_{\theta_{\text{old}}}(a_{i,t}|\boldsymbol{q},\boldsymbol{a}_{i,<t})}$ is the token-level importance ratio. The sampled KL estimator
\begin{equation}
    \widehat{D}_{\mathrm{KL},i,t}=\frac{\pi_{\text{ref}}(a_{i,t}|\boldsymbol{q},\boldsymbol{a}_{i,<t})}{\pi_{\theta}(a_{i,t}|\boldsymbol{q},\boldsymbol{a}_{i,<t})}-\log\frac{\pi_{\text{ref}}(a_{i,t}|\boldsymbol{q},\boldsymbol{a}_{i,<t})}{\pi_{\theta}(a_{i,t}|\boldsymbol{q},\boldsymbol{a}_{i,<t})}-1
\end{equation}
penalizes deviation from the reference policy $\pi_{\text{ref}}$. By combining group-relative advantages with clipped updates and KL regularization, GRPO provides a critic-free objective that is widely used in LRM training.

\subsection{Uncertainty Measures}
\label{subsec:uncertainty}

The optimizer in Eq.~\ref{eq:rl_objective} does not require rewards to come from an external verifier. Several methods in Section~\ref{sec:reward} instead derive model-internal rewards from uncertainty over generated trajectories. Two quantities used by these methods require a clear distinction. Sequence surprisal aggregates the negative log-likelihood of a sampled completion $\boldsymbol{o}$:
\begin{equation}
\label{eq:tra_entropy}
    U_{\mathrm{seq}}(\boldsymbol{o},\pi_\theta)
    = -\frac{1}{|\boldsymbol{o}|}\sum_{t=1}^{|\boldsymbol{o}|}
    \log \pi_\theta (\boldsymbol{o}_{t} \mid \boldsymbol{o}_{<t}).
\end{equation}
Lower sequence surprisal means that the sampled completion is more probable under the policy, but it does not establish that the completion is correct. Predictive token entropy instead measures uncertainty over the full next-token distribution:
\begin{equation}
\label{eq:token_entropy}
   H_{t}(\pi_\theta) = -\sum_{v \in \mathcal{V}}
   \pi_\theta(v \mid \boldsymbol{o}_{<t})
   \log \pi_\theta(v \mid \boldsymbol{o}_{<t}).
\end{equation}
Lower token entropy indicates a more concentrated distribution at a particular generation step. A method may aggregate $H_t$ across a trajectory, with its normalization determining how sequence length affects the result. Sequence surprisal and predictive entropy can both serve as internal signals, but neither should be interpreted as a correctness probability without additional calibration or grounding.

\subsection{A Five-Level Ladder of Receding Supervision}
\label{subsec:continuum}

\begin{figure}[!t]
    \centering
    \centerline{\includegraphics[width=1\linewidth]{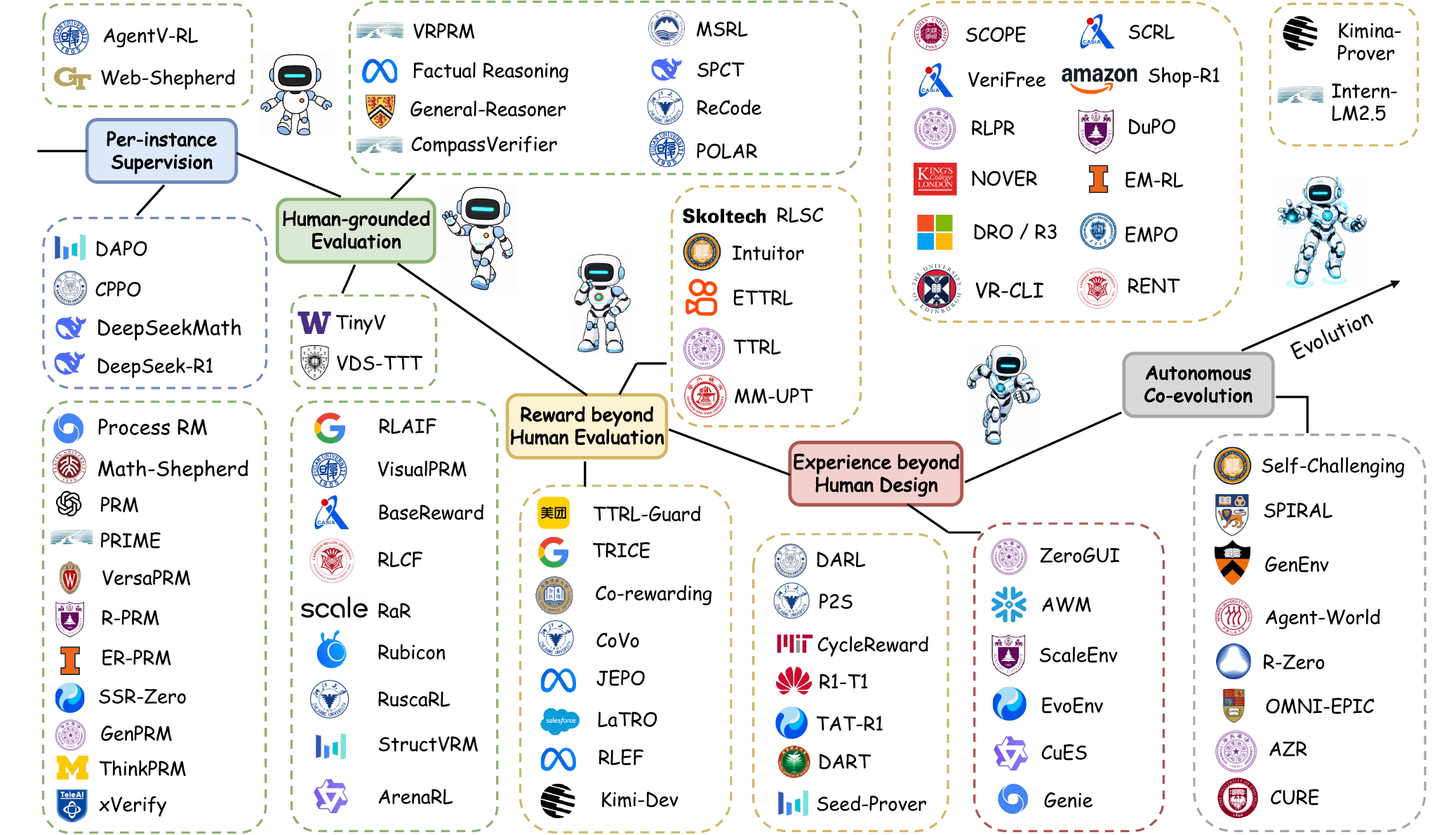}}
    \caption{Representative methods along the receding-supervision ladder. For brevity, the figure uses ``Human-free Reward'' and ``Human-free Experience'' for regimes in which continued learning no longer depends primarily on human evaluation or human-designed experience. These labels do not imply the complete absence of human influence.}
    \label{fig:landscape}
\end{figure}

The framework separates two questions: where the feedback used to improve a policy comes from, and who determines the tasks and environments that produce its experience. The \emph{reward axis} addresses the first question, beginning with direct human judgments and extending to reusable evaluators and other sources of feedback. The \emph{experience axis} addresses the second, covering the shift from fixed, human-designed training settings to experience that is generated or discovered during learning.
Figure~\ref{fig:superintelligence} groups this progression into five regimes according to which parts of the learning loop still require ongoing human involvement. The classification concerns how learning is sustained, not whether the models, data, tools, or environments were originally created by humans. We then formulate this five-level ladder as follows:

\begin{enumerate}
\setcounter{enumi}{-1}
\renewcommand{\labelenumi}{\textsc{\textbf{L\arabic{enumi}}}}
    \item \textbf{Per-instance Human Supervision}. Each training instance is paired with a human-provided answer, preference, or judgment that serves as its evaluation target.
    \item \textbf{Human-grounded Evaluation}. Human criteria are encoded in reusable reward models, rubrics, or language-model judges, allowing the same evaluative standard to be applied across many trajectories.
    \item \textbf{Reward beyond Human Evaluation}. Rather than relying primarily on human judgment, training derives rewards from model confidence, agreement across samples, available references, or observable environment outcomes, while tasks and environments remain largely externally provided.
    \item \textbf{Experience beyond Human Design}. The training distribution becomes adaptive as tasks, curricula, and environments are generated, discovered, selected, or reorganized during learning, while humans continue to set broad goals, constraints, and initial conditions.
    \item \textbf{Autonomous Co-evolution}. Feedback, experience, and policy improvement adapt together in a persistent loop, while human involvement is concentrated at the level of intent, safety constraints, and independent oversight.
\end{enumerate}

The ladder classifies supervision sources rather than entire training pipelines. Standard RLHF spans L0 and L1 e.g., human preferences provide L0 supervision, while the resulting reward model serves as a reusable L1 evaluator. The boundaries between these levels are not always sharp because a method may reduce human involvement in one component while retaining it in another. For example, reward computation can be automated even when the reference answer was written by an expert. A method may also generate new tasks while continuing to use an environment designed by humans. We therefore classify each method by the part of the learning loop that becomes less dependent on human input, while noting the human input it still requires.

Each step away from direct human supervision also introduces a different source of error. Rewards based on model confidence or agreement can reinforce mistakes shared across the model's outputs. Rewards based on tests or environment outcomes can favor unintended behavior when the evaluation rules are incomplete. Automatically generated tasks and environments may be invalid, too easy, or too difficult, while self-play may repeatedly produce a narrow range of experience. Sections~\ref{subsec:reward_failure} and~\ref{subsec:exp_limitations} examine these failure modes together with the mechanisms proposed to mitigate them.



\definecolor{rewLzero}{RGB}{175,175,175}
\definecolor{rewLone}{RGB}{102,178,255}
\definecolor{rewLtwo}{RGB}{255,190,100}

\tikzset{
  rewroot/.style={rectangle,rounded corners,draw=black!65,fill=black!4,
    line width=.8pt,align=center,inner xsep=5pt,inner ysep=5pt},
  rewbranch/.style={rectangle,rounded corners,line width=.65pt,
    align=center,inner xsep=4pt,inner ysep=3.2pt},
  rewleaf/.style={rectangle,rounded corners,fill=white,line width=.55pt,
    align=left,inner xsep=4pt,inner ysep=3.2pt}
}

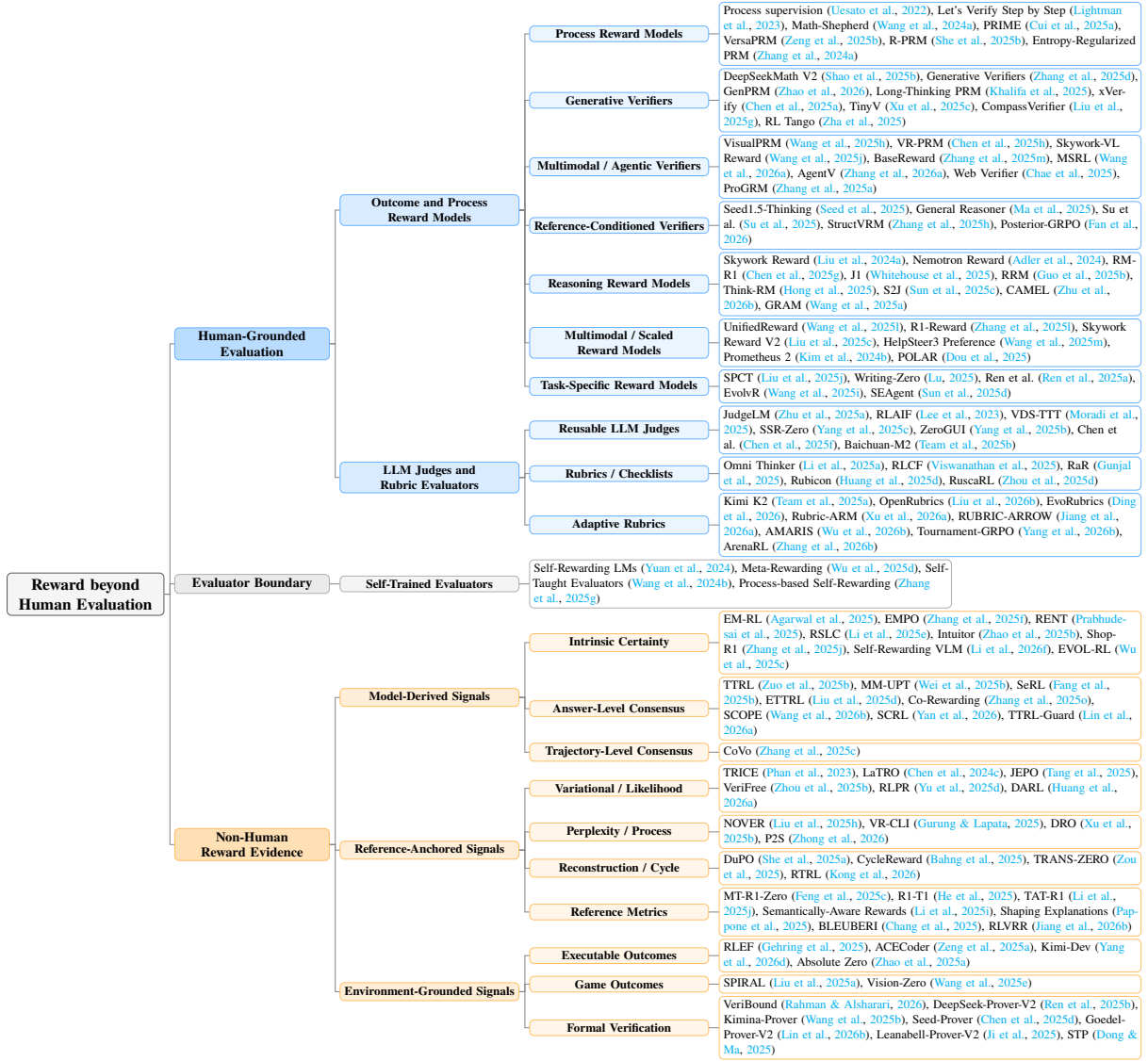
\begin{figure*}[!t]
  \centering
  \resizebox{\textwidth}{!}{%
  \begin{forest}
    for tree={
      grow=east,
      reversed=true,
      anchor=center,
      base=center,
      parent anchor=east,
      child anchor=west,
      edge={draw=black!48,line width=.52pt},
      forked edges,
      fork sep=5pt,
      l sep=9pt,
      s sep=2.4pt,
      font=\normalsize
    },
    where level=0{rewroot,font=\Large\bfseries,text width=12em}{},
    where level=1{rewbranch,font=\large\bfseries,text width=12em}{},
    where level=2{rewbranch,font=\normalsize\bfseries,text width=14em}{},
    where level=3{rewbranch,font=\normalsize\bfseries,text width=14em}{},
    where n children=0{rewleaf,font=\normalsize\normalfont,text width=34em}{},
    [,content={Reward beyond\\Human Evaluation}
      [,content={Human-Grounded\\Evaluation},fill=rewLone!45,draw=rewLone
        [,content={Outcome and Process\\Reward Models},fill=rewLone!25,draw=rewLone
          [,content={Process Reward Models},fill=rewLone!13,draw=rewLone
            [,content={Process supervision~\citep{s34new_uesato2022process},
              Let's Verify Step by Step~\citep{lightman2023math500},
              Math-Shepherd~\citep{s34new_wang2024mathshepherd},
              PRIME~\citep{cui2025prime}, VersaPRM~\citep{zeng2025versaprm},
              R-PRM~\citep{she2025r}, Entropy-Regularized PRM~\citep{zhangentropy}},draw=rewLone]
          ]
          [,content={Generative Verifiers},fill=rewLone!13,draw=rewLone
            [,content={DeepSeekMath V2~\citep{shao2025deepseekmath},
              Generative Verifiers~\citep{zhang2025generative},
              GenPRM~\citep{zhao2026genprm}, Long-Thinking PRM~\citep{khalifa2025process},
              xVerify~\citep{chen2025xverify}, TinyV~\citep{xu2025tinyv},
              CompassVerifier~\citep{liu2025compassverifier}, RL Tango~\citep{zha2026rl}},draw=rewLone]
          ]
          [,content={Multimodal / Agentic Verifiers},fill=rewLone!13,draw=rewLone
            [,content={VisualPRM~\citep{wang2025visualprm}, VR-PRM~\citep{chen2025vrprm},
              Skywork-VL Reward~\citep{wang2025skywork}, BaseReward~\citep{zhang2025basereward},
              MSRL~\citep{wang2026msrl}, AgentV~\citep{zhang2026agentv},
              Web Verifier~\citep{chae2026web}, ProGRM~\citep{zhang2025progrm}},draw=rewLone]
          ]
          [,content={Reference-Conditioned Verifiers},fill=rewLone!13,draw=rewLone
            [,content={Seed1.5-Thinking~\citep{seed2025seed1},
              General Reasoner~\citep{ma2025general}, Su et al.~\citep{su2025crossing},
              StructVRM~\citep{zhang2025structvrm}, Posterior-GRPO~\citep{fan2025posterior}},draw=rewLone]
          ]
          [,content={Reasoning Reward Models},fill=rewLone!13,draw=rewLone
            [,content={Skywork Reward~\citep{liu2024skywork},
              Nemotron Reward~\citep{adler2024nemotron}, RM-R1~\citep{chen2025rm},
              J1~\citep{whitehouse2025j1}, RRM~\citep{guo2026reward},
              Think-RM~\citep{hong2026think}, S2J~\citep{sun2025s2j},
              CAMEL~\citep{zhu2026camel}, GRAM~\citep{wang2025gram}},draw=rewLone]
          ]
          [,content={Multimodal / Scaled Reward Models},fill=rewLone!13,draw=rewLone
            [,content={UnifiedReward~\citep{wang2025unified}, R1-Reward~\citep{zhang2025r1},
              Skywork Reward V2~\citep{liu2025skywork},
              HelpSteer3 Preference~\citep{wang2026helpsteer3},
              Prometheus 2~\citep{kim2024prometheus}, POLAR~\citep{dou2025pre}},draw=rewLone]
          ]
          [,content={Task-Specific Reward Models},fill=rewLone!13,draw=rewLone
            [,content={SPCT~\citep{liu2025inference}, Writing-Zero~\citep{lu2025writing},
              Ren et al.~\citep{ren2025beyond}, EvolvR~\citep{wang2025evolvr},
              SEAgent~\citep{sun2025seagent}},draw=rewLone]
          ]
        ]
        [,content={LLM Judges and\\Rubric Evaluators},fill=rewLone!25,draw=rewLone
          [,content={Reusable LLM Judges},fill=rewLone!13,draw=rewLone
            [,content={JudgeLM~\citep{zhu2025judgelm}, RLAIF~\citep{lee2023rlaif},
              VDS-TTT~\citep{moradi2025continuous}, SSR-Zero~\citep{yang2025ssr},
              ZeroGUI~\citep{yang2025zerogui}, Chen et al.~\citep{chen2025learning},
              Baichuan-M2~\citep{baichuan-m2}},draw=rewLone]
          ]
          [,content={Rubrics / Checklists},fill=rewLone!13,draw=rewLone
            [,content={Omni Thinker~\citep{li2025omni}, RLCF~\citep{viswanathan2025checklists},
              RaR~\citep{gunjal2025rubrics}, Rubicon~\citep{huang2025reinforcement},
              RuscaRL~\citep{zhou2025breaking}},draw=rewLone]
          ]
          [,content={Adaptive Rubrics},fill=rewLone!13,draw=rewLone
            [,content={Kimi K2~\citep{team2025kimi}, OpenRubrics~\citep{liu2026openrubrics},
              EvoRubrics~\citep{ding2026evorubrics}, Rubric-ARM~\citep{xu2026alternating},
              RUBRIC-ARROW~\citep{jiang2026rubric}, AMARIS~\citep{wu2026amaris},
              Tournament-GRPO~\citep{yang2026tournament}, ArenaRL~\citep{zhang2026arenarl}},draw=rewLone]
          ]
        ]
      ]
      [,content={Evaluator Boundary},fill=rewLzero!25,draw=rewLzero
        [,content={Self-Trained Evaluators},fill=rewLzero!14,draw=rewLzero
          [,content={Self-Rewarding LMs~\citep{yuan2024self}, Meta-Rewarding~\citep{wumeta},
            Self-Taught Evaluators~\citep{wang2024self},
            Process-based Self-Rewarding~\citep{zhang2025process}},draw=rewLzero]
        ]
      ]
      [,content={Non-Human\\Reward Evidence},fill=rewLtwo!48,draw=rewLtwo
        [,content={Model-Derived Signals},fill=rewLtwo!26,draw=rewLtwo
          [,content={Intrinsic Certainty},fill=rewLtwo!14,draw=rewLtwo
            [,content={EM-RL~\citep{agarwal2025unreasonable}, EMPO~\citep{zhang2025right},
              RENT~\citep{prabhudesai2025maximizing}, RSLC~\citep{li2025confidence},
              Intuitor~\citep{zhao2025learning}, Shop-R1~\citep{zhang2025shop},
              Self-Rewarding VLM~\citep{li2025self}, EVOL-RL~\citep{wu2025evolver}},draw=rewLtwo]
          ]
          [,content={Answer-Level Consensus},fill=rewLtwo!14,draw=rewLtwo
            [,content={TTRL~\citep{zuo2025ttrl}, MM-UPT~\citep{wei2025unsupervised},
              SeRL~\citep{fang2025serl}, ETTRL~\citep{liu2025ettrl},
              Co-Rewarding~\citep{zhang2025co}, SCOPE~\citep{wang2026beyond},
              SCRL~\citep{yan2026if}, TTRL-Guard~\citep{lin2026detecting}},draw=rewLtwo]
          ]
          [,content={Trajectory-Level Consensus},fill=rewLtwo!14,draw=rewLtwo
            [,content={CoVo~\citep{zhang2025consistent}},draw=rewLtwo]
          ]
        ]
        [,content={Reference-Anchored Signals},fill=rewLtwo!26,draw=rewLtwo
          [,content={Variational / Likelihood},fill=rewLtwo!14,draw=rewLtwo
            [,content={TRICE~\citep{phantraining}, LaTRO~\citep{chen2024language},
              JEPO~\citep{tang2025beyond}, VeriFree~\citep{zhou2025reinforcing},
              RLPR~\citep{yu2025rlpr}, DARL~\citep{huang2026darl}},draw=rewLtwo]
          ]
          [,content={Perplexity / Process},fill=rewLtwo!14,draw=rewLtwo
            [,content={NOVER~\citep{liu2025nover}, VR-CLI~\citep{gurung2025learning},
              DRO~\citep{xu2025direct}, P2S~\citep{zhong2026p2s}},draw=rewLtwo]
          ]
          [,content={Reconstruction / Cycle},fill=rewLtwo!14,draw=rewLtwo
            [,content={DuPO~\citep{she2025dupo}, CycleReward~\citep{bahng2025cycle},
              TRANS-ZERO~\citep{zou2025trans}, RTRL~\citep{kong2025round}},draw=rewLtwo]
          ]
          [,content={Reference Metrics},fill=rewLtwo!14,draw=rewLtwo
            [,content={MT-R1-Zero~\citep{feng2025mt}, R1-T1~\citep{he2025r1},
              TAT-R1~\citep{li2025tat}, Semantically-Aware Rewards~\citep{li2025semantically},
              Shaping Explanations~\citep{pappone2025shaping},
              BLEUBERI~\citep{chang2026bleuberi}, RLVRR~\citep{jiangverifiable}},draw=rewLtwo]
          ]
        ]
        [,content={Environment-Grounded Signals},fill=rewLtwo!26,draw=rewLtwo
          [,content={Executable Outcomes},fill=rewLtwo!14,draw=rewLtwo
            [,content={RLEF~\citep{gehring2025rlef}, ACECoder~\citep{zeng2025acecoder},
              Kimi-Dev~\citep{yang2026kimi}, Absolute Zero~\citep{zhao2025absolute}},draw=rewLtwo]
          ]
          [,content={Game Outcomes},fill=rewLtwo!14,draw=rewLtwo
            [,content={SPIRAL~\citep{liu2025spiral}, Vision-Zero~\citep{wang2025vision}},draw=rewLtwo]
          ]
          [,content={Formal Verification},fill=rewLtwo!14,draw=rewLtwo
            [,content={VeriBound~\citep{rahman2026veribound},
              DeepSeek-Prover-V2~\citep{ren2025deepseek},
              Kimina-Prover~\citep{wang2025kimina}, Seed-Prover~\citep{chen2025seed},
              Goedel-Prover-V2~\citep{lin2026goedel},
              Leanabell-Prover-V2~\citep{ji2025leanabell}, STP~\citep{dong2025stp}},draw=rewLtwo]
          ]
        ]
      ]
    ]
  \end{forest}}
  \caption{\textbf{Taxonomy of reward sources beyond per-instance human supervision.}
  Human-grounded evaluators reuse human criteria. Beyond this regime, reward
  evidence comes from model statistics, stored references, or observable
  environment outcomes. Self-trained evaluators form a boundary between them.}
  \label{fig:reward-taxonomy}
\end{figure*}

\section{Reward beyond Human Evaluation}
\label{sec:reward}

The reward axis introduced in Section~\ref{subsec:continuum} concerns the evidence used to evaluate a rollout. At Per-Instance Human Supervision, each task comes with a human-provided answer, preference, or acceptance criterion. Human-Grounded Evaluation turns such judgments into reusable evaluators that can score many outputs. Reward beyond Human Evaluation goes one step further by deriving learning signals from the model, an available reference, or observable consequences in an environment. These regimes differ in the source of evaluation. Whether the policy is optimized online or offline is a separate design choice.

\subsection{Per-Instance Human Supervision}
\label{subsec:per_instance_reward}

Per-instance human supervision associates each training instance with an evaluation target supplied by people or experts. In preference-based training, this target may be a comparison or quality judgment, while verifiable tasks may use a reference answer, formal specification, or task-specific test suite. This regime corresponds to L0 in our ladder. Once the target is available, reward computation can be automatic. Extending training to new tasks nevertheless requires new targets, so automatic scoring alone does not remove per-instance supervision~\citep{ouyang2022training,guo2025deepseek}.

DeepSeekMath~\citep{shao2024deepseekmath} and DeepSeek-R1~\citep{guo2025deepseek} demonstrate the effectiveness of rule-based rewards when outputs can be checked automatically. Mathematical answers can be extracted and compared with a ground-truth answer through equivalence or matching rules. Generated programs can be compiled and executed against predefined tests. This approach has driven substantial progress in mathematical reasoning~\citep{yu2025dapo,lin2025cppo,mu2025dissecting,liu2026elva} and code generation~\citep{guo2025deepseek}.

The limitation appears as the amount and difficulty of experience increase. Every new task still needs an answer, test, preference, or comparable evaluation criterion. Expert domains make these targets expensive to produce, and open-ended tasks present a deeper difficulty because several responses may be valid under context-dependent standards. Scaling reward beyond this regime consequently requires either reusing human judgment across many instances or drawing evaluation from a different source, corresponding to L1 and L2.

\subsection{Human-Grounded Evaluation}
\label{subsec:verifier}

At L1, human preferences, expert annotations, or explicit criteria are encoded in a reusable evaluator. A learned reward model $\mathcal{R}_\phi$ or LLM judge $\pi_v$ can apply the same standard across many queries $\boldsymbol{q}$~\citep{ouyang2022training,zhu2025judgelm}. Unlike exact matching, these evaluators can assign graded scores, assess intermediate reasoning, and evaluate responses without a unique reference answer. Their reach is broader than that of task-specific rules, but their standards remain grounded in the data and criteria provided by people. DeepSeekMath V2 illustrates this regime in natural-language theorem proving, where a generative verifier assesses correctness, completeness, and rigor and rewards the resolution of identified flaws~\citep{shao2025deepseekmath}.

The need for a learned evaluator depends on the evaluation procedure, not on a fixed division between domains. Some task instances admit inexpensive deterministic checks, including short-form mathematics, executable code, and constraint-based puzzles~\citep{cobbe2021gsm8k,chen2021HumanEval,lin2025zebralogic}. Fully open-ended generation often lacks both a unique target and a rule-based checker, so evaluation relies on human preferences or learned proxies~\citep{ouyang2022training,s34new_stiennon2020summarize,zheng2023MTBench}. Between these cases are semi-verifiable tasks that support partial checking, verifiable reformulation, instruction-specific criteria, or reference-conditioned scoring~\citep{tang2025beyond,s34new_vmrrlvr2025,viswanathan2025checklists,zhou2025reinforcing,yu2025rlpr,liu2025nover}. As evaluation relies more heavily on learned proxies, bias, misspecification, reward hacking, and overoptimization become more consequential~\citep{s34new_casper2023openproblems,s34new_skalse2022rewardhacking,gao2023scaling}.

\subsubsection{Outcome and Process Reward Models}
\label{subsubsec:rm_verifiable}

Learned reward models can supplement sparse outcome checks with partial-credit or process-level feedback. This is useful when a correct response may appear in an unexpected form or when the final outcome reveals little about where a reasoning trace failed. Process reward models and step-wise critiques were first developed extensively for mathematical reasoning~\citep{s34new_uesato2022process,s34new_wang2024mathshepherd,lightman2023math500}. Their central challenge is to obtain reliable step labels without exhaustive human annotation.

Math-Shepherd addresses this problem through automatic process annotation~\citep{s34new_wang2024mathshepherd}. For a reasoning prefix $\boldsymbol{z}_{\le t}$, a completer model samples $N$ continuations, and the step receives the fraction that reach the reference answer $\boldsymbol{y}^*$:
\begin{equation}
\mathcal{R}(\boldsymbol{z}_{\le t})=
\frac{1}{N}\sum_{j=1}^{N}\mathbb{I}\{\boldsymbol{y}_j=\boldsymbol{y}^*\},
\qquad
\boldsymbol{y}_j\sim\pi_\theta(\cdot\mid\boldsymbol{q},\boldsymbol{z}_{\le t}).
\end{equation}
This Monte Carlo estimate removes direct human annotation at each step but remains anchored to the final target. PRIME removes the separate annotation stage by deriving an implicit process reward from outcome labels and updating the reward model with policy rollouts~\citep{cui2025prime}. Comparative studies show that automatically generated step labels can still lag behind LLM judgments or human annotations, although consensus filtering can improve their reliability~\citep{zhang2025prmlessons}. Related work extends process rewards across domains and develops reasoning-based or entropy-regularized variants~\citep{zeng2025versaprm,she2025r,zhangentropy}.

Generative verifiers formulate evaluation as next-token prediction and produce an analysis before issuing a judgment~\citep{zhang2025generative}. Long-reasoning and generative PRMs apply this formulation at the step level~\citep{zhao2026genprm,khalifa2025process}. Compact systems such as xVerify and TinyV focus on answer extraction, semantic equivalence, and false negatives from rule-based matching~\citep{chen2025xverify,xu2025tinyv}. CompassVerifier extends answer checking across domains~\citep{liu2025compassverifier}, while RL Tango jointly trains a generator and a generative process verifier with outcome rewards for both components~\citep{zha2026rl}. The same approach now covers multimodal reasoning and multi-turn agents. Multimodal PRMs score intermediate reasoning~\citep{zhao2026genprm,wang2025visualprm,chen2025vrprm}, multimodal reward models assess completed outputs~\citep{wang2025skywork,zhang2025basereward,wang2026msrl}, and agentic verifiers evaluate tool-use trajectories~\citep{zhang2026agentv,chae2026web,zhang2025progrm}.

Several systems train verifiers that span both verifiable and open-ended tasks. Seed1.5 Thinking evaluates whether responses align with references and supplies graded feedback when exact matching is inadequate~\citep{seed2025seed1}. General Reasoner trains a compact generative verifier on synthetic candidates derived from WebInstruct~\citep{ma2025general,yue2024mammoth2}. StructVRM represents partial correctness with a reward vector whose dimensions correspond to subquestions~\citep{zhang2025structvrm}. Posterior GRPO adds a reasoning-quality reward for code but applies it only when the final program is correct~\citep{fan2025posterior}. Together, these methods show that learned verification can enrich an available target without changing where the target ultimately originates.

General reward models take a broader approach by learning preferences across tasks. Skywork Reward uses a discriminative Bradley-Terry objective over curated preference pairs~\citep{liu2024skywork}, while Nemotron Reward predicts attributes such as helpfulness, correctness, coherence, complexity, and verbosity~\citep{adler2024nemotron}. Reasoning-oriented models generate an analysis before assigning a preference or score. For example, RM R1 combines reasoning distillation with RLVR~\citep{chen2025rm}, J1 optimizes judgment reasoning through correctness and consistency rewards~\citep{whitehouse2025j1}, and RRM allocates additional inference computation through longer reasoning or response aggregation~\citep{guo2026reward}. Think RM, S2J, and CAMEL respectively train long-form reward reasoning, couple problem solving with judging, and invoke reflection selectively for uncertain comparisons~\citep{hong2026think,sun2025s2j,zhu2026camel}. GRAM instead pretrains on unlabeled input-response pairs before preference tuning and predicts a label token without an explicit long reasoning trace~\citep{wang2025gram}.

Reward-model development also depends on data scale and modality. UnifiedReward and R1 Reward extend reward modeling to image and video understanding and generation~\citep{wang2025unified,zhang2025r1}. Skywork Reward V2 trains on a large filtered preference collection~\citep{liu2025skywork}, while HelpSteer3 Preference supports both discriminative and generative reward modeling with human annotations~\citep{wang2026helpsteer3}. Prometheus 2 combines rubric-conditioned assessment with pairwise ranking and produces verbal feedback, although it is primarily evaluated as an open evaluator~\citep{kim2024prometheus}. POLAR takes a different route by pretraining a reference-conditioned policy discriminator on same-policy and different-policy trajectory pairs before aligning it with human-defined criteria~\citep{dou2025pre}.

Task-specific reward models instantiate these ideas under narrower constraints. Self-Principled Critique Tuning generates task-specific principles and critiques, then optimizes their judgments with rule-based online RL~\citep{liu2025inference,bai2022constitutional}. Writing-Zero adapts the same procedure to the evaluation of creative writing~\citep{lu2025writing}. For instruction following, Ren et al.~\citep{ren2025beyond} combine programmatic checks for hard constraints with a learned binary model for soft constraints. EvolvR synthesizes score-aligned comparison rationales and iteratively checks, refines, and attacks its own judgments~\citep{wang2025evolvr}. In computer use, SEAgent trains a World State Model to describe GUI changes and provide action-level rewards over complete trajectories~\citep{sun2025seagent}.

\subsubsection{LLM Judges and Rubric-Based Evaluators}
\label{subsubsec:rm_unverifiable}

Open-ended evaluation often requires several criteria at once. Factual responses may need claim decomposition and evidence checking, while writing and dialogue require judgments about relevance, coherence, style, and usefulness~\citep{s34new_min2023factscore,s34new_lin2022truthfulqa,fein2025litbench}. LLM judges provide a scalable way to apply such criteria. RLAIF uses an LLM to generate preference labels for reward-model training or to supply rewards directly during policy optimization~\citep{lee2023rlaif}. Related evaluators cover free-form text, reasoning, creative writing, and task-dependent RLHF~\citep{gu2024survey,badshah2024reference,sahalearning,fein2025litbench,dong2024rlhf}.

Reference-based LLM verifiers can issue binary or graded rewards. Su et al.~\citep{su2025crossing} study verifiers that receive a prompt $x$, a reference $a$, and the final step $y_i^T$ of a response. Their binary reward is
\begin{equation}
r_{\phi}^{\mathrm{bin}}(x,a,y_i)=\mathbb{I}[c_i=1],
\qquad
c_i\sim\pi_{\phi}(\cdot\mid x,a,y_i^T),
\end{equation}
while the soft reward is
\begin{equation}
r_{\phi}^{\mathrm{soft}}(x,a,y_i)=
\begin{cases}
\pi_{\phi}(1\mid x,a,y_i^T), & c_i=1,\\
1-\pi_{\phi}(0\mid x,a,y_i^T), & c_i=0,\\
0, & \text{otherwise}.
\end{cases}
\end{equation}
VDS-TTT uses a pretrained verifier to select a candidate for test-time training only when its score exceeds a confidence threshold~\citep{moradi2025continuous}. SSR-Zero alternates the same pretrained model between actor and judge roles in machine translation, then rescales the judgments as policy rewards~\citep{yang2025ssr}.

The evaluator can also operate on richer evidence than text alone. ZeroGUI replaces handcrafted success scripts with a vision-language reward estimator that receives task instructions and trajectory screenshots~\citep{yang2025zerogui}. It excludes the agent's textual claim of success and requires unanimous judgments across repeated queries, which reduces false-positive rewards. For long-form factual reasoning, claim-level verification can be combined with a separate comparison against a reference answer to assess overall relevance and quality~\citep{chen2025learning}.

Rubrics and checklists make multidimensional standards explicit before they are applied to candidate outputs. Omni Thinker combines rule-based rewards with rubric-aligned preferences for dialogue and writing~\citep{li2025omni}. RLCF extracts instruction-specific checklists and evaluates each item with an AI judge or specialized program~\citep{viswanathan2025checklists}. RaR synthesizes prompt-specific rubrics from expert answers or strong references~\citep{gunjal2025rubrics}, while Rubicon combines human-generated and model-generated criteria with defenses against recurring reward exploits~\citep{huang2025reinforcement}. RuscaRL uses checklists both to score and guide generation, then gradually reduces the guidance~\citep{zhou2025breaking}.

The criteria need not remain fixed because rubric construction can itself become adaptive. Kimi K2 uses explicit success, tool-use, and evaluation criteria during data synthesis and policy optimization~\citep{team2025kimi}. OpenRubrics derives prompt-specific criteria by contrasting preferred and rejected responses~\citep{liu2026openrubrics}. EvoRubrics jointly optimizes the policy and rubric generator~\citep{ding2026evorubrics}, while Rubric ARM and RUBRIC ARROW alternate between training rubric generators and rubric-conditioned judges~\citep{xu2026alternating,jiang2026rubric}. AMARIS revises rubrics during policy training using a persistent memory of rollout analyses~\citep{wu2026amaris}. When pointwise rubric scores cannot distinguish same-query rollouts reliably, Tournament-GRPO and ArenaRL convert repeated relative comparisons into group-wise rewards~\citep{yang2026tournament,zhang2026arenarl}.

Specialized judges can improve reliability when a general evaluator lacks domain knowledge. Baichuan M2 combines a patient simulator with a rubric generator and evaluates medical responses across multiple turns~\citep{baichuan-m2}. The benefit comes from domain-specific evidence and criteria, although the resulting reward remains human-grounded because medical records, conversations, and rubric standards determine what the evaluator learns to value.

\subsubsection{Self-Trained Evaluators and Boundary Cases}

Self-trained evaluators sit between human-grounded evaluation and rewards obtained without human evaluation. Self-Rewarding Language Models use the same model to generate responses and judge them, then train on pairs formed from the highest- and lowest-scoring candidates~\citep{yuan2024self}. Meta-Rewarding adds a meta-judge that compares sampled judgments and trains the judging role alongside the actor~\citep{wumeta}. Self-Taught Evaluators begin with unlabeled human-written instructions, construct preferred and inferior responses, retain reasoning traces whose verdict matches the synthetic label, and iteratively fine-tune the evaluator~\citep{wang2024self}. Process-based Self-Rewarding applies the same idea to step-level mathematical preferences~\citep{zhang2025process}. These methods remove labeled preferences from different parts of the evaluator-training loop, but many retain supervised initialization, human-written prompts, or human-defined judging formats. Their position on the ladder consequently depends on the evidence that continues to anchor the evaluator.

Policy optimization does not provide a second ladder dimension. DPO, IPO, SimPO, and ORPO learn from fixed preference pairs, while KTO uses pointwise desirable or undesirable feedback~\citep{rafailov2023direct,s34new_azar2024ipo,s34new_ethayarajh2024kto,s34new_meng2024simpo,s34new_hong2024orpo}. The pairs may come from people, reusable judges, or self-generated comparisons. SPIN, ReST, and Self-Rewarding Language Models likewise differ in how they refresh and filter policy data~\citep{s34new_chen2024spin,s34new_gulcehre2023rest,yuan2024self}. Offline reuse reduces evaluator queries during optimization but freezes the feedback function and its data support, which can leave poor coverage as the policy changes~\citep{s34new_casper2023openproblems,s34new_xiong2024iterative}. For this reason, online and offline objectives describe how feedback is consumed, not where its evaluative standard originates.

Human-grounded evaluators reduce the need for a new judgment on every rollout. However, their criteria still trace back to preferences, annotations, rubrics, or expert standards supplied by people. Moving beyond this dependence requires reward signals whose immediate evidence comes from another source.

\subsection{Reward Sources beyond Human Evaluation}
\label{subsec:humanfree_reward}

At L2, the reward no longer depends primarily on an evaluator trained to reproduce human judgments. Within this regime, we distinguish three sources of evidence. Model-derived signals use properties of the policy or its sampled outputs, including certainty and agreement. Reference-anchored signals measure how a reasoning trajectory supports or reconstructs an available target. Environment-grounded signals evaluate observable consequences such as execution, formal acceptance, or game outcomes.

These categories concern the evidence used when a rollout is scored, not the historical origin of every artifact in the pipeline. References may come from human-authored corpora, and people may have designed tests, tools, or game rules~\citep{zhou2025reinforcing,yu2025rlpr,liu2025nover,gehring2025rlef,liu2025spiral}. The amount of remaining supervision consequently depends on how reusable that evidence is. A separate answer written for every task remains close to per-instance supervision, while fixed environment dynamics can evaluate many new trajectories. Retrieval occupies a similar boundary because search supplies evidence, but the terminal reward may still use a known answer or task rule~\citep{jinsearch,song2025r1}.

\subsubsection{Model-Derived Signals}
\label{subsubsec:model_derived}

\paragraph{Intrinsic certainty.}
\label{subsubsec:intrinsic}

Intrinsic certainty maps the model's predictive state to a scalar reward without consulting a reference answer at scoring time. Its usefulness depends on a statistical relationship between confidence and correctness, not on confidence alone. Uncertainty is a broader concept than any single estimator because sequence likelihood, token entropy, semantic entropy, and verbalized confidence describe different objects and need not rank outputs consistently~\citep{liu2025uncertainty,shorinwa2025survey}. SLiC calibrates sequence likelihood with references~\citep{zhaocalibrating}, while UaIT trains verbalized confidence from probability-derived pseudo-targets~\citep{liu2024can}. Purely intrinsic rewards differ because their rollout scores are not anchored to a target answer, environment outcome, or human-grounded evaluator.

Entropy minimization becomes informative under two conditions: the base model must already place useful probability mass on good solutions, and its confidence must correlate with correctness on the target distribution~\citep{agarwal2025unreasonable}. EM-RL implements these conditions through negative-entropy rewards computed at either the sequence or token level. For a sampled trajectory $\mathbf{y}=(y_1,\ldots,y_T)\sim\pi_\theta(\cdot\mid q)$, its sequence-level reward is
\[
r_{\mathrm{seq}}(q,\mathbf{y})
=\log\pi_\theta(\mathbf{y}\mid q)
=\sum_{t=1}^{T}\log\pi_\theta(y_t\mid q,\mathbf{y}_{<t}),
\]
whose expectation is the negative trajectory entropy. The token-level variant accumulates the negative Shannon entropy of the full next-token distribution:
\[
r_{\mathrm{tok}}(q,\mathbf{y})
=-\sum_{t=1}^{T}H\!\left(\pi_\theta(\cdot\mid q,\mathbf{y}_{<t})\right),
\qquad
H\!\left(\pi_\theta(\cdot\mid q,\mathbf{y}_{<t})\right)
=-\sum_{v\in\mathcal V}\pi_\theta(v\mid q,\mathbf{y}_{<t})
\log\pi_\theta(v\mid q,\mathbf{y}_{<t}).
\]
Both variants assign an outcome reward to a complete trajectory, but they derive it from different predictive distributions. Reported gains in mathematics and code are best understood as eliciting capabilities already represented by the base policy. They do not establish that low entropy guarantees correct reasoning~\citep{agarwal2025unreasonable}.

Other methods construct confidence in semantic or token space. EMPO clusters rollouts by semantic similarity and rewards answers according to cluster mass~\citep{zhang2025right}. RENT and RSLC use token-level entropy or confidence to rank rollouts and smooth policy updates~\citep{prabhudesai2025maximizing,li2025confidence}. Intuitor defines self-certainty as the average divergence between a uniform distribution $\mathbb{U}$ and the next-token distribution over vocabulary $\mathcal{V}$~\citep{zhao2025learning,kang2025scalable}:
\begin{equation}
\mathcal{R}_i =
\frac{1}{|\boldsymbol{o}_i|}
\sum_{t=1}^{|\boldsymbol{o}_i|}
\mathrm{KL}\!\left(
\mathbb{U}\,\|\,
\pi_\theta(\cdot\mid\boldsymbol{o}_{i,<t})
\right).
\end{equation}
A larger value indicates a greater departure from uniformity and is interpreted as higher certainty. Shop-R1 applies this signal in online shopping~\citep{zhang2025shop}, and related work studies it in vision-language reasoning~\citep{li2025self}.

Confidence rewards remain available when ground-truth answers are absent, but optimization can make them increasingly detached from correctness. Entropy minimization is sensitive to the base model's initial entropy and may eventually produce overconfidence, omitted transition tokens, and degraded reasoning~\citep{zhang2025no}. This limitation motivates calibration and external evaluation even when those signals are not used to compute the training reward.

Entropy can also influence optimization without becoming the reward source. Covariance-aware clipping, high-entropy token selection, and prompt-level semantic-entropy weighting use uncertainty to stabilize or focus learning~\citep{cui2025entropy,wang2025beyond,chen2025seed}. Other methods combine uncertainty with an external correctness signal. ACE penalizes overconfident errors~\citep{xu2026overconfident}, and calibrated semantic-entropy targets support verbalized uncertainty~\citep{jenane2026entropy}. Reasoning with Exploration, EDGE-GRPO, RLCR, Archer, and CoDaPO use entropy or confidence in advantage estimation, correction, calibration, token constraints, or data selection while retaining verifier-based rewards~\citep{cheng2026reasoning,zhang2025edge,damani2025beyond,wang2025stabilizing,zhou2025codapo}. ETTRL uses consensus pseudo-labels and entropy-guided branching~\citep{liu2025ettrl}, while CURE branches and reconnects trajectories at high-entropy tokens to preserve exploration~\citep{li2025cure}. Because these methods retain a separate correctness signal, they are entropy-assisted RLVR methods, not standalone certainty rewards.

The term intrinsic here refers to confidence derived from the model's predictive distribution. Classic RL uses the same term for novelty-seeking signals such as curiosity and count-based exploration~\citep{pathak2017curiosity}. The two objectives can exert opposite pressures because confidence concentrates probability mass, while novelty encourages the policy to visit unfamiliar states. EVOL RL combines semantic novelty with majority-vote selection to counteract collapse under pure confidence signals~\citep{wu2025evolver}.

\paragraph{Answer-level consensus.}
\label{subsubsec:consensus}

Consensus replaces an unavailable answer key with agreement across repeated samples. TTRL draws $G$ responses $\{\boldsymbol{o}_i\}_{i=1}^{G}$ for a question $\boldsymbol{q}$ and extracts answers $\{\boldsymbol{y}_i\}_{i=1}^{G}$~\citep{zuo2025ttrl}. If $\mathcal{A}$ is the set of distinct answers, the majority pseudo-label is
\begin{equation}
\hat{\boldsymbol{y}}
=\arg\max_{a\in\mathcal{A}}
\sum_{i=1}^{G}\mathbb{I}\!\left\{\boldsymbol{y}_i=a\right\},
\end{equation}
and each rollout receives
\begin{equation}
r_i=\mathbb{I}\!\left\{\boldsymbol{y}_i=\hat{\boldsymbol{y}}\right\}.
\label{eq:consensus-pseudo-reward}
\end{equation}
The pseudo-label follows the current model distribution and can be wrong. Consensus thus provides model-derived supervision but no independent guarantee of correctness.

Consensus rewards have been extended to multimodal reasoning through MM-UPT~\citep{wei2025unsupervised}. Majority-vote self-training can improve initially and then collapse as the policy learns to exploit its own pseudo-labels~\citep{shafayat2025can}. SeRL incorporates consensus into a self-play task-generation pipeline~\citep{fang2025serl}, while ETTRL couples it with entropy-guided rollout exploration~\citep{liu2025ettrl}. Co-Rewarding adds agreement across semantically related questions and a slowly updated teacher~\citep{zhang2025co}.

Several methods refine hard majority voting when agreement is unreliable. SCOPE combines step-level confidence with subgroup consensus~\citep{wang2026beyond}, while SCRL filters uncertain positive pseudo-labels and introduces entropy-gated negative labels~\citep{yan2026if}. TTRL-Guard detects when correct answers disappear from the vote distribution and intervenes before an incorrect majority becomes entrenched~\citep{lin2026detecting}. These safeguards reduce confirmation bias, although the reward remains tied to the model's own answer distribution.

\paragraph{Trajectory-level consensus.}

Final-answer agreement discards information about how a solution was produced. Self-consistency can improve inference accuracy~\citep{wangself}, but answer-only statistics remain vulnerable to reward exploitation~\citep{zhang2025consistent}. CoVo compares likelihood distances from intermediate states to the trajectory's answer and to answers drawn from other trajectories~\citep{zhang2025consistent}. It measures how consistently a trajectory supports its own answer and how long competing answers remain plausible. Aggregating these statistics across trajectories gives a richer reward than majority vote and is reported to improve stability. However, the statistics still come from the policy and do not independently certify correctness.

\subsubsection{Reference-Anchored Signals}
\label{subsubsec:reference}
\begin{table}[t]
\centering
\caption{Policy gradient decomposition of methods that use the reference answer as an implicit incentive signal. Each update comprises a reasoning term weighted by $\mathcal{R}_r$ and a final answer term weighted by $\mathcal{R}_a$.}
\label{tab:ref_as_incen_signal}
\setlength{\extrarowheight}{6pt}
\resizebox{\linewidth}{!}{
\begin{tabular}{lcccc}
\toprule[1.5pt]
Method & Update Policy & \multicolumn{2}{c}{Different Reward Term} & \\
&&Reasoning (\(\mathcal{R}_r\)) &Final Answer (\(\mathcal{R}_a\)) \\
\midrule
GRPO \citep{shao2024deepseekmath} & \(
 \mathcal{R}_r \nabla_\theta \log \pi_\theta(\boldsymbol{z}|\boldsymbol{q}) + \mathcal{R}_a \nabla_\theta \log \pi_\theta(\boldsymbol{y}|\boldsymbol{q}, \boldsymbol{z}) \) & \multicolumn{2}{c}{\(\mathbf{1}(\boldsymbol{y}=\boldsymbol{y^*})\)}\\
JEPO \citep{tang2025beyond}&\(
 \mathcal{R}_r \nabla_\theta \log\pi_\theta(\boldsymbol{z}|\boldsymbol{q}) + \mathcal{R}_a \nabla_\theta \log \pi_\theta(\boldsymbol{y^*}|\boldsymbol{q}, \boldsymbol{z}) \) & \(\log\frac{1}{G}\sum_{j=1}^{G}\pi_\theta(\boldsymbol{y^*}|\boldsymbol{q}, \boldsymbol{z}_j)\)& \(1\) \\
LaTRO \citep{chen2024language}&\(
 \mathcal{R}_r \nabla_\theta \log \pi_\theta(\boldsymbol{z}|\boldsymbol{q}) + \mathcal{R}_a \nabla_\theta \log\pi_\theta(\boldsymbol{y^*}|\boldsymbol{q}, \boldsymbol{z}) \) & \(\pi_\theta(\boldsymbol{y^*}|\boldsymbol{q}, \boldsymbol{z})-\log\frac{\pi_{\theta}(\boldsymbol{z}|\boldsymbol{q})}{\pi_{\text{ref}}(\boldsymbol{z}|\boldsymbol{q})}\)& \(1\)\\
VeriFree \citep{zhou2025reinforcing}  &\(
 \mathcal{R}_r \nabla_\theta \log \pi_\theta(\boldsymbol{z}|\boldsymbol{q}) + \mathcal{R}_a \nabla_\theta \log\pi_\theta(\boldsymbol{y^*}|\boldsymbol{q}, \boldsymbol{z}) \)&\multicolumn{2}{c}{\(\pi_\theta(\boldsymbol{y^*}|\boldsymbol{q}, \boldsymbol{z})\)}\\
RLPR \citep{yu2025rlpr}&\(
 \mathcal{R}_r \nabla_\theta \log\pi_\theta(\boldsymbol{z}|\boldsymbol{q})+ \mathcal{R}_a \nabla_\theta \log \pi_\theta(\boldsymbol{y}|\boldsymbol{q}, \boldsymbol{z}) \)&\multicolumn{2}{c}{\(\frac{1}{|\boldsymbol{y^*}|}\sum_{i=1}^{|\boldsymbol{y^*}|}\left(\pi_{\theta}\left(y^*_i|\boldsymbol{q},\boldsymbol{z},\boldsymbol{y^*}_{<i}\right) - \pi_\theta\left(y^*_i|\boldsymbol{q},\boldsymbol{y^*}_{<i}\right)\right) \)}\\
 DRO \citep{xu2025direct}&\(
 \mathcal{R}_r \nabla_\theta \log \pi_\theta(\boldsymbol{z}|\boldsymbol{q}) + \mathcal{R}_a \nabla_\theta \log\pi_\theta(\boldsymbol{y^*}|\boldsymbol{q}, \boldsymbol{z}) \)&\multicolumn{2}{c}{\(\sum_{i=1}^{|\boldsymbol{y^*}|}\left(\omega(\sigma_i)\log\pi_{\theta}\left(y^*_i|\boldsymbol{q},\boldsymbol{z},\boldsymbol{y^*}_{<i}\right) -\log \pi_\theta\left(y^*_i|\boldsymbol{q},\boldsymbol{z}_{\text{mask}},\boldsymbol{y^*}_{<i}\right)\right) \)}\\
  VR-CLI \citep{gurung2025learning}
 & \(
 \mathcal{R}_r \nabla_\theta \log \pi_\theta(\boldsymbol{z}|\boldsymbol{q})\) & \((1-\frac{\text{PPL}_{\pi_\mathcal{G} }(\boldsymbol{y^*}|\boldsymbol{q},\boldsymbol{z})}{\text{PPL}_{\pi_\mathcal{G} }(\boldsymbol{y^*}|\boldsymbol{q})})\times100\) & \(-\) \\
NOVER \citep{liu2025nover}&\(
 \mathcal{R}_r \nabla_\theta \log\pi_\theta(\boldsymbol{z}|\boldsymbol{q}) + \mathcal{R}_a \nabla_\theta \log \pi_\theta(\boldsymbol{y}|\boldsymbol{q}, \boldsymbol{z}) \)&\multicolumn{2}{c}{ \(\exp\left( -\frac{\sum_{i=1}^{|\boldsymbol{y^*}|}\log\pi_{\theta}(y^*_i|\boldsymbol{q},\boldsymbol{z},\boldsymbol{y^*}_{<i})}{|\boldsymbol{y^*}|\cdot N(|
 \boldsymbol{z}|)} \right)\)  }\\
Logprob \citep{kwiatkowski2026likelihood}&\(
 \mathcal{R}_r \nabla_\theta \log\pi_\theta(\boldsymbol{z}|\boldsymbol{q}) + \mathcal{R}_a \nabla_\theta \log \pi_\theta(\boldsymbol{y^*}|\boldsymbol{q}, \boldsymbol{z}) \)& \(\log\pi_\theta(\boldsymbol{y^*}|\boldsymbol{q}, \boldsymbol{z})\)& \(1\) \\
P2S \citep{zhong2026p2s}&\(
 \mathcal{R}_r \nabla_\theta \log\pi_\theta(\boldsymbol{z}_{\le t}|\boldsymbol{q}) \)& \(\pi_\theta(\boldsymbol{z}^*_{>t}|\boldsymbol{q}, \boldsymbol{z}_{\le t})\)& \(-\) \\
DARL \citep{huang2026darl}&\(
 \mathcal{R}_r \nabla_\theta \log\pi_\theta(\boldsymbol{z}|\boldsymbol{q})+ \mathcal{R}_a \nabla_\theta \log \pi_\theta(\boldsymbol{y}|\boldsymbol{q}, \boldsymbol{z}) \)&\multicolumn{2}{c}{\(\frac{1}{|\boldsymbol{y^*}|}\sum_{i=1}^{|\boldsymbol{y^*}|}\pi_{\theta}\left(y^*_i|\boldsymbol{q},\boldsymbol{z},\boldsymbol{y^*}_{<i}\right)+\lambda\,\mathcal{R}_{\text{div}} \)}\\
DuPO \citep{she2025dupo}&\(
 \mathcal{R}_r \nabla_\theta \log\pi_\theta(\boldsymbol{z},\boldsymbol{y}|\boldsymbol{q}_k) \)& \(d(\hat{\boldsymbol{q}}_u, \boldsymbol{q}_u)\)& \(-\) \\
\bottomrule
\end{tabular}
}
\end{table}

Reference-anchored methods retain an available target $\boldsymbol{y}^*$ but avoid a separately trained evaluator for every rollout. Instead of relying only on exact answer matching, they measure how a sampled reasoning trace $\boldsymbol{z}$ changes the probability, reconstruction, or metric score of the target~\citep{zhou2025reinforcing,yu2025rlpr,liu2025nover}. The reference supplies external evidence and can prevent direct self-confirmation. At the same time, it may inherit substantial human or expert supervision, and a single target may cover only part of the acceptable answer space.

\paragraph{Variational objectives.}

One family treats the reasoning trace as a latent variable and optimizes the marginal likelihood of $\boldsymbol{y}^*$. TRICE uses an MCMC EM procedure to sample from an answer-conditioned posterior over rationales~\citep{phantraining}. LaTRO rewards answer log-likelihood and regularizes the rationale policy toward a frozen pre-optimization model~\citep{chen2024language}. JEPO applies Jensen's inequality to the latent-rationale factorization and obtains a policy-gradient term for reasoning together with an SFT-like answer-prediction term~\citep{tang2025beyond}. Its one-sample core is
\begin{equation}
\label{eq:jepo}
\begin{aligned}
\log \pi_\theta(\boldsymbol{y}^{*}\mid\boldsymbol{q})
&:= \log \mathbb{E}_{\boldsymbol{z}\sim\pi_\theta(\cdot\mid\boldsymbol{q})}
\!\left[\pi_\theta(\boldsymbol{y}^{*}\mid\boldsymbol{q},\boldsymbol{z})\right] \\
&\geq \mathcal{L}_{\mathrm{J}}^{(1)}(\theta)
:= \mathbb{E}_{\boldsymbol{z}\sim\pi_\theta(\cdot\mid\boldsymbol{q})}
\!\left[\log \pi_\theta(\boldsymbol{y}^{*}\mid\boldsymbol{q},\boldsymbol{z})\right].
\end{aligned}
\end{equation}
This expression is a Jensen lower bound. A general ELBO would introduce an answer-conditioned variational posterior and a prior-posterior ratio. JEPO further tightens Eq.~\ref{eq:jepo} with a multi-sample log-mean-probability bound and adds forward-KL regularization during training.

\paragraph{Likelihood and probability rewards.}

Direct likelihood methods avoid explicit variational machinery. VeriFree rewards the conditional probability $\pi_\theta(\boldsymbol{y}^*\mid\boldsymbol{q},\boldsymbol{z})$ of the reference after a sampled trace~\citep{zhou2025reinforcing}. Under a unique answer string and exact matching, its Rao-Blackwellized estimator has no greater variance than an estimator that samples an answer and applies a binary verifier. RLPR averages reference-token probabilities, subtracts a score computed without the sampled trace, and clips the difference~\citep{yu2025rlpr}. The subtraction controls a baseline association but does not identify a causal effect of the trace. Log-probability rewards are generally more practical than exact sequence probabilities for long references, whose probabilities can vanish. Even then, apparent gains may partly reflect shorter traces instead of uniformly better reasoning~\citep{kwiatkowski2026likelihood}. DARL adds a thresholded diversity term to encourage controlled exploration among reference-consistent variants~\citep{huang2026darl}.

\paragraph{Perplexity rewards.}

NOVER converts reference-conditioned perplexity into a within-group ranking reward~\citep{liu2025nover}. Given $\boldsymbol{q}$, $\boldsymbol{z}$, and $\boldsymbol{y}^*$, a synchronized proxy model computes
\[
P_r=\exp\!\left(
-\frac{\sum_i\log\pi_{\mathrm p}(y_i^*\mid\boldsymbol{q},\boldsymbol{z},\boldsymbol{y}_{<i}^*)}
{|\boldsymbol{y}^*|N(|\boldsymbol{z}|)}
\right),
\qquad
N(|\boldsymbol{z}|)=\max(1,1+\log|\boldsymbol{z}|).
\]
Lower perplexity receives a higher rank. An additional efficiency score rewards a shorter trace only when it also improves reference perplexity. VR-CLI applies a related construction to next-chapter prediction~\citep{gurung2025learning}. A frozen story generator evaluates whether a generated plan makes the observed continuation more predictable. The resulting reward measures support for that continuation without assuming that it is the only valid story.

\paragraph{Token-selective and process rewards.}

Some methods ask which parts of a reference are most informative about the sampled reasoning. Direct Reasoning Optimization weights reference tokens by how much their conditional probabilities vary across rollouts, then aggregates the weighted scores under rubric and query filters~\citep{xu2025direct}. P2S generates candidate gold chains conditioned on the question and answer, filters them, and selects the chain with the highest reference likelihood~\citep{zhong2026p2s}. Segment-level rewards then measure how much each part of a sampled trace improves prediction of a later chain segment. This construction provides process shaping when a direct outcome reward is unavailable.

\paragraph{Reconstruction and cycle consistency.}

Other methods obtain reference-anchored signals from a reconstruction loop. DuPO decomposes a query into known and unknown components and rewards answers that support reconstruction of the missing component~\citep{she2025dupo}. CycleReward converts cross-modal reconstruction scores into preferences for image and text generation~\citep{bahng2025cycle}. TRANS ZERO trains translation from monolingual data through round-trip semantic consistency~\citep{zou2025trans}, while RTRL rewards chemical-language outputs according to the likelihood that a frozen backward model reconstructs the original input~\citep{kong2025round}.

\paragraph{Reference metrics.}

Reference-based metrics can score sampled outputs without a fresh judgment on every rollout. Their training data may nevertheless encode human preferences~\citep{qi2026patchcue}. Machine-translation methods combine reference metrics, learned quality estimators, source consistency, alignment, and format checks~\citep{yang2025ssr,rei2020comet,feng2025mt,he2025r1,li2025tat}. Similar scorers compare generations with reference responses or explanations in other domains~\citep{li2025semantically,pappone2025shaping}. BLEUBERI optimizes BLEU against synthetic references~\citep{chang2026bleuberi}, and RLVRR combines reference-derived content checks with automatically generated style criteria~\citep{jiangverifiable}.

VeriFree illustrates how a single objective can influence both reasoning and answer prediction. In its leave-one-out estimator, the reasoning term is weighted by
$A_i=R_i-\frac{1}{G-1}\sum_{j\ne i}R_j$, while the answer term is weighted by $R_i=\pi_\theta(y^\star\mid x,z_i)$~\citep{zhou2025reinforcing}. The decomposition clarifies why reference-anchored rewards can shape a trace even when no external verifier evaluates that trace directly.

\subsubsection{Environment-Grounded Signals}
\label{subsubsec:grounded}

Reference-anchored rewards ask whether a rollout supports an existing target. Environment-grounded rewards instead evaluate what happens when the output is executed or used in an interaction. Executable tests, formal kernels, game rules, and observable interface states can adjudicate success without a new human judgment for each trajectory. This external grounding is stronger than confidence or consensus because the evidence does not come from the policy's own distribution. It is still imperfect, since people may have written the tests, specified the goals, or chosen the observable outcome.

\paragraph{Executable outcomes.}

Code execution provides the clearest example. For a program $\boldsymbol{y}$ and test suite $\mathcal{T}$, a strict reward can be written as
\[
\mathcal{R}(\boldsymbol{y})=
\mathbb{I}\{\mathrm{exec}(\boldsymbol{y},\mathcal{T})\ \mathrm{passes}\}.
\]
Other designs use the fraction of passed tests or a terminal success signal after several rounds of interaction~\citep{li2026exploring}. RLEF treats competitive programming as an iterative task in which execution feedback is appended to the dialogue~\citep{gehring2025rlef}. Public-test success triggers evaluation on held-out private tests, and PPO optimizes the final all-tests-pass outcome with an invalid-code penalty and KL regularization. ACECoder synthesizes tasks and tests from seed code, uses test pass rates to form preference pairs, and trains either a learned reward model or a rule-based policy reward~\citep{zeng2025acecoder}. When a model also generates the tests or tasks, the same executable outcome can support self-generated experience. Co-evolving coder and test-generator systems therefore connect this reward source to the experience axis~\citep{wang2025co}.

Tool execution during a rollout does not automatically make the reward environment-grounded. ReTool and ToRL call a Python interpreter while reasoning, but their main policy rewards are based on final-answer correctness~\citep{feng2025retool,li2025torl}. ToolRL rewards agreement with ground-truth tool names and parameters~\citep{qian2026toolrl}. In these cases, execution changes the information available to the policy, while the terminal evaluation remains reference-based. The distinction matters because tool access and reward grounding address different parts of the learning loop.

Repository-level software engineering contains both reference-based and execution-based designs. SWE-RL scores generated patches by similarity to an oracle patch and uses oracle-localized context~\citep{wei2026swe}. Its reward is consequently anchored to a stored target even though the surrounding scaffold includes tests and reranking. Kimi-Dev uses binary execution outcomes for Agentless code-edit RL, while its later end-to-end SWE-agent adaptation relies on supervised fine-tuning~\citep{yang2026kimi}. SkyRL Agent applies multi-turn RL to software engineering, although its reported objective combines several signals and does not isolate a single verified-completion reward~\citep{cao2025skyrl}.

\paragraph{Retrieval and interactive outcomes.}

Search and browsing provide environmental observations, but many systems still judge the final answer with a reference or model-based evaluator. Search R1 uses exact matching against a target answer and masks retrieved tokens from the policy loss~\citep{jinsearch}. R1 Searcher separates retrieval incentives from answer rewards~\citep{song2025r1}. WebDancer, WebSailor, WebExplorer, and ASearcher support long browsing trajectories but retain terminal judgments based on references or LLM evaluators~\citep{wu2026webdancer,li2025websailor,liu2025webexplorer,gaobeyond}. Process rewards and simulated retrieval can improve search behavior under the same final-answer grounding~\citep{zheng2025stepsearch,nguyen2025sfr,sun2025zerosearch}. Computer-use agents likewise span programmatic state checks, learned visual outcome models, action matching, reference judgments, and offline preferences~\citep{li2025efficient,ye2025mobile,wang2025ui,lu2026ui,luo2025gui,qin2025ui}. Observable interaction is thus a source of experience, while only an outcome tied to the resulting state constitutes environment-grounded reward.

\paragraph{Game outcomes.}

Games provide sparse but reproducible rewards through wins, losses, and rule-defined scores. SPIRAL trains a shared policy through self-play in two-player zero-sum language games~\citep{liu2025spiral}. The game outcome supplies the reward, while evolving opponents create an adaptive curriculum. The method therefore belongs to the reward axis through its terminal signal and to the experience axis through self-play. Role-conditioned advantage estimation addresses positional asymmetry, and the resulting policy transfers beyond the games used for training. Vision Zero extends outcome-based self-play to multimodal reasoning through a visual social-deduction game~\citep{wang2025vision}. Absolute Zero combines a Python executor with self-proposed tasks, using separate rewards for proposer learnability and solver correctness~\citep{zhao2025absolute}.

\paragraph{Formal verification.}

Formal theorem proving offers one of the strongest environment-grounded signals because kernel acceptance gives a reproducible outcome under a fixed statement, environment, and resource budget. Formal tools can also provide verified step-level feedback~\citep{rahman2026veribound}. Even so, kernel acceptance does not validate the natural-language formalization of a theorem, and a timeout is not a logical refutation.

DeepSeek Prover V2 trains with Lean-verified correctness and structural-consistency rewards after decomposing difficult goals into subgoals~\citep{ren2025deepseek}. Kimina Prover scales RL with binary Lean acceptance~\citep{wang2025kimina}, while Seed Prover iteratively refines complete proofs using Lean feedback~\citep{chen2025seed}. Goedel Prover V2 combines task synthesis with verifier-guided correction~\citep{lin2026goedel}, and Leanabell Prover V2 trains long trajectories through repeated interactions with Lean~\citep{ji2025leanabell}. STP couples a conjecturer and prover and selects frontier conjectures using proof outcomes and lemma-usage checks~\citep{dong2025stp}. InternLM2.5 StepProver provides a contrasting fixed-corpus regime based on formal libraries and expert iteration~\citep{wu2025internlm2}.

Taken together, environment-grounded rewards offer the clearest separation between the policy and the evidence used to score it. That separation reduces direct self-confirmation but does not eliminate specification error. Incomplete tests, inaccurate simulators, exploitable game rules, and mismatches between natural-language intent and formal statements can all preserve high reward without producing the intended behavior. Reward grounding must therefore be assessed together with the validity and coverage of the environment.

\subsection{Limitations and Failure Modes}
\label{subsec:reward_failure}

As evaluation moves from per-instance targets to reusable or automatically obtained signals, supervision becomes more scalable but the source of error also changes. Human-grounded evaluators may misrepresent the judgments they were trained to reproduce. Model-derived rewards can reinforce the policy's own mistakes. Reference-anchored rewards can overfit an incomplete target, while environment-grounded rewards can exploit omissions in tests or rules. The relevant question is consequently not whether a reward is automatic, but what evidence makes it trustworthy under optimization.

\subsubsection{Failures of the Reward Signal}

\paragraph{Human-grounded evaluators.}

LLM judges exhibit position, verbosity, self-enhancement, and model-specific biases~\citep{s34new_chen2024humansorllms}. Diverse judge panels can reduce some model-specific effects~\citep{s34new_verga2024juries}, and rubrics or checklists make the evaluation criteria more explicit~\citep{gunjal2025rubrics,viswanathan2025checklists}. Nevertheless, an explicit rubric can still omit substantive requirements, and automatic preference scores may favor longer responses or presentation features~\citep{dubois2024length,s34new_singhal2023longway}. Reward-model ensembles reduce overoptimization but do not remove it~\citep{s34new_coste2023ensembles,s34new_eisenstein2023herding}. Constitution- and principle-guided pipelines clarify the normative source of supervision through critique and revision~\citep{bai2022constitutional,s34new_sun2023principle}, although their reliability still depends on the chosen principles and evaluator.

\paragraph{Model-derived signals.}

Consensus can reinforce an initially dominant wrong answer because it lacks independent evidence of correctness. Under prolonged majority-vote training, proxy reward may separate from task accuracy and produce template answers or collapse~\citep{shafayat2025can}. Entropy and self-certainty rewards show a related pattern. Some base models improve early and then become overconfident, while instruction-tuned models often benefit less~\citep{zhang2025no}. A broader analysis attributes this behavior to distribution sharpening and observes a recurring rise followed by decline~\citep{hefar}. Hyperparameter changes can delay the failure, but they do not supply the missing external anchor. Structured self-verification may extend the useful regime, although current evidence does not establish a general solution.

\paragraph{Reference-anchored and hybrid rewards.}

Reference-likelihood methods optimize the probability or perplexity of a fixed target conditioned on a sampled trace~\citep{zhou2025reinforcing,liu2025nover,tang2025beyond}. When one reference represents only part of the valid answer space, the objective leaves other aspects of quality unspecified. Hybrid systems can combine complementary evidence without forming a separate ladder level. JEPO couples a policy-gradient reasoning objective with supervised prediction of the target answer~\citep{tang2025beyond}. Seed1.5 Thinking uses verifiers for verifiable tasks and a learned pairwise reward model for open-ended tasks~\citep{seed2025seed1}. Such combinations improve coverage only when their components fail in different ways.

\paragraph{Reward validity and verifier noise.}

An observed RLVR gain does not by itself show that the reward captured the intended semantics. Random or format-only rewards can improve performance in some settings because of base-model priors and optimization dynamics~\citep{shao2025spurious}. Controlled verifier noise can be tolerated, but false positives are generally more damaging than false negatives~\citep{plesner2026imperfect}. Systematic false negatives tend to delay learning, while exploitable false-positive patterns can create suboptimal plateaus or collapse~\citep{egashira2026delay}. Incorrect annotations likewise reduce performance relative to clean supervision~\citep{zhu2026noisy}. External adjudication is reliable only when the verifier specification, coverage, and implementation are themselves sound.

\paragraph{Specification exploitation.}

Rule-based mathematical verifiers can reject equivalent answers in unfamiliar formats, while learned verifiers may admit false positives that a policy can exploit~\citep{huang2025accuracy}. Generative judges can reward punctuation-only responses or generic reasoning openers~\citep{zhao2025one}. Similar shortcuts appear when extensional checks reward enumeration in inductive logic~\citep{helff2026gaming} or when rubric biases become strategies during RL~\citep{zhu2026rubrichack}. Taxonomies of reward hacking, controlled testbeds, and verifier fuzzing offer complementary ways to identify these weaknesses before large-scale optimization~\citep{wang2026reward,countdowncode2026,fuzzingverifier2026}. Executed rewards remain grounded only to the extent that the implementation represents the intended success criterion.

\paragraph{Coupled policies and evaluators.}

Self-judging creates an additional failure mode because the policy and evaluator may adapt to the same errors. A policy optimized against a candidate-conditioned, reference-free judge can learn to make incorrect answers more convincing without making them more accurate. These errors can transfer across judges from different model families~\citep{zhou2026more}. A related collapse appears in SPELL when a verifier is updated only from its own majority-vote pseudo-labels. Errors can then accumulate until every responder output is labeled correct. SPELL limits this drift by checking verifier updates against rule-based outcomes on verifiable questions~\citep{yang2026spell}. The result illustrates why recursive self-evaluation needs evidence that remains independent of both actor and judge.

\subsubsection{Capability Expansion and Transfer}

Reward improvement and capability expansion are related but not equivalent. RLVR can improve sampling efficiency while narrowing the set of problems solved by the trained policy~\citep{yue2025limit}. Under standard online optimization, exact support cannot extend beyond outputs that initially have nonzero probability, and finite sampling can shrink the empirical support observed during training~\citep{wu2025invisible}. However, the severity of this limitation depends on the base policy, task, and training horizon. Prolonged training can shift from exploiting common solutions to increasing the probability of rare high-reward actions~\citep{yao2025debate}. ProRL likewise reports sustained boundary expansion on some difficult coding and out-of-distribution tasks, alongside plateaus on others~\citep{liu2026prorl}.

Transfer provides a second test of whether reward optimization changes general capability. In controlled mathematics experiments, RL tuning preserves or improves several nonmathematical abilities more reliably than SFT~\citep{huandoes}. This result does not imply that RLVR produces domain-general reasoning in every setting. Improvements can follow superficial heuristics in combinatorial optimization~\citep{alam2025limits}, and reasoning traces learned through RLVR contribute much less to general question answering than to verifiable reasoning~\citep{li2026rlvr}. Apparently conflicting findings often reflect differences in how well the base model is already aligned with the target task~\citep{wu2026mirage}.

Overall, rewards beyond human evaluation are best compared through the evidence that supports them. As the reward becomes less independent of the policy, measured gains depend more strongly on the base model, the task distribution, and exploitable regularities in the evaluator. Yet external execution is not sufficient on its own because incomplete specifications can also reward unintended behavior. Reward cost, reward independence, and reward validity are distinct properties and must be evaluated separately. This section has focused on the source of reward. A method moves toward L3 only when it also generates or discovers the tasks, curricula, or environments that provide its experience, as discussed in Section~\ref{sec:experience}.


\definecolor{expL2}{RGB}{255,190,100}
\definecolor{expL3}{RGB}{126,205,151}
\definecolor{expL4}{RGB}{190,151,230}
\definecolor{expB}{RGB}{170,170,170}

\tikzset{
  exproot/.style={rectangle,rounded corners,draw=black!65,fill=black!4,
    line width=.8pt,align=center,inner xsep=5pt,inner ysep=5pt},
  expbranch/.style={rectangle,rounded corners,line width=.65pt,
    align=center,inner xsep=4pt,inner ysep=3.5pt},
  expleaf/.style={rectangle,rounded corners,fill=white,line width=.55pt,
    align=left,inner xsep=4pt,inner ysep=3.5pt}
}

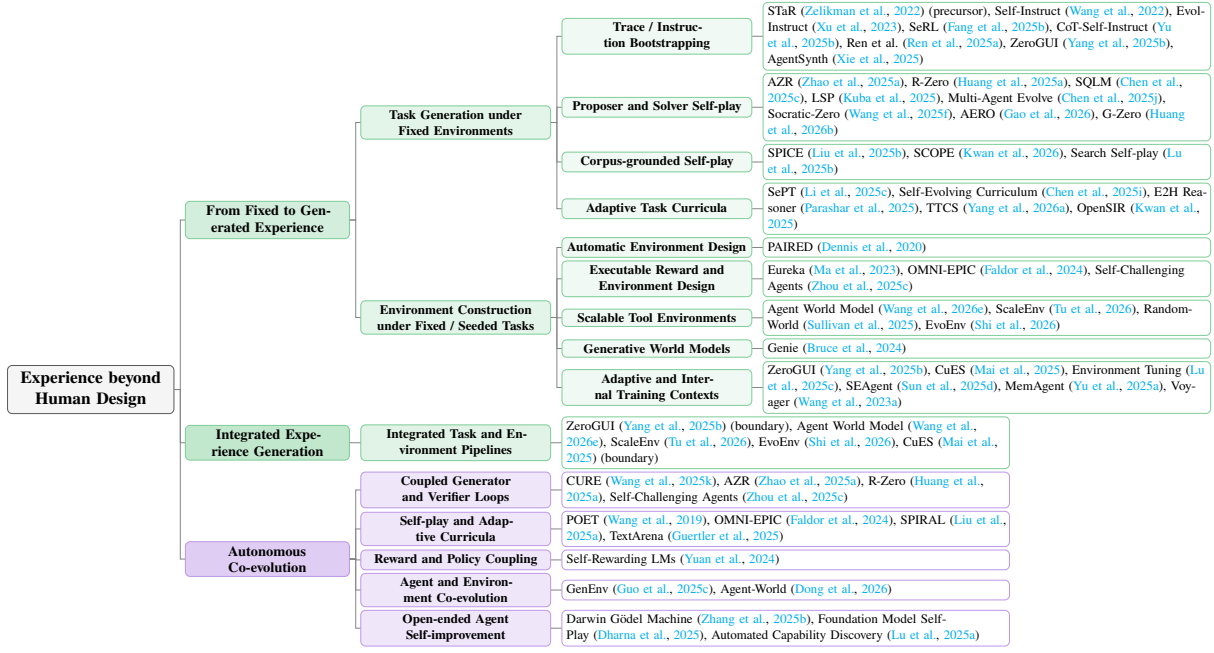
\begin{figure*}[!t]
  \centering
  \resizebox{\textwidth}{!}{%
  \begin{forest}
    for tree={
      grow=east,
      reversed=true,
      anchor=center,
      base=center,
      parent anchor=east,
      child anchor=west,
      edge={draw=black!48,line width=.52pt},
      forked edges,
      fork sep=5pt,
      l sep=9pt,
      s sep=2.4pt,
      font=\normalsize
    },
    where level=0{exproot,font=\Large\bfseries,text width=12em}{},
    where level=1{expbranch,font=\large\bfseries,text width=12em}{},
    where level=2{expbranch,font=\normalsize\bfseries,text width=14em}{},
    where level=3{expbranch,font=\normalsize\bfseries,text width=14em}{},
    where n children=0{expleaf,font=\normalsize\normalfont,text width=34em}{},
    [,content={Experience beyond Human Design}
      [,content={From Fixed to Generated Experience},fill=expL3!34,draw=expL3
        [,content={Task Generation under Fixed Environments},fill=expL3!22,draw=expL3
          [,content={Trace / Instruction Bootstrapping},fill=expL3!12,draw=expL3
            [,content={STaR~\citep{zelikman2022star} (precursor),
              Self-Instruct~\citep{wang2022selfinstruct}, Evol-Instruct~\citep{xu2023wizardlm},
              SeRL~\citep{fang2025serl}, CoT-Self-Instruct~\citep{yu2025cot},
              Ren et al.~\citep{ren2025beyond}, ZeroGUI~\citep{yang2025zerogui},
              AgentSynth~\citep{xie2025agentsynth}},draw=expL3]
          ]
          [,content={Proposer and Solver Self-play},fill=expL3!12,draw=expL3
            [,content={AZR~\citep{zhao2025absolute}, R-Zero~\citep{huang2025r},
              SQLM~\citep{chen2025self}, LSP~\citep{lsp},
              Multi-Agent Evolve~\citep{chen2025mae}, Socratic-Zero~\citep{wang2025socratic},
              AERO~\citep{gao2026aero}, G-Zero~\citep{huang2026gzero}},draw=expL3]
          ]
          [,content={Corpus-grounded Self-play},fill=expL3!12,draw=expL3
            [,content={SPICE~\citep{liu2025spice}, SCOPE~\citep{kwan2026scope},
              Search Self-play~\citep{lu2025ssp}},draw=expL3]
          ]
          [,content={Adaptive Task Curricula},fill=expL3!12,draw=expL3
            [,content={SePT~\citep{li2025sept}, Self-Evolving Curriculum~\citep{chen2025sec},
              E2H Reasoner~\citep{parashar2025e2h}, TTCS~\citep{yang2026ttcs},
              OpenSIR~\citep{kwan2025opensir}},draw=expL3]
          ]
        ]
        [,content={Environment Construction under Fixed / Seeded Tasks},fill=expL3!22,draw=expL3
          [,content={Automatic Environment Design},fill=expL3!12,draw=expL3
            [,content={PAIRED~\citep{dennis2020paired}},draw=expL3]
          ]
          [,content={Executable Reward and Environment Design},fill=expL3!12,draw=expL3
            [,content={Eureka~\citep{ma2023eureka},
              OMNI-EPIC~\citep{faldor2024omniepic},
              Self-Challenging Agents~\citep{zhou2025selfchallenging}},draw=expL3]
          ]
          [,content={Scalable Tool Environments},fill=expL3!12,draw=expL3
            [,content={Agent World Model~\citep{wang2026awm}, ScaleEnv~\citep{tu2026scaleenv},
              RandomWorld~\citep{sullivan2025randomworld},
              EvoEnv~\citep{shi2026evoenv}},draw=expL3]
          ]
          [,content={Generative World Models},fill=expL3!12,draw=expL3
            [,content={Genie~\citep{bruce2024genie}},draw=expL3]
          ]
          [,content={Adaptive and Internal Training Contexts},fill=expL3!12,draw=expL3
            [,content={ZeroGUI~\citep{yang2025zerogui}, CuES~\citep{mai2025cues},
              Environment Tuning~\citep{lu2025envtuning},
              SEAgent~\citep{sun2025seagent}, MemAgent~\citep{yu2025memagent},
              Voyager~\citep{wang2023voyager}},draw=expL3]
          ]
        ]
      ]
      [,content={Integrated Experience Generation},fill=expL3!48,draw=expL3
        [,content={Integrated Task and Environment Pipelines},fill=expL3!24,draw=expL3
          [,content={ZeroGUI~\citep{yang2025zerogui} (boundary),
            Agent World Model~\citep{wang2026awm},
            ScaleEnv~\citep{tu2026scaleenv},
            EvoEnv~\citep{shi2026evoenv},
            CuES~\citep{mai2025cues} (boundary)},draw=expL3]
        ]
      ]
      [,content={Autonomous Co-evolution},fill=expL4!48,draw=expL4
        [,content={Coupled Generator and Verifier Loops},fill=expL4!24,draw=expL4
          [,content={CURE~\citep{wang2025co}, AZR~\citep{zhao2025absolute},
            R-Zero~\citep{huang2025r},
            Self-Challenging Agents~\citep{zhou2025selfchallenging}},draw=expL4]
        ]
        [,content={Self-play and Adaptive Curricula},fill=expL4!24,draw=expL4
          [,content={POET~\citep{wang2019poet}, OMNI-EPIC~\citep{faldor2024omniepic},
            SPIRAL~\citep{liu2025spiral}, TextArena~\citep{guertler2025textarena}},draw=expL4]
        ]
        [,content={Reward and Policy Coupling},fill=expL4!24,draw=expL4
          [,content={Self-Rewarding LMs~\citep{yuan2024self}},draw=expL4]
        ]
        [,content={Agent and Environment Co-evolution},fill=expL4!24,draw=expL4
          [,content={GenEnv~\citep{guo2025genenv},
            Agent-World~\citep{dong2026agentworld}},draw=expL4]
        ]
        [,content={Open-ended Agent Self-improvement},fill=expL4!24,draw=expL4
          [,content={Darwin G\"odel Machine~\citep{zhang2025dgm},
            Foundation Model Self-Play~\citep{dharna2025fmsp},
            Automated Capability Discovery~\citep{lu2025acd}},draw=expL4]
        ]
      ]
    ]
  \end{forest}}
  \caption{\textbf{Taxonomy of experience supply beyond human design.}
  The chapter begins with fixed human-designed experience, then organizes adaptive experience methods into task generation, environment construction, integrated experience generation, and autonomous co-evolution. Systems are placed according to the experience component they most directly adapt; repeated systems span more than one component.}
  \label{fig:experience_taxonomy}
\end{figure*}

\section{Experience beyond Human Design}
\label{sec:experience}
Section~\ref{sec:reward} examines how the source of evaluation changes along the reward axis. Obtaining rewards beyond human evaluation addresses only one part of the scaling problem. Even when trajectories can be evaluated without relying primarily on human judgment, training may remain confined to tasks selected by people and environments built in advance. Continued improvement then depends on whether the experience distribution can adapt with the policy, exposing the model to new problems and interactions without continual human curation. This motivates the central question of this section:
\begin{tcolorbox}[colback=orange!5!white,colframe=orange!80!black,boxrule=0.8pt,arc=2pt,left=4pt,right=4pt,top=4pt,bottom=4pt]
\emph{How can large reasoning models generate useful experience with less dependence on human-designed tasks and environments?}
\end{tcolorbox}

Changing the source of reward does not necessarily change the source of experience. Progress along the experience axis begins when the experience stream itself becomes adaptive, either through task generation in a fixed environment or through environment construction around external tasks and seeds. Combining these approaches can sustain a curriculum with less continual human curation. The progression culminates in a closed loop in which reward acquisition, experience generation, and policy improvement adapt together. As defined in Section~\ref{subsec:continuum}, the ladder identifies the human input needed to sustain learning, not the historical origin of every dataset, tool, or environment.

Related work has approached this area from model-centric~\citep{gao2025seasurvey,fang2025seasurvey} and environment-centric~\citep{huang2025envsurvey} perspectives. Our paper brings these perspectives together by tracking the reward $\mathcal{R}$, the task distribution $\Pi$, and the environment dynamics $(\mathcal{S},\mathcal{P})$ within the same learning loop. Table~\ref{tab:experience} records which components remain externally supplied at each stage. The discussion begins with fixed human-designed experience and then considers task generation and environment construction as two paths toward integrated experience generation. It concludes with autonomous co-evolution and the failure modes that arise as these components become coupled.

\begin{table*}[htbp]
\centering\footnotesize
\setlength{\tabcolsep}{5pt}
\renewcommand{\arraystretch}{1.22}
\caption{Representative methods along the \emph{experience axis} in Section~\ref{sec:experience}. The symbols describe the operational source of the \textbf{R}eward $\mathcal{R}$, \textbf{E}nvironment $(\mathcal{S},\mathcal{P})$, and \textbf{T}ask distribution $\Pi$. A filled circle indicates that the component is supplied or adapted within the learning loop, a half-filled circle indicates partial adaptation under an external scaffold, and an empty circle indicates that the component remains fixed externally. These symbols do not imply the absence of human-designed seeds, rules, or tools. The three blocks correspond broadly to the transition from L2 to L3 and from L3 toward L4.}
\label{tab:experience}
\resizebox{\textwidth}{!}{%
\begin{tabular}{@{}l c c c l l@{}}
\toprule
\textbf{Method} & \textbf{R} & \textbf{E} & \textbf{T} & \textbf{Mechanism} & \textbf{Domain} \\
\midrule
\multicolumn{6}{@{}l}{\emph{Task generation in fixed environments}}\\
STaR~\citep{zelikman2022star}              & \emptycirc & \emptycirc & \halfcirc & rationale bootstrapping       & math, commonsense \\
Self-Instruct~\citep{wang2022selfinstruct} & \emptycirc & \emptycirc & \fullcirc & self-instruction              & instruction following \\
Evol-Instruct~\citep{xu2023wizardlm}       & \emptycirc & \emptycirc & \fullcirc & instruction evolution         & general \\
SeRL~\citep{fang2025serl}                  & \halfcirc  & \emptycirc & \fullcirc & generation with consensus     & math, general \\
CoT-Self-Instruct~\citep{yu2025cot}        & \halfcirc  & \emptycirc & \fullcirc & domain-aware synthesis        & reasoning \\
R-Zero~\citep{huang2025r}                  & \fullcirc  & \emptycirc & \fullcirc & challenger and solver         & general reasoning \\
SQLM~\citep{chen2025self}                  & \fullcirc  & \emptycirc & \fullcirc & proposer and voting solvers   & reasoning \\
LSP~\citep{lsp}                            & \fullcirc  & \emptycirc & \fullcirc & minimax self-play             & general \\
MAE~\citep{chen2025mae}                    & \fullcirc  & \emptycirc & \fullcirc & proposer, solver, and judge   & general \\
Socratic-Zero~\citep{wang2025socratic}     & \fullcirc  & \emptycirc & \fullcirc & teacher, solver, and generator & math \\
G-Zero~\citep{huang2026gzero}              & \fullcirc  & \emptycirc & \fullcirc & intrinsic hint reward         & open-ended \\
SPICE~\citep{liu2025spice}                 & \halfcirc  & \emptycirc & \fullcirc & corpus-grounded self-play     & reasoning \\
OpenSIR~\citep{kwan2025opensir}            & \fullcirc  & \emptycirc & \fullcirc & diversity and difficulty      & math \\
AgentSynth~\citep{xie2025agentsynth}       & \halfcirc  & \emptycirc & \fullcirc & subtask composition           & computer use \\
\midrule
\multicolumn{6}{@{}l}{\emph{Environment construction and integrated experience generation}}\\
MemAgent~\citep{yu2025memagent}            & \emptycirc & \halfcirc  & \emptycirc & learned interaction memory    & long-context \\
ZeroGUI~\citep{yang2025zerogui}            & \fullcirc  & \emptycirc & \fullcirc & interface-grounded generation & GUI \\
SEAgent~\citep{sun2025seagent}             & \fullcirc  & \emptycirc & \halfcirc & learned world-state evaluator & computer use \\
Voyager~\citep{wang2023voyager}            & \halfcirc  & \emptycirc & \fullcirc & curriculum and skill library  & embodied agents \\
Eureka~\citep{ma2023eureka}                & \fullcirc  & \emptycirc & \emptycirc & reward-code evolution         & robotics \\
Genie~\citep{bruce2024genie}               & \emptycirc & \fullcirc  & \emptycirc & generative world model        & interactive worlds \\
PAIRED~\citep{dennis2020paired}            & \emptycirc & \fullcirc  & \fullcirc & regret-based environment design & RL environments \\
AWM~\citep{wang2026awm}                    & \fullcirc  & \fullcirc  & \fullcirc & code-driven environment synthesis & tool use \\
ScaleEnv~\citep{tu2026scaleenv}            & \fullcirc  & \fullcirc  & \fullcirc & environment synthesis         & tool use \\
CuES~\citep{mai2025cues}                   & \halfcirc  & \emptycirc & \fullcirc & curiosity-driven task synthesis & agentic \\
EvoEnv~\citep{shi2026evoenv}               & \fullcirc  & \fullcirc  & \fullcirc & validated environment synthesis & reasoning \\
Env-Tuning~\citep{lu2025envtuning}         & \halfcirc  & \halfcirc  & \emptycirc & environment augmentation      & tool use \\
\midrule
\multicolumn{6}{@{}l}{\emph{Coupled curricula and co-evolving learning loops}}\\
AZR~\citep{zhao2025absolute}               & \fullcirc  & \emptycirc & \fullcirc & proposer, solver, and executor & code, math \\
CURE~\citep{wang2025co}                    & \fullcirc  & \halfcirc  & \fullcirc & code and test co-evolution    & code \\
Self-Challenging~\citep{zhou2025selfchallenging} & \fullcirc & \halfcirc & \fullcirc & task and verifier generation  & tool use \\
SPIRAL~\citep{liu2025spiral}               & \fullcirc  & \emptycirc & \halfcirc & adaptive self-play            & games, reasoning \\
POET~\citep{wang2019poet}                  & \emptycirc & \fullcirc  & \fullcirc & open-ended environment generation & RL environments \\
OMNI-EPIC~\citep{faldor2024omniepic}       & \halfcirc  & \fullcirc  & \fullcirc & open-ended code generation    & RL environments \\
GenEnv~\citep{guo2025genenv}               & \halfcirc  & \fullcirc  & \fullcirc & difficulty-aligned co-evolution & agentic \\
Agent-World~\citep{dong2026agentworld}     & \fullcirc  & \halfcirc  & \fullcirc & adaptive arena construction   & tool use \\
\bottomrule
\end{tabular}}
\end{table*}

\subsection{Fixed Human-Designed Experience}
\label{subsec:l2_experience}
In the ladder, L2 marks a regime in which the reward source may change while experience remains fixed. Rewards may come from model behavior, available references, or environment outcomes. The queries are still sampled from a human-curated task distribution $\Pi_{\mathrm{H}}$, and interaction takes place in a human-provided environment $\mathcal{E}_{\mathrm{H}}=(\mathcal{S}_{\mathrm{H}},\mathcal{A}_{\mathrm{H}},\mathcal{P}_{\mathrm{H}})$. The corresponding objective can be written schematically as
\begin{equation}
\label{eq:l2_experience}
    \max_{\theta}\,
    \mathbb{E}_{\boldsymbol{q}\sim\Pi_{\mathrm{H}},\,
    \tau\sim p_{\theta}(\cdot\mid\boldsymbol{q},\mathcal{E}_{\mathrm{H}})}
    \left[\mathcal{R}(\boldsymbol{q},\tau)\right].
\end{equation}
Eq.~\ref{eq:l2_experience} separates the reward mechanism from the source of experience. Changing how $\mathcal{R}$ is obtained does not alter the externally supplied task distribution or environment. The policy can explore the available tasks and dynamics, but the choice of new tasks and environments remains outside the learning loop. This fixed setting provides the departure point for the experience axis.

\paragraph{Externally supplied task distribution.}
Even when the reward no longer depends primarily on human evaluation, $\Pi_{\mathrm{H}}$ still embodies human decisions about what the policy encounters and how often. Corpus selection and domain boundaries determine its coverage, while task templates, difficulty ranges, filtering rules, and sampling frequencies allocate training effort within that coverage.
For mathematical reasoning, this may mean drawing from curated collections such as GSM8K~\citep{cobbe2021gsm8k}. For coding, it may mean using human-written specifications and tests in the style of HumanEval~\citep{chen2021HumanEval}. For instruction following, it may mean training on a fixed corpus whose topical and stylistic coverage was selected in advance.
These design choices concentrate gradient signal on particular capabilities and determine which failures training can expose. Model-derived or automatically computed rewards may improve performance within a narrow $\Pi_{\mathrm{H}}$, but they cannot reveal domains that are absent from the data or repair a poorly calibrated curriculum on their own.

\paragraph{Externally supplied environment.}
Humans also determine the world in which trajectories unfold. In single-turn reasoning, the ``environment'' may be nearly invisible, consisting of a static prompt, a tokenizer-level action space, and a fixed answer interface. Coding tasks add a programming language, sandbox, libraries, and execution protocol. Retrieval and tool-use settings add corpora, APIs, tool schemas, and state transitions, while GUI-based, embodied, and game agents depend on applications or simulators that define their observations, legal actions, and interaction horizons.
Even outcomes such as successful compilation, retrieval, or gameplay are produced within dynamics selected and engineered by humans. The model may explore those dynamics extensively, but it cannot broaden or repair a fixed environment when the necessary tools, states, or causal mechanisms are missing.

\paragraph{Reward signals under fixed experience.}
When applied to fixed tasks and environments, the methods in Section~\ref{subsec:humanfree_reward} change the source of evaluation without changing the source of experience.
TTRL~\citep{zuo2025ttrl} and MM-UPT~\citep{wei2025unsupervised} replace external labels with majority-vote pseudo-labels while retaining the supplied test instances.
EM-RL~\citep{agarwal2025unreasonable} derives reward from the policy's own entropy, but the questions on which certainty is optimized remain externally chosen.
VeriFree, RLPR, and NOVER~\citep{zhou2025reinforcing,yu2025rlpr,liu2025nover} use the likelihood or perplexity of a supplied reference as a soft consistency signal. They remove the need for a separate verifier while retaining the query and reference pairs that define the experience.
Similarly, execution feedback can provide $\mathcal{R}$ through program outcomes while the interpreter, unit-test interface, or game rules remain fixed.
Across these methods, the reward source changes while $(\Pi_{\mathrm{H}},\mathcal{E}_{\mathrm{H}})$ remains externally supplied.

\paragraph{Benefits and limits of fixed experience.}
Fixing the source of experience provides a stable basis for comparing reward signals because the target distribution does not change with the policy and an external source of grounding is preserved. With $\Pi_{\mathrm{H}}$ and $\mathcal{P}_{\mathrm{H}}$ held stationary, gains can be attributed more cleanly to policy optimization than to changes in the curriculum. This stability makes fixed experience a useful starting point, allowing a policy to learn from consensus, certainty, reference consistency, or world outcomes before it begins to generate new experience~\citep{zuo2025ttrl,agarwal2025unreasonable,zhou2025reinforcing,guo2025deepseek}.

The same stability also limits further improvement because the diversity and horizon of experience cannot exceed what $\Pi_{\mathrm{H}}$ and $\mathcal{E}_{\mathrm{H}}$ support. Repeated sampling and additional test-time compute can improve coverage on existing problems, but they do not broaden the task or environment distribution by themselves~\citep{s34new_brown2024monkeys,snell2024scaling}. Moving beyond this setting requires adaptation in \emph{what} problems are encountered, \emph{where} their consequences unfold, or both. Adaptive curricula and open-ended environment generation pursue this expansion by changing which tasks or environments enter learning~\citep{chen2025sec,wang2019poet,faldor2024omniepic}. A new reward source alone does not provide that adaptation.

\subsection{From Fixed to Generated Experience}
\label{subsec:l2_l3_transition}
Some recent methods generally relax one source of human design at a time. Some generate tasks while retaining a fixed environment, whereas others construct environments around externally supplied tasks or seeds, adapting $\Pi$ and $(\mathcal{S},\mathcal{P})$, respectively. Neither route by itself produces a continuing stream of both new tasks and the contexts needed to execute them, but each provides a building block for the integrated systems discussed in Section~\ref{subsec:l3_experience}.

\subsubsection{Task Generation in Fixed Environments}
\label{subsec:task_gen}
One way to reduce reliance on a human-curated task distribution is to let the model generate or select tasks while keeping the environment fixed. Depending on the domain, this fixed substrate may include a code interpreter, mathematical checker, or retrieval corpus. In the objective of Eq.~\ref{eq:general_goal}, queries are drawn from a model-parameterized proposer $\boldsymbol{q}\sim\Pi_{\theta}$ instead of the fixed distribution $\Pi_{\mathrm{H}}$, while the transition dynamics $\mathcal{P}$ and the verifier remain externally supplied. Generated tasks must be solvable, diverse, and calibrated to the policy's current ability. Proposer objectives often capture this requirement through \emph{learnability}, which favors tasks that remain informative without becoming trivial or effectively unsolvable. Because validation still relies on external dynamics and rules, these methods adapt the task distribution without yet adapting the environment.

\paragraph{Self-generated reasoning traces and instruction data.}
STaR~\citep{zelikman2022star} provides an early template by keeping the questions, environment, and answer-checking reward fixed while generating intermediate reasoning traces $\boldsymbol{z}$ and retaining only those that reach the correct answer $\boldsymbol{y}^*$. Later work extends generation from traces to the problems themselves. Self-Instruct~\citep{wang2022selfinstruct} synthesizes instruction, input, and output tuples and removes near-duplicates before fine-tuning, while Evol-Instruct~\citep{xu2023wizardlm} rewrites seed instructions with harder constraints or new topics. SeRL~\citep{fang2025serl} generates reasoning instances from a limited seed set, filters them by content, length, similarity, and difficulty, and trains on the retained instances with consensus-based rewards. CoT-Self-Instruct~\citep{yu2025cot} conditions synthesis on the intended domain and complexity, using answer consistency to filter verifiable tasks and a preference model to filter unverifiable ones. Ren et al.~\citep{ren2025beyond} control compositional difficulty by adding or removing constraints. The same pattern extends to agentic settings, where ZeroGUI~\citep{yang2025zerogui} proposes GUI tasks from the current interface and AgentSynth~\citep{xie2025agentsynth} combines individually solvable subtasks into longer computer-use tasks. These pipelines reduce dependence on human-authored corpora while leaving the verification environment fixed.

\paragraph{Proposer and solver self-play.}
Task generation can also be framed as an asymmetric game between a \emph{proposer} $\pi_{\theta_P}$ and a \emph{solver} $\pi_{\theta_S}$, which are often instantiated by the same base model in different roles. The proposer reward determines whether this game yields a useful curriculum. Rewarding only solver failure pushes $\Pi$ toward unsolvable noise, while rewarding only successful solutions encourages trivial tasks. A \emph{learnability} signal avoids both extremes by peaking at intermediate solver success. Writing $\bar{p}_{\theta_S}(\boldsymbol{q})$ for the empirical solve rate of a proposed query over $G$ solver rollouts,
\begin{equation}
\label{eq:learnability}
    \bar{p}_{\theta_S}(\boldsymbol{q}) = \frac{1}{G}\sum_{i=1}^{G}\mathbb{1}\!\left[\boldsymbol{y}_i = \hat{\boldsymbol{y}}^*(\boldsymbol{q})\right],
    \qquad
    \mathcal{R}_{\text{propose}}(\boldsymbol{q}) = \bar{p}_{\theta_S}(\boldsymbol{q})\bigl(1-\bar{p}_{\theta_S}(\boldsymbol{q})\bigr),
\end{equation}
so that queries that are neither always solved ($\bar{p}{=}1$) nor never solved ($\bar{p}{=}0$) receive the highest proposer reward. Absolute Zero Reasoner (AZR)~\citep{zhao2025absolute} lets the model act as both proposer and solver, combining a learnability score of the form in Eq.~\ref{eq:learnability} with environment-validated feedback to target the solver's competence frontier. R-Zero~\citep{huang2025r} trains a challenger to pose questions near that frontier while the solver learns from majority-vote pseudo-labels. SQLM~\citep{chen2025self} uses the same structure and obtains the proxy label $\hat{\boldsymbol{y}}^*$ through voting across independent solvers. Language Self-Play (LSP)~\citep{lsp} removes the initial dataset by formulating task generation as a minimax game:
\begin{equation}
\label{eq:selfplay}
    \max_{\theta_S}\ \min_{\theta_P}\ \mathbb{E}_{\boldsymbol{q}\sim\pi_{\theta_P},\ \boldsymbol{a}\sim\pi_{\theta_S}(\cdot|\boldsymbol{q})}\!\left[\mathcal{R}(\boldsymbol{q},\boldsymbol{a})\right].
\end{equation}
The minimax coupling allows the task distribution to respond to the solver's changing performance without an initial dataset. Recent work enriches the game with additional roles and internal signals. Multi-Agent Evolve~\citep{chen2025mae} uses proposer, solver, and judge roles from a shared backbone to estimate the difficulty and quality of open-ended questions without rule-checkable answers. Socratic-Zero~\citep{wang2025socratic} coordinates teacher, solver, and generator roles, allowing the teacher to target the solver's weaknesses and the generator to distill that teaching strategy into a scalable curriculum. AERO~\citep{gao2026aero} uses entropy-based positioning and counterfactual correction to locate the solver's learning frontier. G-Zero~\citep{huang2026gzero} replaces an external judge with an intrinsic $\text{Hint-}\delta$ reward that measures how a self-generated hint shifts the solver's predictive distribution. Across these methods, the task supply adapts while the surrounding environment or validation mechanism remains fixed. Extending the loop further requires evaluation and experience generation to adapt together.

\paragraph{Corpus- and document-grounded self-play.}
Because ungrounded self-play tends to recycle concepts the solver already knows, corpus-grounded methods draw task content from external documents. SPICE~\citep{liu2025spice} alternates one model between a Challenger that mines documents and writes reasoning problems and a Reasoner that solves them. SCOPE~\citep{kwan2026scope} applies a related design to open-ended tasks, with a Challenger that writes document-grounded questions near the Solver's frontier and a frozen copy of the base model that produces task-specific rubrics and grades the responses. Search Self-play~\citep{lu2025ssp} targets deep-search agents by generating search queries and using retrieval-augmented generation over each search trajectory to provide training targets. In each case, grounding expands the task content beyond the proposer's existing priors.

\paragraph{Reward-free self-training and self-generated curricula.}
Several boundary methods leave the source of tasks and the environment essentially unchanged but adapt how the model trains on its own outputs. SePT~\citep{li2025sept} alternates between response sampling and fine-tuning on refreshed self-generated traces without a reward model, verifier, or teacher. Self-Evolving Curriculum~\citep{chen2025sec} models task ordering as a non-stationary multi-armed bandit whose arms represent difficulty levels or problem categories and whose payoff is the absolute policy-gradient advantage. E2H Reasoner~\citep{parashar2025e2h} similarly orders tasks from easy to hard and gradually reduces exposure to easy problems. At test time, TTCS~\citep{yang2026ttcs} co-evolves a synthesizer and solver by generating local curriculum variants that help the solver bootstrap from easier neighboring problems under self-consistency rewards. OpenSIR~\citep{kwan2025opensir} combines diversity and difficulty rewards so that a teacher and student pair can expand beyond concepts represented by a minimal seed. These methods improve the organization of experience without independently changing both its task and environment sources.

\subsubsection{Environment Construction for Fixed or Seeded Tasks}
\label{subsec:env_construction}
A complementary route leaves the task distribution largely intact but lets the model construct or expand the environment $(\mathcal{S},\mathcal{P})$ in which experience is gathered. The limiting factor is often the cost and brittleness of the interactive world needed to produce and verify trajectories, not the availability of questions. This constraint is especially important for long-horizon agentic tasks~\citep{li2026recursive} in the sequential POMDP regime ($T{>}1$) introduced in Section~\ref{subsec:mdp}, where a human-authored simulator or verifier may account for much of the development effort. Current approaches use adversarial environment design, executable code synthesis, learned world models, live applications, and persistent memory to construct the dynamics and interfaces through which seeded tasks become useful experience.

\paragraph{Automatic environment design.}
Work on Unsupervised Environment Design (UED) predates LRMs, with PAIRED~\citep{dennis2020paired} formalizing the approach through a game between an environment-generating adversary and a protagonist and antagonist pair.
To avoid generating worlds that are merely impossible, the adversary maximizes \emph{regret} instead of the protagonist's failure. Regret is the gap between the return of the best-responding antagonist $\pi^{A}$ and the protagonist $\pi^{P}$ on a generated query $\boldsymbol{q}$,
\begin{equation}
\label{eq:regret}
    \mathrm{Regret}(\boldsymbol{q}) = U(\pi^{A},\boldsymbol{q}) - U(\pi^{P},\boldsymbol{q}),
\end{equation}
which becomes informative when the antagonist can solve an environment that challenges the protagonist.
For environment construction, regret serves the same purpose that the learnability signal in Eq.~\ref{eq:learnability} serves for task generation. Both create an automatic curriculum that favors experience near the policy's competence frontier. Subsequent UED variants refine the estimation of regret and the sampling of environment parameters, but retain the principle that useful environments should neither under-challenge nor overwhelm the current policy.

\paragraph{Environments as executable code.}
Representing the environment as code offers a practical route for LRMs. Eureka~\citep{ma2023eureka} uses an LLM to write reward code and refine it through simulator feedback, thereby automating part of reward design within $\mathcal{E}$. OMNI-EPIC~\citep{faldor2024omniepic} generates environment and termination code and then applies an interestingness model to retain tasks that are both learnable and novel. For tool use, Self-Challenging agents~\citep{zhou2025selfchallenging} pair each generated task with an executable verification function, allowing code execution to evaluate the resulting trajectory.

\paragraph{Scalable synthesis of executable tool environments.}
Because interactive environments are often scarcer than questions in agentic RL, recent work seeks to increase their supply without sacrificing reliability. Agent World Model~\citep{wang2026awm} decomposes an environment into a stateful backend, tool interface, and success criteria, then uses a staged LLM pipeline to synthesize database-backed environments and correct invalid components. ScaleEnv~\citep{tu2026scaleenv} expands tool-dependency graphs and verifies actions through execution, producing interactive environments together with verifiable tasks. RandomWorld~\citep{sullivan2025randomworld} procedurally generates tools and compositional tasks from a fine-grained type hierarchy. CuES~\citep{mai2025cues} explores a new environment through intrinsic curiosity and abstracts observed interactions into reusable task schemas. EvoEnv~\citep{shi2026evoenv} constructs Python environments through staged validation, semantic review, difficulty calibration, and novelty checks. Environment Tuning~\citep{lu2025envtuning} augments the environment with corrective feedback and a progress-based curriculum, allowing training to move beyond static trajectories.

\paragraph{Generative world models.}
An alternative to symbolic code is to learn the dynamics $\mathcal{P}$ directly. Genie~\citep{bruce2024genie} trains an action-controllable world model from unlabeled videos without ground-truth action labels, showing how $(\mathcal{S},\mathcal{P})$ can be learned from passive data instead of being specified entirely by hand.

\paragraph{Persistent and adaptive interaction contexts.}
Live applications and persistent memory offer another way to extend the context in which experience is gathered. ZeroGUI~\citep{yang2025zerogui} uses a vision-language model to generate tasks from the current interface and estimate rewards from trajectory screenshots, reducing its reliance on handwritten verifier scripts and annotated success trajectories. SEAgent~\citep{sun2025seagent} trains a World State Model to describe the state of a computer environment and evaluate actions along a trajectory. MemAgent~\citep{yu2025memagent} learns memory that can be read and updated across conversation turns, extending the horizon over which experience can accumulate. Voyager~\citep{wang2023voyager} combines an automatic curriculum with a growing library of executable skills in an embodied environment. More broadly, experience-driven lifelong learning frameworks~\citep{cai2025building} emphasize the accumulation and reuse of past interactions, consistent with the Experience Era view that sustained interaction can support durable capability growth~\citep{silver2025welcome}.

\subsection{Integrated Experience Generation}
\label{subsec:l3_experience}
Integrated experience generation corresponds to L3 in the ladder, where the task and environment pipeline can continue to produce training trajectories after receiving only limited external seeds. The policy need not create every artifact itself because another model may synthesize the experience, a corpus may supply its content, or a live environment may provide the interactions. Such sources can continue to ground learning without requiring people to curate each new task and setting. A method enters this regime when its generated or discovered experience supports continued training with limited additional curation.

\paragraph{Integrated task and environment pipelines.}
Several recent systems combine automatically supplied tasks, executable environments, and automatic feedback in one pipeline. ZeroGUI~\citep{yang2025zerogui} illustrates a boundary case in the GUI domain by using a vision-language model to propose tasks from a live interface and estimate rewards from trajectory screenshots under majority confirmation. Because the application dynamics and pretrained visual evaluator remain externally supplied, the method reduces task curation and handcrafted reward annotation without independently generating its verifier or environment.
Agent World Model~\citep{wang2026awm} uses a staged LLM pipeline with automated correction to synthesize stateful backends, tool interfaces, tasks, and success criteria from compact schemas. ScaleEnv~\citep{tu2026scaleenv} combines tool-dependency-graph expansion with executable action verification to construct solvable tasks and interactive environments across multiple domains. EvoEnv~\citep{shi2026evoenv} admits a synthesized Python environment only after validation, semantic review, difficulty calibration, and novelty checks. These systems reduce continued human curation by generating both the setting and the tasks used for interaction. CuES~\citep{mai2025cues} marks a boundary because it derives reusable task schemas through curiosity-driven exploration without handcrafted task seeds while retaining an externally supplied initial environment.

\paragraph{Boundary cases.}
When a proposer and solver loop invents mathematical questions while relying on a fixed Python interpreter, the task distribution adapts but execution and checking remain externally designed. A generative world model such as Genie~\citep{bruce2024genie} changes the environment while leaving the downstream task distribution and reward externally specified. These cases reduce human design along only one part of the experience axis. Integrated experience generation requires the generated or discovered tasks and the contexts needed to execute them to provide an ongoing sequence of $(\boldsymbol{q},\tau,\mathcal{R})$ for training. This criterion still permits seeds, corpora, and live applications to provide external grounding.

\subsection{Autonomous Co-evolution}
\label{subsec:co_construction}
Autonomous co-evolution corresponds to L4 in the ladder, where reward acquisition, task generation, environment construction, and policy improvement form a persistent adaptive loop in which changes to one component reshape the others. Humans may still define broad objectives, operational boundaries, and independent audits without maintaining the curriculum or evaluator throughout training. This regime connects current LRM research with earlier work on open-ended learning. POET~\citep{wang2019poet} co-evolves populations of environments and agents while transferring solutions across an expanding curriculum, although its reward remains externally specified. OMNI-EPIC~\citep{faldor2024omniepic} generates environment and reward code before filtering for tasks that are both learnable and interesting. Self-Rewarding Language Models~\citep{yuan2024self} couple reward and policy by using the policy as its own judge across successive iterations, while leaving the task distribution and environment human-designed. These examples couple different subsets of the learning loop, whereas the full regime requires experience generation and evaluation to continue improving together without an externally maintained curriculum.

\paragraph{Coupled generator and verifier loops.}
CURE~\citep{wang2025co} couples code and unit-test generation so that programs expose weaknesses in generated tests and stronger tests provide more informative feedback for subsequent programs. The executable runtime remains fixed, but the effective verification environment adapts with the generator. AZR~\citep{zhao2025absolute} and R-Zero~\citep{huang2025r} close the loop between proposer and solver while retaining a fixed execution or validation mechanism, making them examples of co-evolving curricula within an externally defined environment. Self-Challenging agents~\citep{zhou2025selfchallenging} synthesize \emph{Code-as-Task} instances that bundle an instruction, executable verification function, reference solution, and failure cases, thereby coupling task generation with executable feedback while leaving the underlying tool system externally defined.

\paragraph{Self-play and adaptive curricula.}
SPIRAL~\citep{liu2025spiral} trains a single policy in both roles of a two-player zero-sum language game against improving copies of itself. As the opponent strengthens under an objective of the form in Eq.~\ref{eq:selfplay}, the sequence of matchups becomes an adaptive curriculum that can elicit transferable reasoning strategies. Because the game rules remain fixed, SPIRAL adapts its opponents and curriculum without generating the underlying environment. The shifting opponent also introduces non-stationarity and higher gradient variance. Role-conditioned advantage estimation addresses this instability by normalizing each rollout against the expected return of its role, analogous to the group-relative baseline of GRPO in Eq.~\ref{eq:GRPO}. Closed-loop experience generation depends on the same recurrent relationship: policy improvements must induce more informative tasks and environments, which must then support further policy improvement. POET~\citep{wang2019poet} and OMNI-EPIC~\citep{faldor2024omniepic} pursue this form of open-ended coupling in RL settings~\citep{silver2025welcome}.

\paragraph{Difficulty-aligned agent and environment co-evolution.}
GenEnv~\citep{guo2025genenv} makes the environment a learned component by pairing an agent LLM with a generative simulator. An $\alpha$-curriculum reward steers the generated tasks toward the agent's zone of proximal development, extending the regret objective in Eq.~\ref{eq:regret} and the learnability signal in Eq.~\ref{eq:learnability} to a learned environment. Agent-World~\citep{dong2026agentworld} constructs an arena from databases and executable toolsets, then uses successive rounds of multi-environment RL to diagnose capability gaps and synthesize targeted tasks. Competitive game engines provide another source of adaptive experience. TextArena~\citep{guertler2025textarena} offers language games whose self-play dynamics create an automatic difficulty curriculum for social and strategic skills while providing a substrate for methods such as SPIRAL.

\paragraph{Open-ended self-improvement of the agent itself.}
Some systems extend co-evolution beyond tasks and environments to the design of the agent itself~\citep{liu2026structured,liu2026graphir,liu2026ai4ai}. The Darwin G\"odel Machine~\citep{zhang2025dgm} maintains an archive of coding agents that modify their own code and branch from archived ancestors to explore alternative designs. Foundation Model Self-Play~\citep{dharna2025fmsp} proposes and refines agent strategies through competitive play and quality-diversity search. Automated Capability Discovery~\citep{lu2025acd} uses one foundation model to propose and evaluate tasks for another, exposing capabilities and failures without relying on a fixed hand-designed benchmark.

\subsection{Failure Modes across the Experience Axis}
\label{subsec:exp_limitations}
As task and environment generation become adaptive, errors in the generated experience can feed directly into subsequent policy updates. On the task side, poor difficulty calibration and limited diversity can deprive the policy of useful learning signals. On the environment side, inaccurate or expensive simulations can undermine the trajectories they produce. Once these components are coupled, non-stationarity and mutual reinforcement create additional paths for errors to persist.

\paragraph{Difficulty miscalibration.}
If an unconstrained proposer drifts toward trivial or unsolvable tasks, group rewards become nearly constant and the GRPO advantages $\hat{A}_i$ in Eq.~\ref{eq:GRPO} vanish, causing learning to stall. The learnability signal in Eq.~\ref{eq:learnability}, the regret objective in Eq.~\ref{eq:regret}, and the frontier-targeting challenger in R-Zero~\citep{huang2025r} all seek to preserve an informative difficulty range. Their effectiveness depends on estimates of the solve rate $\bar{p}_{\theta_S}$, which may be noisy early in training.

\paragraph{Co-adaptive reward hacking.}
When $\mathcal{R}$, $\Pi$, and $(\mathcal{S},\mathcal{P})$ are generated within the learning loop, the solver may exploit blind spots in the verifier instead of acquiring genuine capability, extending the reward-hacking problem discussed in Section~\ref{sec:reward}. Drift becomes harder to detect when the verifier changes with the policy, and the generator and grader may converge on conventions that are internally consistent but externally invalid. Unit tests in CURE~\citep{wang2025co}, code verification in Self-Challenging agents~\citep{zhou2025selfchallenging}, and game outcomes in SPIRAL~\citep{liu2025spiral} constrain this behavior without eliminating systematic bias in the generated tests or games. Internal grading creates a further asymmetry because models may improve at producing answers without becoming equally reliable at checking them~\citep{chen2026selfverify}. Reliable verification may therefore require its own training objective and independent audits.

\paragraph{Distributional narrowing and mode collapse.}
Self-generated task streams can concentrate on regions the model already handles well, reducing the diversity of $\Pi$ over successive generations. Multidimensional filtering in SeRL~\citep{fang2025serl}, the adversarial objective of LSP~\citep{lsp} in Eq.~\ref{eq:selfplay}, and interestingness models in open-ended learning~\citep{faldor2024omniepic,wang2019poet} introduce diversity pressure that helps keep the curriculum expanding.

\paragraph{Non-stationarity and instability.}
In fully coupled loops, the environment or opponent changes as the policy learns, departing from the fixed data and reward assumptions common in single-turn RLVR. The resulting non-stationarity can increase gradient variance in multi-agent training, although role-conditioned or group-relative baselines~\citep{liu2025spiral} can improve stability. It also compounds the long-horizon credit-assignment difficulty of the $T{>}1$ regime, where a single scalar outcome must be distributed over an entire self-generated trajectory.

\paragraph{Environment fidelity and cost.}
The reliability of a constructed environment depends on the models that build and evaluate it. Vision-language reward estimation~\citep{yang2025zerogui} and world-state modeling~\citep{sun2025seagent} inherit errors and biases from their underlying models, which can propagate through every trajectory collected in the environment. Verifying a generated simulator or test suite may also reintroduce much of the supervision cost that autonomous experience generation seeks to reduce. Evidence from closed-loop self-evolution further suggests that internally generated supervision may plateau below oracle supervision~\citep{qi2026generalization}, supporting the use of independent anchors even in a largely autonomous experience pipeline.

\paragraph{Domain landscape and evaluation implications.}
The main constraint on autonomous experience generation differs by domain. In mathematics and code, executable checkers provide relatively strong grounding for generated tasks, although proposers may still collapse to recycled templates without corpus grounding~\citep{liu2025spice,kwan2025opensir}. In tool use and computer control, environment fidelity is often a tighter constraint than question scarcity~\citep{wang2026awm,tu2026scaleenv,zhou2025selfchallenging}. Evaluation must therefore measure simulator correctness and transfer to held-out applications as discussed in Section~\ref{subsec:eval_experience}. In open-ended and social settings, self-play can produce an adaptive curriculum~\citep{liu2025spiral,guertler2025textarena}, but the same coupling can amplify collusion between the proposer and judge~\citep{yuan2024self}.

The goal is not to maximize autonomy at the expense of grounding. Stable learning may still require consensus, execution feedback, corpus grounding, or occasional human audits. The relevant question is whether continual human curation can be reduced while enough independent evidence remains to detect drift and collapse, which motivates the evaluation criteria developed in the next section.


\definecolor{evalCap}{RGB}{238,142,176}
\definecolor{evalRew}{RGB}{255,190,100}
\definecolor{evalExp}{RGB}{126,205,151}
\definecolor{evalPro}{RGB}{190,151,230}

\tikzset{
  evalroot/.style={rectangle,rounded corners,draw=black!65,fill=black!4,
    line width=.8pt,align=center,inner xsep=5pt,inner ysep=5pt},
  evalbranch/.style={rectangle,rounded corners,line width=.65pt,
    align=center,inner xsep=4pt,inner ysep=3.5pt},
  evalleaf/.style={rectangle,rounded corners,fill=white,line width=.55pt,
    align=left,inner xsep=4pt,inner ysep=3.5pt}
}

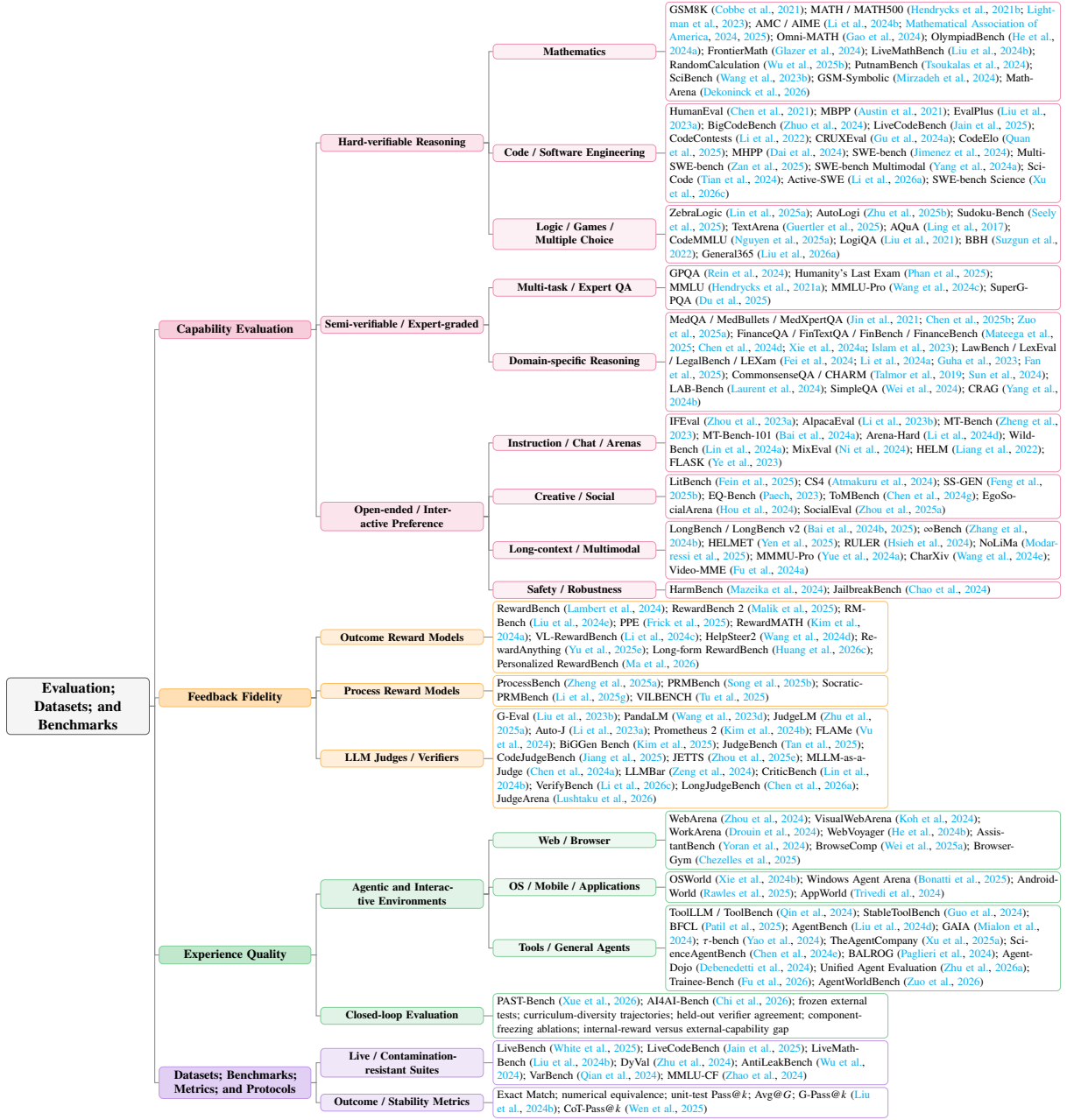
\begin{figure*}[!t]
  \centering
  \resizebox{\textwidth}{!}{%
  \begin{forest}
    for tree={
      grow=east,
      reversed=true,
      anchor=center,
      base=center,
      parent anchor=east,
      child anchor=west,
      edge={draw=black!48,line width=.52pt},
      forked edges,
      fork sep=5pt,
      l sep=9pt,
      s sep=2.2pt,
      font=\normalsize
    },
    where level=0{evalroot,font=\Large\bfseries,text width=12em}{},
    where level=1{evalbranch,font=\large\bfseries,text width=13em}{},
    where level=2{evalbranch,font=\normalsize\bfseries,text width=14em}{},
    where level=3{evalbranch,font=\normalsize\bfseries,text width=14em}{},
    where n children=0{evalleaf,font=\normalsize\normalfont,text width=35em}{},
    [,content={Evaluation; Datasets; and Benchmarks}
      [,content={Capability Evaluation},fill=evalCap!45,draw=evalCap
        [,content={Hard-verifiable Reasoning},fill=evalCap!24,draw=evalCap
          [,content={Mathematics},fill=evalCap!13,draw=evalCap
            [,content={GSM8K~\citep{cobbe2021gsm8k}; MATH / MATH500~\citep{hendrycks2021MATH,lightman2023math500};
              AMC / AIME~\citep{li2024numinamath,AIME2024,AIME2025};
              Omni-MATH~\citep{gao2024OmniMATH}; OlympiadBench~\citep{he2024OlympiadBench};
              FrontierMath~\citep{glazer2024FrontierMath}; LiveMathBench~\citep{liu2024LiveMathBench};
              RandomCalculation~\citep{wu2025RandomCalculation}; PutnamBench~\citep{tsoukalas2024PutnamBench};
              SciBench~\citep{wang2023SciBench}; GSM-Symbolic~\citep{mirzadeh2024GSMSymbolic};
              MathArena~\citep{dekoninck2026matharena}},draw=evalCap]
          ]
          [,content={Code / Software Engineering},fill=evalCap!13,draw=evalCap
            [,content={HumanEval~\citep{chen2021HumanEval}; MBPP~\citep{austin2021MBPP};
              EvalPlus~\citep{liu2023EvalPlus}; BigCodeBench~\citep{zhuo2024BigCodeBench};
              LiveCodeBench~\citep{jainlivecodebench}; CodeContests~\citep{li2022competition};
              CRUXEval~\citep{gu2024cruxeval}; CodeElo~\citep{quan2025codeelo}; MHPP~\citep{dai2024mhpp};
              SWE-bench~\citep{jimenez2024SWEBench}; Multi-SWE-bench~\citep{zan2025MultiSWEbench};
              SWE-bench Multimodal~\citep{yang2024SWEbenchMultimodal}; SciCode~\citep{tian2024SciCode};
              Active-SWE~\citep{li2026activeswe}; SWE-bench Science~\citep{xu2026swebenchscience}},draw=evalCap]
          ]
          [,content={Logic / Games / Multiple Choice},fill=evalCap!13,draw=evalCap
            [,content={ZebraLogic~\citep{lin2025zebralogic}; AutoLogi~\citep{zhu2025autologi};
              Sudoku-Bench~\citep{seely2025sudoku}; TextArena~\citep{guertler2025textarena};
              AQuA~\citep{ling2017program}; CodeMMLU~\citep{nguyen2025codemmlu};
              LogiQA~\citep{liu2021logiqa}; BBH~\citep{suzgun2022BBH};
              General365~\citep{liu2026general365}},draw=evalCap]
          ]
        ]
        [,content={Semi-verifiable / Expert-graded},fill=evalCap!24,draw=evalCap
          [,content={Multi-task / Expert QA},fill=evalCap!13,draw=evalCap
            [,content={GPQA~\citep{rein2024gpqa}; Humanity's Last Exam~\citep{phan2025humanity};
              MMLU~\citep{hendrycks2021MMLU}; MMLU-Pro~\citep{wang2024mmlupro};
              SuperGPQA~\citep{du2025SuperGPQA}},draw=evalCap]
          ]
          [,content={Domain-specific Reasoning},fill=evalCap!13,draw=evalCap
            [,content={MedQA / MedBullets / MedXpertQA~\citep{jin2021MedQA,chen2025Medbullets,zuo2025medxpertqa};
              FinanceQA / FinTextQA / FinBench / FinanceBench~\citep{mateega2025financeqa,chen2024fintextqa,xie2024finben,islam2023financebench};
              LawBench / LexEval / LegalBench / LEXam~\citep{fei2024lawbench,li2024lexeval,guha2023legalbench,fan2025lexam};
              CommonsenseQA / CHARM~\citep{talmor2019commonsenseqa,sun2024CHARM};
              LAB-Bench~\citep{laurent2024LABBench}; SimpleQA~\citep{wei2024SimpleQA}; CRAG~\citep{yang2024CRAG}},draw=evalCap]
          ]
        ]
        [,content={Open-ended / Interactive Preference},fill=evalCap!24,draw=evalCap
          [,content={Instruction / Chat / Arenas},fill=evalCap!13,draw=evalCap
            [,content={IFEval~\citep{zhou2023IFEval}; AlpacaEval~\citep{li2023alpacaeval};
              MT-Bench~\citep{zheng2023MTBench}; MT-Bench-101~\citep{bai2024MTBench101};
              Arena-Hard~\citep{li2024ArenaHard}; WildBench~\citep{lin2024WildBench};
              MixEval~\citep{ni2024MixEval}; HELM~\citep{liang2022HELM}; FLASK~\citep{ye2023FLASK}},draw=evalCap]
          ]
          [,content={Creative / Social},fill=evalCap!13,draw=evalCap
            [,content={LitBench~\citep{fein2025litbench}; CS4~\citep{atmakuru2024cs4};
              SS-GEN~\citep{feng2025ssgn}; EQ-Bench~\citep{paech2023eqbench};
              ToMBench~\citep{chen2024tombench}; EgoSocialArena~\citep{hou2024egosocialarena};
              SocialEval~\citep{zhou2025socialeval}},draw=evalCap]
          ]
          [,content={Long-context / Multimodal},fill=evalCap!13,draw=evalCap
            [,content={LongBench / LongBench v2~\citep{bai2024LongBench,bai2024LongBenchv2};
              $\infty$Bench~\citep{zhang2024InfinityBench}; HELMET~\citep{yen2025HELMET};
              RULER~\citep{hsieh2024RULER}; NoLiMa~\citep{modarressi2025NoLiMa};
              MMMU-Pro~\citep{yue2024MMMUPro}; CharXiv~\citep{wang2024CharXiv};
              Video-MME~\citep{fu2024VideoMME}},draw=evalCap]
          ]
          [,content={Safety / Robustness},fill=evalCap!13,draw=evalCap
            [,content={HarmBench~\citep{mazeika2024HarmBench}; JailbreakBench~\citep{chao2024JailbreakBench}},draw=evalCap]
          ]
        ]
      ]
      [,content={Feedback Fidelity},fill=evalRew!48,draw=evalRew
        [,content={Outcome Reward Models},fill=evalRew!24,draw=evalRew
          [,content={RewardBench~\citep{lambert2024rewardbench}; RewardBench 2~\citep{malik2025rewardbench};
            RM-Bench~\citep{liu2024rm}; PPE~\citep{frick2024PPE};
            RewardMATH~\citep{kim2024RewardMATH}; VL-RewardBench~\citep{li2024VLRewardBench};
            HelpSteer2~\citep{wang2024HelpSteer2}; RewardAnything~\citep{yu2025rewardanything};
            Long-form RewardBench~\citep{huang2026longformrewardbench};
            Personalized RewardBench~\citep{ma2026personalizedrewardbench}},draw=evalRew]
        ]
        [,content={Process Reward Models},fill=evalRew!24,draw=evalRew
          [,content={ProcessBench~\citep{zheng2024ProcessBench}; PRMBench~\citep{song2025PRMBench};
            Socratic-PRMBench~\citep{li2025socratic}; VILBENCH~\citep{tu2025vilbench}},draw=evalRew]
        ]
        [,content={LLM Judges / Verifiers},fill=evalRew!24,draw=evalRew
          [,content={G-Eval~\citep{liu2023GEval}; PandaLM~\citep{wang2023PandaLM};
            JudgeLM~\citep{zhu2025judgelm}; Auto-J~\citep{li2023AutoJ};
            Prometheus 2~\citep{kim2024prometheus}; FLAMe~\citep{vu2024FLAMe};
            BiGGen Bench~\citep{kim2024BigGenBench}; JudgeBench~\citep{tan2024judgebench};
            CodeJudgeBench~\citep{jiang2025codejudgebench}; JETTS~\citep{zhou2025evaluating};
            MLLM-as-a-Judge~\citep{chen2024mllm}; LLMBar~\citep{zeng2024LLMBar};
            CriticBench~\citep{lin2024CriticBench}; VerifyBench~\citep{li2026verifybench};
            LongJudgeBench~\citep{chen2026longjudgebench}; JudgeArena~\citep{lushtaku2026judgearena}},draw=evalRew]
        ]
      ]
      [,content={Experience Quality},fill=evalExp!48,draw=evalExp
        [,content={Agentic and Interactive Environments},fill=evalExp!24,draw=evalExp
          [,content={Web / Browser},fill=evalExp!14,draw=evalExp
            [,content={WebArena~\citep{zhou2024WebArena}; VisualWebArena~\citep{koh2024VisualWebArena};
              WorkArena~\citep{drouin2024WorkArena}; WebVoyager~\citep{he2024WebVoyager};
              AssistantBench~\citep{yoran2024AssistantBench}; BrowseComp~\citep{wei2025BrowseComp};
              BrowserGym~\citep{chezelles2025BrowserGym}},draw=evalExp]
          ]
          [,content={OS / Mobile / Applications},fill=evalExp!14,draw=evalExp
            [,content={OSWorld~\citep{xie2024OSWorld}; Windows Agent Arena~\citep{bonatti2024WindowsAgentArena};
              AndroidWorld~\citep{rawles2024AndroidWorld}; AppWorld~\citep{trivedi2024AppWorld}},draw=evalExp]
          ]
          [,content={Tools / General Agents},fill=evalExp!14,draw=evalExp
            [,content={ToolLLM / ToolBench~\citep{qin2024ToolLLM}; StableToolBench~\citep{guo2024StableToolBench};
              BFCL~\citep{patil2025BFCL}; AgentBench~\citep{liu2023AgentBench};
              GAIA~\citep{mialon2024GAIA}; $\tau$-bench~\citep{yao2024Taubench};
              TheAgentCompany~\citep{xu2025AgentCompany}; ScienceAgentBench~\citep{chen2024ScienceAgentBench};
              BALROG~\citep{paglieri2024BALROG}; AgentDojo~\citep{debenedetti2024AgentDojo};
              Unified Agent Evaluation~\citep{zhu2026unifiedagent};
              Trainee-Bench~\citep{fu2026traineebench};
              AgentWorldBench~\citep{zuo2026qwenagentworld}},draw=evalExp]
          ]
        ]
        [,content={Closed-loop Evaluation},fill=evalExp!24,draw=evalExp
          [,content={PAST-Bench~\citep{xue2026pastbench}; AI4AI-Bench~\citep{chi2026ai4ai};
            frozen external tests; curriculum-diversity trajectories; held-out verifier agreement;
            component-freezing ablations; internal-reward versus external-capability gap},draw=evalExp]
        ]
      ]
      [,content={Datasets; Benchmarks; Metrics; and Protocols},fill=evalPro!45,draw=evalPro
        [,content={Live / Contamination-resistant Suites},fill=evalPro!23,draw=evalPro
          [,content={LiveBench~\citep{white2025LiveBench}; LiveCodeBench~\citep{jainlivecodebench};
            LiveMathBench~\citep{liu2024LiveMathBench}; DyVal~\citep{zhu2023DyVal};
            AntiLeakBench~\citep{wu2024AntiLeakBench}; VarBench~\citep{qian2024VarBench};
            MMLU-CF~\citep{zhao2024MMLUCF}},draw=evalPro]
        ]
        [,content={Outcome / Stability Metrics},fill=evalPro!23,draw=evalPro
          [,content={Exact Match; numerical equivalence; unit-test Pass@$k$; Avg@$G$;
            G-Pass@$k$~\citep{liu2024LiveMathBench}; CoT-Pass@$k$~\citep{wen2025reinforcement}},draw=evalPro]
        ]
      ]
    ]
  \end{forest}}
  \caption{\textbf{Evaluation taxonomy for scaling beyond human supervision.}
  A complete evaluation jointly assesses capability on frozen benchmarks, the fidelity of training feedback, the quality of generated experience, and reproducibility under contamination-aware protocols.}
  \label{fig:evaluation_taxonomy}
\end{figure*}

\section{Evaluation, Datasets, and Benchmarks}
\label{sec:evaluation}

Evaluating learning beyond human supervision requires evidence about three distinct objects: the policy, the feedback used to train it, and the experience from which it learns. A higher held-out score establishes a change in capability, but does not establish that the training reward remained faithful or that a generated curriculum remained broad and valid. Our evaluation framework consequently separates policy capability, feedback fidelity, and experience quality. These forms of evidence become jointly important as reward acquisition and experience generation move into the learning loop.

The balance among these three forms of evidence changes across the ladder in Sec.~\ref{subsec:continuum}. At L0, human judgments directly define the held-out standard. At L1 and L2, evaluation must also test the reusable evaluators and reward sources used for learning. At L3, task and environment quality become explicit evaluation targets, while L4 requires longitudinal evidence about the stability of the coupled learning loop. Tables~\ref{tab:evaluation_benchmarks} and~\ref{tab:evaluation_metrics} summarize representative suites and metrics. Table~\ref{tab:eval_by_level} translates the three-part framework into reporting requirements for the five ladder levels.

\subsection{Capability Evaluation}
\label{subsec:eval_capability}

Capability evaluation measures whether the policy $\pi_\theta$ solves the intended problems under a held-out standard. The source of the training reward does not determine the appropriate capability test. A policy trained with model-derived rewards may still be evaluated on human-authored mathematics problems, while an agent trained in evolving environments may still be tested in a fixed sandbox. We group capability benchmarks by how their outputs are assessed because this choice determines the reliability, cost, and scalability of evaluation.

\subsubsection{Hard-verifiable reasoning}

Hard-verifiable tasks admit deterministic checking through exact numerical or symbolic matching, unit tests, game outcomes, or constraint satisfaction. They support scalable evaluation and provide a comparatively reliable external anchor for methods trained with reduced supervision.

\paragraph{Mathematical reasoning.}
Mathematics is among the most widely used domains for evaluating LRMs~\citep{ahn2024large}. Established suites include GSM8K~\citep{cobbe2021gsm8k}, MATH and MATH500~\citep{hendrycks2021MATH,lightman2023math500}, and competition problems from AMC and AIME~\citep{li2024numinamath,AIME2024,AIME2025}. Omni-MATH~\citep{gao2024OmniMATH}, OlympiadBench~\citep{he2024OlympiadBench}, FrontierMath~\citep{glazer2024FrontierMath}, LiveMathBench~\citep{liu2024LiveMathBench}, and RandomCalculation~\citep{wu2025RandomCalculation} extend coverage to harder or more dynamic settings. PutnamBench evaluates formal theorem proving~\citep{tsoukalas2024PutnamBench}, while SciBench tests college-level scientific problem solving~\citep{wang2023SciBench}.

MathArena broadens this landscape with a continuously maintained collection spanning olympiad problems, proof-oriented competitions, research questions drawn from arXiv, and formal proof generation in Lean~\citep{dekoninck2026matharena}. Its tasks distinguish final-answer accuracy from proof construction and formalization. This distinction matters because Pass@$1$ does not reveal whether a model is consistent across samples or whether its intermediate reasoning is valid. Pass@$k$, mean sample accuracy, G-Pass@$k$, and CoT-Pass@$k$ provide complementary evidence, as discussed in Sec.~\ref{subsec:eval_metrics}. Results also benefit from difficulty and subset breakdowns. A single aggregate on MATH can conceal whether gains are confined to routine algebra or extend to olympiad-level problems~\citep{gao2024OmniMATH,glazer2024FrontierMath}. Fixed grading scripts further improve comparisons across models. GSM-Symbolic adds a robustness check by changing names, values, and templates while preserving the underlying reasoning structure~\citep{mirzadeh2024GSMSymbolic}.

\paragraph{Code generation and software engineering.}
Executable tests provide a direct measure of functional correctness in code generation. HumanEval~\citep{chen2021HumanEval}, MBPP~\citep{austin2021MBPP}, and EvalPlus~\citep{liu2023EvalPlus} are widely used for this purpose. BigCodeBench~\citep{zhuo2024BigCodeBench}, LiveCodeBench~\citep{jainlivecodebench}, CodeContests~\citep{li2022competition}, CRUXEval~\citep{gu2024cruxeval}, CodeElo~\citep{quan2025codeelo}, and MHPP~\citep{dai2024mhpp} expand functional and distributional coverage. Similarity metrics such as CodeBLEU~\citep{ren2020codebleu}, CodeBERTScore~\citep{zhou2023codebertscore}, and ICE-Score~\citep{zhuo2024ice} can complement execution, but they are less direct measures of task completion when reliable tests are available.

Repository-level benchmarks assess a broader form of software engineering. SWE-bench~\citep{jimenez2024SWEBench}, Multi-SWE-bench~\citep{zan2025MultiSWEbench}, and SWE-bench Multimodal~\citep{yang2024SWEbenchMultimodal} evaluate repository patches, multilingual issue resolution, and visually grounded tasks. Active-SWE removes the supplied issue report and requires agents to discover and repair defects~\citep{li2026activeswe}. SWE-bench Science covers issue-driven repair, expert exploration, and system integration in scientific repositories~\citep{xu2026swebenchscience}. These suites evaluate problem identification, tool use, environment interaction, and patch generation together. For methods trained with executable rewards, the evaluation harness must be disjoint from the training tests to limit leakage. SciCode provides an additional transfer test through scientist-curated research programming problems~\citep{tian2024SciCode}.

\paragraph{Logic puzzles, games, and multiple choice.}
ZebraLogic~\citep{lin2025zebralogic}, AutoLogi~\citep{zhu2025autologi}, and Sudoku-Bench~\citep{seely2025sudoku} evaluate structured reasoning through constraint satisfaction. TextArena uses competitive language games to assess strategic behavior through observable outcomes~\citep{guertler2025textarena}. Multiple-choice suites such as AQuA~\citep{ling2017program}, CodeMMLU~\citep{nguyen2025codemmlu}, LogiQA~\citep{liu2021logiqa}, and BBH~\citep{suzgun2022BBH} enable inexpensive and reproducible comparisons. Their format can understate the difficulty of generating a complete solution and can reward option elimination. General365 addresses a different confound by using reasoning problems that require only K-12 background knowledge~\citep{liu2026general365}. Its controlled variants help separate reasoning difficulty from specialized factual knowledge.

\begin{table*}[!t]
\centering
\scriptsize
\setlength{\tabcolsep}{3.2pt}
\renewcommand{\arraystretch}{1.13}
\caption{Representative suites and protocols for evaluating large reasoning models. \textbf{Target} identifies the component being evaluated. \textbf{Evidence} records the external signal used to support the measurement. The selection emphasizes coverage across evaluation targets and is not exhaustive.}
\label{tab:evaluation_benchmarks}
\resizebox{\linewidth}{!}{%
\begin{tabular}{@{}lllll@{}}
\toprule
\textbf{Suite or protocol} & \textbf{Target} & \textbf{Domain} & \textbf{Evidence} & \textbf{Representative metrics} \\
\midrule
\multicolumn{5}{@{}l}{\emph{Policy capability}} \\
MATH / MATH500~\citep{hendrycks2021MATH,lightman2023math500}
& policy & mathematics & reference answers & Acc. / Pass@$k$ \\
Omni-MATH / FrontierMath~\citep{gao2024OmniMATH,glazer2024FrontierMath}
& policy & advanced mathematics & exact or expert grading & Acc. / Pass@$k$ \\
LiveBench / DyVal~\citep{white2025LiveBench,zhu2023DyVal}
& policy & multiple domains & refreshed or perturbed ground truth & Acc. \\
HumanEval / EvalPlus~\citep{chen2021HumanEval,liu2023EvalPlus}
& policy & code generation & held-out unit tests & Pass@$k$ \\
BigCodeBench / LiveCodeBench~\citep{zhuo2024BigCodeBench,jainlivecodebench}
& policy & code generation & execution & Pass@$k$ \\
SWE-bench~\citep{jimenez2024SWEBench}
& policy & software engineering & repository tests & resolve rate \\
GPQA / HLE~\citep{rein2024gpqa,phan2025humanity}
& policy & expert reasoning & expert-authored labels & Acc. \\
SimpleQA / CRAG~\citep{wei2024SimpleQA,yang2024CRAG}
& policy & factuality and RAG & references or retrieval-backed grading & Acc. / graded score \\
Arena-Hard / WildBench~\citep{li2024ArenaHard,lin2024WildBench}
& policy & open-ended interaction & frozen judges or human preferences & win rate \\
LongBench / HELMET / RULER~\citep{bai2024LongBench,yen2025HELMET,hsieh2024RULER}
& policy & long context & task-specific ground truth & Acc. / recall \\
WebArena / OSWorld~\citep{zhou2024WebArena,xie2024OSWorld}
& policy & web and computer use & executable task success & success rate \\
AgentBench / GAIA~\citep{liu2023AgentBench,mialon2024GAIA}
& policy & general agents & task completion & Acc. / success rate \\
\midrule
\multicolumn{5}{@{}l}{\emph{Feedback fidelity}} \\
RewardBench / RewardBench~2~\citep{lambert2024rewardbench,malik2025rewardbench}
& feedback & outcome reward models & disjoint preference data & pairwise / best-of-$N$ Acc. \\
RM-Bench / PPE~\citep{liu2024rm,frick2024PPE}
& feedback & reward robustness and utility & controlled rewrites / downstream outcomes & Acc. / correlation \\
ProcessBench / PRMBench~\citep{zheng2024ProcessBench,song2025PRMBench}
& feedback & process reward models & step-error annotations & step F1 / PRM-Score \\
Socratic-PRMBench / VILBENCH~\citep{li2025socratic,tu2025vilbench}
& feedback & text and multimodal reasoning & fine-grained error labels & PRM-Score \\
JudgeBench / CodeJudgeBench~\citep{tan2024judgebench,jiang2025codejudgebench}
& feedback & LLM judges & verifiable response pairs & Acc. / consistency \\
JETTS~\citep{zhou2025evaluating}
& feedback & judge utility & downstream search and refinement & quality gain \\
VerifyBench / LongJudgeBench~\citep{li2026verifybench,chen2026longjudgebench}
& feedback & expert and long-form judging & expert answers / controlled protocols & Acc. / agreement \\
LLMBar / CriticBench~\citep{zeng2024LLMBar,lin2024CriticBench}
& feedback & instruction judging and critique & adversarial pairs / correction targets & Acc. / correction gain \\
\midrule
\multicolumn{5}{@{}l}{\emph{Experience quality}} \\
Generated-task audit
& experience & self-generated curricula & frozen solver and external seeds & solve-rate profile / learnability / diversity \\
Corpus-grounded task audit~\citep{liu2025spice,kwan2026scope}
& experience & document-grounded curricula & source attribution & novelty / source coverage \\
Synthesized-environment audit
& experience & interactive environments & executable validation and frozen agents & validity / fidelity / transfer \\
AgentWorldBench~\citep{zuo2026qwenagentworld}
& experience & learned world models & observed environment transitions & transition fidelity \\
PAST-Bench~\citep{xue2026pastbench}
& experience & persistent agents & matched sequences with and without memory & retained-experience gain \\
AI4AI-Bench~\citep{chi2026ai4ai}
& experience & algorithm improvement & fixed repository and evaluator & external improvement \\
Closed-loop audit
& experience & co-evolving learning loops & frozen external tests and component ablations & diversity / drift / stability \\
\bottomrule
\end{tabular}}
\end{table*}
\begin{table*}[!t]
\centering
\footnotesize
\setlength{\tabcolsep}{5pt}
\renewcommand{\arraystretch}{1.2}
\caption{Evaluation metrics grouped by the object they measure. No metric is sufficient in isolation when the feedback or experience stream is generated within the learning loop.}
\label{tab:evaluation_metrics}
\resizebox{\linewidth}{!}{%
\begin{tabular}{@{}lll@{}}
\toprule
\textbf{Family} & \textbf{Representative metrics} & \textbf{Primary use} \\
\midrule
Outcome correctness
& Exact match, numerical equivalence, unit-test Pass@$k$, success rate
& Verifiable capability \\
Sampling reliability
& Avg@$G$, G-Pass@$k$~\citep{liu2024LiveMathBench}, CoT-Pass@$k$~\citep{wen2025reinforcement}
& Typical performance, consistency, and reasoning validity \\
Preference agreement
& Pairwise Acc.~\citep{lambert2024rewardbench}, listwise Acc., best-of-$N$ Acc.
& Outcome reward models and LLM judges \\
Judge utility and bias
& Reranking gain~\citep{zhou2025evaluating}, human-judge agreement, position and style sensitivity
& Practical value and robustness of $\pi_v$ \\
Process fidelity
& Step F1, PRM-Score~\citep{li2025socratic,song2025PRMBench}
& Error detection within reasoning traces \\
Open-ended quality
& Win rate, rubric score, G-Eval~\citep{liu2023GEval}
& Open-ended capability under a specified evaluator \\
Task-supply quality
& Solve-rate distribution, $\bar{p}(1-\bar{p})$, diversity, novelty, rejection rate
& Self-generated curricula \\
Environment validity
& Executability, validation pass rate, transition fidelity, frozen-agent solve rate
& Synthesized environments \\
Closed-loop health
& Internal reward versus external capability, diversity over time, non-stationarity, component-freezing ablations
& Co-evolving learning loops \\
Evaluation validity
& Contamination~\citep{singh2024ConTAM,dekoninck2024ConStat}, position bias~\citep{shi2024judging}, protocol variance
& Trustworthy comparison \\
\bottomrule
\end{tabular}}
\end{table*}
\begin{table*}[!t]
\centering
\footnotesize
\setlength{\tabcolsep}{5pt}
\renewcommand{\arraystretch}{1.2}
\caption{Minimum evaluation evidence for claims at each ladder level. Capability remains an evaluation target throughout the ladder. Feedback and experience require additional source-matched audits as their provision moves into the learning loop.}
\label{tab:eval_by_level}
\resizebox{\linewidth}{!}{%
\begin{tabular}{@{}p{0.07\textwidth}p{0.27\textwidth}p{0.29\textwidth}p{0.31\textwidth}@{}}
\toprule
\textbf{Level} & \textbf{Policy capability} & \textbf{Feedback fidelity} & \textbf{Experience quality} \\
\midrule
L0
& Held-out performance under a fixed evaluation protocol
& Annotation quality, agreement, and coverage
& Separation of human-supplied training and evaluation instances \\

L1
& Domain-matched capability and transfer under an external evaluator
& Accuracy, calibration, robustness, and bias of the reusable evaluator on disjoint data
& Coverage and distribution shift within externally supplied tasks and environments \\

L2
& Held-out capability with stability or reasoning-validity metrics when relevant
& Audits matched to the reward source, including calibration, consensus fragility, reference dependence, and checker validity
& Generalization across fixed externally supplied tasks and environments \\

L3
& Capability and transfer on frozen suites disjoint from the generated training stream
& Agreement with an independent verifier, executable checker, or sparse human audit
& Task diversity, learnability, novelty, environment validity, solvability, and transfer to external experience \\

L4
& External capability and transfer tracked throughout co-evolution
& Drift between internal reward and independent external evaluation, including generator-verifier asymmetry
& Closed-loop stability, non-stationarity, diversity over time, the relation between internal reward and external capability, and component-freezing ablations \\
\bottomrule
\end{tabular}}
\end{table*}

\subsubsection{Semi-verifiable and expert-graded tasks}

Some tasks have concise answers but require expertise or retrieval-backed evidence for reliable grading. GPQA~\citep{rein2024gpqa}, Humanity's Last Exam~\citep{phan2025humanity}, MMLU~\citep{hendrycks2021MMLU}, MMLU-Pro~\citep{wang2024mmlupro}, and SuperGPQA~\citep{du2025SuperGPQA} test broad knowledge and multi-step reasoning across disciplines. Their expensive labels make them useful external anchors for assessing transfer beyond mathematics and code.

Domain-specific suites cover medicine~\citep{jin2021MedQA,chen2025Medbullets,zuo2025medxpertqa}, finance~\citep{mateega2025financeqa,chen2024fintextqa,xie2024finben,islam2023financebench}, law~\citep{fei2024lawbench,li2024lexeval,guha2023legalbench,fan2025lexam}, commonsense reasoning~\citep{talmor2019commonsenseqa,sun2024CHARM}, and practical biology research~\citep{laurent2024LABBench}. SimpleQA isolates short-form factuality~\citep{wei2024SimpleQA}, while CRAG evaluates retrieval-augmented generation~\citep{yang2024CRAG}. These benchmarks retain an external grading standard even when the model was trained with rewards that did not depend on direct human evaluation. They consequently measure transfer under residual expert grounding, not the independence of the training reward itself.

\subsubsection{Open-ended generation and interactive preference}

Open-ended tasks lack a unique reference answer, so evaluation relies on preferences, rubrics, or live comparison. This setting creates a particular risk of evaluator entanglement when related LLM judges are used for both training and testing.

\paragraph{Instruction following, chat, and arenas.}
IFEval~\citep{zhou2023IFEval}, AlpacaEval~\citep{li2023alpacaeval,dubois2024length}, MT-Bench~\citep{zheng2023MTBench}, MT-Bench-101~\citep{bai2024MTBench101}, Arena-Hard~\citep{li2024ArenaHard}, WildBench~\citep{lin2024WildBench}, and MixEval~\citep{ni2024MixEval} evaluate instruction following and conversational quality. HELM~\citep{liang2022HELM} and FLASK~\citep{ye2023FLASK} provide broader evaluation frameworks. Arena-style win rates capture comparative preference at scale, but remain sensitive to verbosity, response order, and judge choice~\citep{shi2024judging,dubois2024length}. A stronger protocol fixes the judge panel and measures its agreement with an independent human subset.

\paragraph{Creative, social, and long-context evaluation.}
LitBench~\citep{fein2025litbench}, CS4~\citep{atmakuru2024cs4}, and SS-GEN~\citep{feng2025ssgn} probe creative and narrative writing. EQ-Bench~\citep{paech2023eqbench}, ToMBench~\citep{chen2024tombench}, EgoSocialArena~\citep{hou2024egosocialarena}, and SocialEval~\citep{zhou2025socialeval} examine social reasoning and theory of mind. Long-context suites include LongBench~\citep{bai2024LongBench}, LongBench v2~\citep{bai2024LongBenchv2}, $\infty$Bench~\citep{zhang2024InfinityBench}, HELMET~\citep{yen2025HELMET}, RULER~\citep{hsieh2024RULER}, and NoLiMa~\citep{modarressi2025NoLiMa}. Multimodal evaluation is represented by MMMU-Pro~\citep{yue2024MMMUPro}, CharXiv~\citep{wang2024CharXiv}, and Video-MME~\citep{fu2024VideoMME}. These benchmarks commonly combine objective items with LLM-based rubric scores and selective human review~\citep{zheng2023MTBench,liu2023GEval}. Their scale is valuable, but judge bias can cause stylistic fluency to be mistaken for substantive quality.

\paragraph{Safety and adversarial robustness.}
HarmBench standardizes automated red teaming and measures both harmful compliance and excessive refusal~\citep{mazeika2024HarmBench}. JailbreakBench supplies reproducible attack, defense, and threat-model protocols~\citep{chao2024JailbreakBench}. Fixed external safety suites remain important when the training curriculum adapts because self-generated experience can underrepresent rare harms or amplify unsafe behavior.

Capability benchmarks establish whether performance improved under a specified evaluator. They cannot by themselves determine whether the feedback used during training was faithful or whether the experience distribution remained informative. A method may raise GSM8K accuracy through a poorly calibrated intrinsic reward, improve HumanEval while generating progressively narrower tasks, or gain arena wins by exploiting judge preferences~\citep{dubois2024length,liu2024rm}. The next two subsections examine these missing forms of evidence directly.

\subsection{Feedback Fidelity}
\label{subsec:eval_reward}

Feedback evaluation tests whether the learning signal rewards the behavior it is intended to promote. This requires examining the reward mechanism itself, not only the capability of the resulting policy. The relevant object may be an outcome reward model, a process reward model, an LLM judge, or a model-derived signal such as certainty or consensus.

\subsubsection{Outcome reward models}

RewardBench evaluates outcome reward models through pairwise accuracy on preference triples spanning chat, reasoning, and safety~\citep{lambert2024rewardbench}. RewardBench~2 adds unseen human prompts, best-of-four selection, and tests of downstream utility~\citep{malik2025rewardbench}. Long-form RewardBench extends evaluation to responses that require document-level assessment and tests sensitivity to response length and error position~\citep{huang2026longformrewardbench}. Personalized RewardBench examines whether a reward model follows user-specific rubrics when responses have similar overall quality~\citep{ma2026personalizedrewardbench}.

Several benchmarks target specific sources of error. RM-Bench rewrites semantically equivalent responses in different styles to expose spurious preferences for length and formatting~\citep{liu2024rm}. PPE tests whether reward-model proxy metrics predict downstream RLHF outcomes~\citep{frick2024PPE}. RewardMATH studies robustness and overoptimization in mathematical reward models~\citep{kim2024RewardMATH}, and VL-RewardBench extends reward evaluation to vision-language responses~\citep{li2024VLRewardBench}. HelpSteer2~\citep{wang2024HelpSteer2} and RewardAnything~\citep{yu2025rewardanything} support multi-attribute or principle-conditioned evaluation. Across these settings, the central question is whether $\mathcal{R}_\phi$ represents the intended criterion or a convenient correlate of it.

\subsubsection{Process reward models}

Process reward models require evaluation at the level where credit is assigned. ProcessBench~\citep{zheng2024ProcessBench} and PRMBench~\citep{song2025PRMBench} ask verifiers to identify erroneous steps in mathematical trajectories. Socratic-PRMBench introduces fine-grained errors and summarizes their detection with PRM-Score~\citep{li2025socratic}. VILBENCH extends process-reward evaluation to vision-language reasoning~\citep{tu2025vilbench}. Final-answer discrimination and process-error detection measure different properties, so outcome accuracy must be accompanied by step-level F1 or PRM-Score.

\subsubsection{LLM judges and verifier utility}

Generative judges can be tested for agreement with people, factual discrimination, and utility within an inference or critique pipeline. G-Eval~\citep{liu2023GEval}, PandaLM~\citep{wang2023PandaLM}, JudgeLM~\citep{zhu2025judgelm}, Auto-J~\citep{li2023AutoJ}, Prometheus~2~\citep{kim2024prometheus}, FLAMe~\citep{vu2024FLAMe}, and BiGGen Bench~\citep{kim2024BigGenBench} cover rubric scoring, pairwise comparison, and fine-grained skill assessment. JudgeBench uses correct and incorrect responses drawn from verifiable sources to reduce reliance on stylistic cues~\citep{tan2024judgebench}. CodeJudgeBench applies the same principle to execution-checked code~\citep{jiang2025codejudgebench}, while JETTS evaluates judge utility in reranking, beam search, and critique refinement~\citep{zhou2025evaluating}.

Other suites probe judge reliability across modalities and contexts. MLLM-as-a-Judge covers multimodal inputs~\citep{chen2024mllm}. VerifyBench compares specialized verifiers with general LLM judges on expert responses in science and mathematics~\citep{li2026verifybench}. LongJudgeBench evaluates long-form outputs under different judging protocols~\citep{chen2026longjudgebench}, and JudgeArena supports controlled variation of the benchmark, judge, prompt, and inference backend~\citep{lushtaku2026judgearena}. LLMBar distinguishes instruction adherence from superficial preference~\citep{zeng2024LLMBar}, while CriticBench evaluates generation, critique, and correction as a pipeline~\citep{lin2024CriticBench}. Position-bias tests quantify order effects through repeated and swapped comparisons~\citep{shi2024judging}.

When an LLM judge supplies the training reward, reusing the same judge family for final evaluation creates a circular test. A disjoint judge, a held-out human subset, or a verifiable proxy provides a more credible measure of improvement.

\subsubsection{Intrinsic certainty and consensus signals}

Rewards derived from uncertainty, self-consistency, or likelihood relative to a reference require diagnostics that differ from preference accuracy. Confidence can be calibrated against correctness on verifiable tasks. Consensus rewards can be tested across sample sizes, temperatures, and distribution shifts. Reference-based signals require sensitivity tests for changes in reference wording. In every case, an external checker can reveal whether optimization improves reasoning or merely amplifies patterns already favored by the model.

The feedback audit must match the training mechanism. Reward-model policies require evaluation of the frozen $\mathcal{R}_\phi$ through RewardBench-style tests or downstream utility measures~\citep{lambert2024rewardbench,frick2024PPE}. Judge-trained policies require direct evaluation of $\pi_v$ and a secondary score from an independent evaluator~\citep{tan2024judgebench,jiang2025codejudgebench}. Process-supervised policies require both step-error detection and final-outcome measures~\citep{zheng2024ProcessBench,song2025PRMBench}. Policies trained with intrinsic rewards require calibration and external-checker audits. Matching the audit to the reward source separates policy improvement from the reliability of the signal that produced it.

\subsection{Experience Quality}
\label{subsec:eval_experience}

Experience evaluation asks whether the tasks and environments supplied to learning remain valid, diverse, and informative. This issue arises when the task distribution $\Pi$ or environment dynamics $(\mathcal{S},\mathcal{P})$ are generated or adapted during training. A rising internal reward can coexist with a narrowing curriculum, an inaccurate environment, or a verifier that co-adapts to the policy. Evaluation must consequently examine the experience stream as well as the trained policy.

\subsubsection{Agentic and interactive environments}

Web and computer-use benchmarks provide externally specified success criteria for long-horizon agents. WebArena~\citep{zhou2024WebArena}, VisualWebArena~\citep{koh2024VisualWebArena}, WorkArena~\citep{drouin2024WorkArena}, WebVoyager~\citep{he2024WebVoyager}, AssistantBench~\citep{yoran2024AssistantBench}, BrowseComp~\citep{wei2025BrowseComp}, and BrowserGym~\citep{chezelles2025BrowserGym} cover navigation, enterprise knowledge work, multimodal browsing, and information seeking. OSWorld~\citep{xie2024OSWorld}, Windows Agent Arena~\citep{bonatti2024WindowsAgentArena}, AndroidWorld~\citep{rawles2024AndroidWorld}, and AppWorld~\citep{trivedi2024AppWorld} extend evaluation to desktop, mobile, and multi-application workflows.

Tool-use suites include ToolLLM and ToolBench~\citep{qin2024ToolLLM}, StableToolBench~\citep{guo2024StableToolBench}, and the Berkeley Function Calling Leaderboard~\citep{patil2025BFCL}. AgentBench~\citep{liu2023AgentBench}, GAIA~\citep{mialon2024GAIA}, $\tau$-bench~\citep{yao2024Taubench}, TheAgentCompany~\citep{xu2025AgentCompany}, ScienceAgentBench~\citep{chen2024ScienceAgentBench}, and BALROG~\citep{paglieri2024BALROG} broaden coverage across general, scientific, workplace, and game settings. AgentDojo adds security evaluation through prompt-injection attacks and defenses in dynamic tool environments~\citep{debenedetti2024AgentDojo}.

The same suites can evaluate both policy capability and environment quality. A unified framework can vary agent scaffolds and environmental volatility to separate policy performance from execution conditions~\citep{zhu2026unifiedagent}. Trainee-Bench tests scheduling, information gathering, and reuse of experience across workplace episodes~\citep{fu2026traineebench}. AgentWorldBench evaluates whether a language world model reproduces state transitions observed in established agent benchmarks~\citep{zuo2026qwenagentworld}. For synthesized environments, relevant measures include validation pass rate, transition reproducibility, success-criterion correctness, solve rate under a frozen policy, and transfer to a public agent suite. These measures apply directly to the environment-generation methods discussed in Sec.~\ref{subsec:env_construction}.

\subsubsection{Task quality under self-generated curricula}

When the task distribution is generated during learning, solver accuracy must be accompanied by evidence about the proposed curriculum. Let $\bar{p}_{\theta_S}(\boldsymbol{q})$ denote the empirical solve rate of query $\boldsymbol{q}$. The learnability score $\bar{p}(1-\bar{p})$ in Eq.~\ref{eq:learnability} is largest at intermediate solve rates and vanishes when every sampled solution succeeds or fails. It identifies tasks near the current solver's learning frontier, although it does not measure novelty or semantic diversity.

Complementary statistics include the solve-rate distribution, embedding or $n$-gram diversity, novelty relative to the seed set, and rejection rates for invalid, toxic, or near-duplicate proposals. Corpus-grounded proposers enable document-level attribution as an external novelty check~\citep{liu2025spice,kwan2026scope}. Repeated concentration on a small source subset signals curriculum recycling. Trace-bootstrapping and instruction-evolution pipelines also need retention rates and filtering criteria because these choices shape the final task distribution~\citep{zelikman2022star,wang2022selfinstruct,xu2023wizardlm,fang2025serl}.

Capability leaderboards often conceal failures in the experience stream. Trivial or impossible tasks produce weak learning signals under group-relative optimization. A generator may repeatedly sample skills the policy has already mastered~\citep{fang2025serl,lsp}. In interactive settings, hallucinated rewards, flaky tests, and underspecified tools can corrupt every trajectory collected in a generated environment~\citep{yang2025zerogui,shi2026evoenv}. Curriculum and environment metrics are needed to distinguish genuine expansion of experience from these forms of collapse.

\subsubsection{Closed-loop evaluation}

The transition from experience beyond human design to autonomous co-evolution requires the strongest evaluation protocol. When policies, rewards, tasks, or environments adapt together, internal progress can diverge from externally measured capability. Relevant risks include distributional narrowing, non-stationary opponents or arenas~\citep{liu2025spiral,guertler2025textarena,dong2026agentworld}, and asymmetry between generation and verification~\citep{chen2026selfverify}. A credible closed-loop evaluation needs frozen external tests, trajectories of curriculum diversity, agreement between training and held-out verifiers, and ablations that freeze individual components of the loop.

PAST-Bench compares matched task sequences with and without retained experience to determine whether later gains follow the intended save, retrieve, and update pathway~\citep{xue2026pastbench}. AI4AI-Bench tests whether agents can improve learning algorithms inside frozen research repositories~\citep{chi2026ai4ai}. The modified procedures are rerun from scratch and scored by a fixed evaluator. This protocol tests a concrete component of recursive improvement without treating success on that component as evidence of a fully autonomous loop.

Public agent suites and synthetic environments play complementary roles. Success in generated worlds can reflect generator-specific regularities if it does not transfer to environments such as WebArena or OSWorld~\citep{zhou2024WebArena,xie2024OSWorld}. Public-suite gains are also incomplete evidence when the validity of the generated training environments is unknown. Credible evaluation combines learning curves in synthetic environments with transfer to frozen public environments under the same action interface.

\subsection{Datasets and Benchmark Suites}
\label{subsec:eval_datasets}

The role of a dataset depends on how it enters the learning and evaluation pipeline. A verifiable collection can serve as training experience, a held-out capability test, or a source for constructing an evaluator. These roles must be stated explicitly because reuse across them can invalidate the apparent independence of evaluation.

\paragraph{Verifiable training and evaluation sets.}
GSM8K, MATH, HumanEval, MBPP, and competition-code corpora are attractive for training because their checkers are inexpensive and objective. Their broad use also creates saturation and contamination risks~\citep{singh2024ConTAM,xu2024ContaminationSurvey}. LiveBench~\citep{white2025LiveBench}, LiveCodeBench~\citep{jainlivecodebench}, LiveMathBench~\citep{liu2024LiveMathBench}, DyVal~\citep{zhu2023DyVal}, AntiLeakBench~\citep{wu2024AntiLeakBench}, VarBench~\citep{qian2024VarBench}, and MMLU-CF~\citep{zhao2024MMLUCF} reduce particular leakage risks through fresh data, controlled perturbations, or closed-test construction. They do not eliminate protocol bias, so static and refreshed suites remain useful as complementary views.

\paragraph{Preference and evaluator-development sets.}
Outcome and process reward models are developed from preference comparisons, step labels, and rubric annotations. RewardBench-style data~\citep{lambert2024rewardbench,malik2025rewardbench}, HelpSteer2~\citep{wang2024HelpSteer2}, ProcessBench~\citep{zheng2024ProcessBench}, PRMBench~\citep{song2025PRMBench}, and Socratic-PRMBench~\citep{li2025socratic} instantiate different standards of judgment. An evaluator trained on one of these distributions requires a disjoint test. Otherwise, the benchmark measures familiarity with its annotation scheme as much as general evaluator fidelity.

\paragraph{Interaction logs and environments.}
WebArena-style environments~\citep{zhou2024WebArena,xie2024OSWorld,chezelles2025BrowserGym} and tool-use corpora~\citep{qin2024ToolLLM,patil2025BFCL} supply trajectories for long-horizon evaluation. Experience-generating methods additionally need a frozen split of synthesized environments, together with generator prompts and validation scripts~\citep{mei2026searchart,zhu2026unifiedagent}. These artifacts make environment-fidelity claims auditable.

Table~\ref{tab:eval_by_level} maps these dataset roles to the ladder. The essential distinction is not whether a dataset was originally produced by a person. It is whether the same data, evaluator, or environment participates in both learning and final assessment. As more components enter the learning loop, frozen external suites and experience-quality measures become increasingly important.

\paragraph{Sampling, decoding, and comparability.}
Results on the same dataset can vary with temperature, sample count $G$, token budget, tool access, and requirements on the reasoning trace. Reports need a complete decoding profile that includes temperature, top-$p$, maximum tokens, stopping rules, and random seeds. When sampling is part of the method, greedy and sampled results answer different questions and require separate reporting. Agent evaluations also require the action budget, observation format, available tools, reset procedure, and environment version~\citep{zhou2024WebArena,xie2024OSWorld,chezelles2025BrowserGym}. Without these controls, protocol differences can dominate the effect attributed to a learning signal.

\paragraph{Cross-domain transfer.}
Transfer provides a useful stress test for methods trained on a narrow verifiable domain. A complete report includes performance in the training domain, performance in the target domain, and the gap between them. Large in-domain gains with little transfer are compatible with several explanations, including overfitting, reward exploitation, and curriculum narrowing. Improvements across science QA, tool use, or open-ended interaction provide stronger evidence of generalization, although additional analysis is still needed to identify the cause~\citep{wei2024SimpleQA,wang2023SciBench,mialon2024GAIA}.

\subsection{Metrics, Protocols, and Reproducibility}
\label{subsec:eval_metrics}

\paragraph{Outcome and stability metrics.}
Exact match, numerical equivalence, and unit-test Pass@$k$ remain the primary metrics for verifiable domains. Let $G$ be the number of samples for each query and $c$ the number of correct samples. The unbiased Pass@$k$ estimator is
\begin{equation}
\mathrm{Pass@}k = \mathbb{E}_{q}\!\left[1-\frac{\binom{G-c}{k}}{\binom{G}{k}}\right].
\end{equation}
Mean sample accuracy is $c/G$, sometimes denoted Avg@$G$. G-Pass@$k$ measures consistent correctness across samples~\citep{liu2024LiveMathBench},
\begin{equation}
\mathrm{G\text{-}Pass@}k = \mathbb{E}_{q}\!\left[\binom{c}{k}/\binom{G}{k}\right],
\end{equation}
while CoT-Pass@$k$ additionally requires a valid reasoning trace~\citep{wen2025reinforcement}. The combination distinguishes occasional success from reliable reasoning across repeated samples.

\paragraph{Preference and judge metrics.}
Pairwise accuracy measures agreement between an outcome reward model and preference labels. Given triples $(\boldsymbol{q},\boldsymbol{y}_c,\boldsymbol{y}_r)$,
\begin{equation}
\text{Accuracy} = \frac{1}{|\mathcal{D}|} \sum_{(\boldsymbol{q}, \boldsymbol{y}_c, \boldsymbol{y}_r) \in \mathcal{D}} \mathbb{I}\!\left[ \mathcal{R}_\phi(\boldsymbol{q}, \boldsymbol{y}_c) > \mathcal{R}_\phi(\boldsymbol{q}, \boldsymbol{y}_r) \right],
\label{eq:pairwise_accuracy}
\end{equation}
RewardBench~2 extends this protocol to best-of-four selection, while PPE assesses metrics through their relationship with downstream RLHF outcomes~\citep{malik2025rewardbench,frick2024PPE}. Open-ended evaluation commonly reports win rates, rubric scores, and structured G-Eval scores~\citep{liu2023GEval}. Process fidelity is measured with step-level F1 or PRM-Score~\citep{li2025socratic,song2025PRMBench}. Every judge-based result also needs a diagnostic for position, length, and style bias~\citep{shi2024judging,liu2024rm,dubois2024length}.

\paragraph{Experience-quality metrics.}
Self-generated task streams require solve-rate distributions, mean learnability $\mathbb{E}[\bar{p}(1-\bar{p})]$, diversity, novelty, and filtering statistics. Synthesized environments require executability, validation pass rate, transition fidelity, and transfer to public agent benchmarks. Closed loops require trajectories of both internal reward and external held-out capability. Persistent divergence between the two signals warns of co-adaptive reward exploitation or collapse, although it does not identify a single cause on its own.

\paragraph{Contamination, bias, and tooling.}
ConTAM~\citep{singh2024ConTAM}, ConStat~\citep{dekoninck2024ConStat}, and related analyses~\citep{xu2024ContaminationSurvey,fu2024ContaminationAssumptions} illustrate the vulnerability of static suites to leakage. Refreshed and procedurally varied benchmarks reduce this risk~\citep{white2025LiveBench,zhu2023DyVal,wu2024AntiLeakBench,qian2024VarBench,zhao2024MMLUCF}. LLM-based evaluators also require explicit audits for position and style bias~\citep{shi2024judging,liu2024rm}.

Community tools including lm-evaluation-harness,\footnote{\url{https://github.com/EleutherAI/lm-evaluation-harness}} LightEval,\footnote{\url{https://github.com/huggingface/lighteval}} ZeroEval,\footnote{\url{https://github.com/WildEval/ZeroEval}} Math-evaluation-harness,\footnote{\url{https://github.com/ZubinGou/math-evaluation-harness}} and BigCode-evaluation-harness\footnote{\url{https://github.com/bigcode-project/bigcode-evaluation-harness}} standardize prompting, decoding, and aggregation. Reproducible reports must identify decoding seeds, judge prompts, environment settings, and exact harness revisions because these choices can materially change the comparison.

\paragraph{Recommended minimum scorecard.}
Our paper recommends frozen capability tests in the training domain and at least one transfer setting, together with an audit matched to the feedback source. Methods that generate experience also need direct measurements of task or environment quality. Claims based on heavily reused benchmarks require a contamination analysis, and all comparisons need compute-matched baselines under a shared protocol. Methods approaching L4 additionally require a longitudinal comparison between internal reward and frozen external evaluation. Table~\ref{tab:eval_by_level} presents the same requirements by ladder level.

\subsection{Limitations and Open Evaluation Gaps}
\label{subsec:eval_gaps}

Current evaluation practice remains much stronger for policy capability than for feedback fidelity or experience quality. Public leaderboards rarely test whether a training judge agrees with an independent evaluator. Metrics for curriculum diversity, environment fidelity, and closed-loop stability are also fragmented across individual methods. These weaknesses make it possible for internal reward to rise while the evidence for general capability remains unchanged~\citep{qi2026generalization,chen2026selfverify}.

Several practical gaps compound this problem. Contamination controls and dynamic evaluation are unevenly adopted outside a small set of benchmarks~\citep{white2025LiveBench,zhu2023DyVal,singh2024ConTAM}. Intrinsic-reward methods and verifier-based RLVR are seldom compared under matched compute, data access, and decoding budgets. Agent benchmarks offer realistic interaction but remain expensive and partly non-reproducible because live websites and graphical interfaces change over time~\citep{zhou2024WebArena,chezelles2025BrowserGym,xie2024OSWorld}. Multimodal and long-context evaluation is also only weakly connected to work on learning with reduced supervision, despite the growing use of multimodal, tool-using agents~\citep{chen2024mllm,yen2025HELMET,hsieh2024RULER}.

The three-part evaluation framework provides a practical response to these gaps. Capability is measured on frozen verifiable, expert, open-ended, or agentic suites. Feedback is tested with reward-model, process-model, judge, or intrinsic-signal audits matched to the training method. Experience is assessed through curriculum quality, environment validity, and closed-loop stability. Refreshed suites can reduce contamination in capability tests~\citep{white2025LiveBench,zhu2023DyVal,jainlivecodebench}, while expert checks and frozen judge panels provide sparse external anchors for feedback and experience.

No single leaderboard can establish progress toward self-sustaining learning. Claims must identify which components were evaluated externally and which remained inside the training loop. Missing evidence does not invalidate a method, but it limits the conclusion that can be drawn from its reported gains.

\section{Challenges and Future Directions}
\label{sec:future}

The methods discussed in the preceding sections expand the sources of reward and experience available to large reasoning models. Reusable evaluators reduce the need for repeated human judgment, while self-generated tasks and environments allow training to continue beyond a fixed corpus. Yet the same mechanisms weaken some of the external constraints that make conventional training easy to audit. Proxy exploitation, curriculum collapse, generator-verifier collusion, and overlap between training and evaluation can all remain hidden behind rising internal reward. Accordingly, the central challenge is to sustain learning with less direct intervention while preserving evidence that remains independent of the learning loop.

\subsection{Stabilizing Rewards beyond Per-Instance Human Evaluation}
\label{subsec:future_reward}

Section~\ref{sec:reward} reveals a tradeoff between the scalability of a reward and the independence of its evidence. Entropy, self-certainty, consensus, and self-judgment can score large numbers of rollouts at low marginal cost. However, prolonged optimization can sharpen the policy distribution, reinforce majority errors, and decouple confidence from correctness. Execution, games, and formal verification provide stronger external grounding, but incomplete tests and exploitable rules still leave room for unintended behavior~\citep{zhang2025no,shafayat2025can,hefar,huang2025accuracy,zhao2025one,wang2026reward}. Progress along the reward axis consequently depends on preserving independent evidence without recreating dense per-instance supervision.

\paragraph{Independent anchors for intrinsic and consensus rewards.}
Majority voting and confidence maximization avoid external labels at reward time, but neither signal can detect an error shared by the policy's own samples~\citep{shafayat2025can,zhang2025no}. Sparse executable checks, frozen evaluators from an independent model family, corpus-grounded consistency tests, and periodic expert audits can provide complementary evidence. These anchors need to remain outside the optimization loop so that they detect divergence between internal reward and held-out capability. Their purpose is not to restore dense supervision. Instead, they establish a reference against which model-derived rewards can be calibrated and monitored.

\paragraph{Verifier robustness under optimization pressure.}
Learned and generative verifiers cover outputs that brittle string matching cannot assess~\citep{chen2025xverify,xu2025tinyv,zhang2025generative}. This flexibility creates a larger attack surface because false positives can become highly rewarding under policy optimization~\citep{zhao2025one,huang2025accuracy,helff2026gaming,zhu2026rubrichack}. Robustness therefore has to be tested after exposure to an optimizing policy, not only on a static benchmark. Adversarial evaluation, differential testing against stricter checkers, and asymmetric objectives can expose or discourage collusion between generators and verifiers~\citep{zha2026rl,wang2025co}. Ensembles provide an additional defense by reducing dependence on a single evaluator~\citep{s34new_verga2024juries,s34new_coste2023ensembles}. Process-level rewards offer evidence beyond final-answer matching, although their intermediate judgments require the same adversarial scrutiny~\citep{cui2025prime,she2025r}. An open problem is how to certify that evaluator fidelity survives policy optimization when success on RewardBench-style probes does not guarantee robustness to newly optimized outputs~\citep{lambert2024rewardbench,malik2025rewardbench}.

\paragraph{Semi-verifiable and open-ended tasks.}
Evidence for scalable reward remains concentrated in mathematics and code because these domains support inexpensive outcome checks. Scientific reasoning, long-form factuality, multimodal interaction, and preference-sensitive generation admit multiple valid responses and more context-dependent criteria. Extending RL to these settings requires structured evidence from rubrics, claim-level verification, retrieval, and calibrated judges~\citep{viswanathan2025checklists,gunjal2025rubrics,chen2025learning,fein2025litbench}. Such evaluators must recognize substantive variation without collapsing quality into style or verbosity. Hybrid designs provide a practical intermediate approach by sending verifiable subproblems to deterministic checkers and open-ended components to learned evaluators~\citep{seed2025seed1}. The unresolved question is which criteria can safely be delegated and which still require an independent external standard.

\subsection{Scaling Experience without Curriculum Collapse}
\label{subsec:future_experience}

As task and environment generation enter the learning loop, the main difficulty shifts from collecting more trajectories to preserving their educational value. A proposer can generate tasks that are uniformly easy, effectively impossible, or confined to patterns the solver already knows. A synthesized environment can also be internally consistent while failing to represent the external world. Difficulty miscalibration, mode collapse, non-stationarity, and co-adaptive reward exploitation already appear in proposer-solver systems and generated environments~\citep{huang2025r,zhao2025absolute,fang2025serl,qi2026generalization}.

\paragraph{Learnability and diversity over time.}
Frontier-targeting proposers and regret-based environment design seek tasks near the solver's current learning boundary~\citep{dennis2020paired,huang2025r,zhao2025absolute}. Early solve-rate estimates $\bar{p}$ are noisy, however, and a well-calibrated difficulty distribution can still lack semantic diversity. A useful curriculum therefore requires longitudinal measurements of both learnability and coverage. Corpus-grounded and interaction-grounded proposers can reduce template recycling by drawing on external content~\citep{liu2025spice,kwan2025opensir,lu2025ssp}. Archives of unsolved and out-of-distribution tasks provide a complementary mechanism because they preserve difficult experience for later policies instead of discarding it when the current solver fails.

\paragraph{Environment fidelity and executable grounding.}
Generated simulators, unit tests, and tool stacks inherit errors from the models that construct them~\citep{yang2025zerogui,sun2025seagent,zhou2025selfchallenging}. Environment generation must consequently be paired with validation. Staged executability checks can reject malformed components before training, while differential testing against public harnesses such as SWE-bench, OSWorld, and WebArena can reveal discrepancies in task outcomes~\citep{jimenez2024SWEBench,xie2024OSWorld,zhou2024WebArena}. Transfer from synthetic training environments to held-out real interactions provides a stronger test of fidelity than success inside the generated environment alone. Without these controls, a learning loop may optimize against artificial dynamics that do not support capability outside its own simulator.

\paragraph{Stable closed loops.}
CURE, Self-Challenging agents, SPIRAL, and the Darwin G\"odel Machine couple different combinations of generators, verifiers, opponents, and agent modification~\citep{wang2025co,zhou2025selfchallenging,liu2025spiral,zhang2025dgm}. None yet closes the entire learning loop because fixed runtimes, game rules, task sources, or evaluation objectives continue to supply external structure. More complete co-evolution will require optimization and diagnostics that account for changing opponents and evaluators. Role-conditioned baselines can reduce variance in asymmetric self-play~\citep{liu2025spiral}. Frozen capability tests can reveal divergence from internal reward, while explicit verifier training can address the assumption that evaluation improves automatically with generation~\citep{chen2026selfverify}. Stopping criteria are also needed when internal reward rises but external transfer declines. Existing closed-loop experiments show improvement over the initial policy followed by a plateau below oracle-supervised training~\citep{qi2026generalization}. Extending this limit without allowing the loop to validate its own errors remains a defining challenge for autonomous co-evolution.

\subsection{Efficiency of Experience Sampling and Token Utility}
\label{subsec:future_efficiency}

Reducing annotation cost can increase inference and interaction cost. Group-based RL samples several rollouts per query, consensus rewards require repeated generation, and self-generated experience adds proposer, verifier, or environment-construction passes. Long reasoning traces further consume computation even when their tokens do not affect the final reward. Under Eq.~\ref{eq:GRPO}, tasks with nearly constant group rewards can spend substantial compute while producing little learning signal.

Experience generation is consequently a resource-allocation problem in which every sampled trajectory must justify its computational cost. Samples can be concentrated near the learnability frontier, and low-information rollouts can be rejected before full generation. Shorter process feedback may preserve credit assignment without scoring every token, while replay buffers can reuse expensive interactions across proposer and solver updates. At inference time, uncertainty and estimated task difficulty can guide the allocation of test-time computation. These mechanisms connect efficiency to capability because the number and diversity of useful experiences available to a policy are limited by the compute budget. Credible reports pair capability gains with the amount of generation, verification, and environment interaction required to obtain them.

\subsection{Evaluation under Receding Supervision}
\label{subsec:future_eval}

Capability, feedback fidelity, and experience quality provide complementary evidence, as developed in Section~\ref{sec:evaluation}. Current leaderboards emphasize capability on static test sets, while reward robustness, curriculum diversity, environment fidelity, and closed-loop stability remain less standardized~\citep{white2025LiveBench,qi2026generalization,chen2026selfverify}. The reuse of related judges for training and evaluation creates an additional risk because the same bias can improve both the policy's reward and its reported score.

Table~\ref{tab:eval_by_level} summarizes the minimum evidence required as more components enter the learning loop. Reward-generating methods need audits matched to the source of feedback. Methods that generate tasks or environments additionally need measurements of diversity, solvability, validity, and transfer. Live and procedurally refreshed benchmarks reduce dependence on fixed evaluation distributions~\citep{white2025LiveBench,zhu2023DyVal,jainlivecodebench}, although freshness alone cannot rule out contamination or protocol bias. Shared evaluation harnesses are equally important because comparisons among intrinsic-reward, verifier-based, and self-play methods require matched compute, data access, and decoding budgets. Automated evaluation remains credible only when its external anchors are frozen and independent of training.

\subsection{Trust, Safety, and Value Preservation under Autonomy}
\label{subsec:future_safety}

Human values do not disappear when direct supervision becomes sparse. Their influence moves into initial data, constitutions, tools, environments, objectives, and evaluation criteria~\citep{bai2022constitutional,s34new_sun2023principle}. As these components begin to co-adapt, the feedback channels that support learning can also amplify misalignment. Self-reward may reinforce sycophancy or unsafe high-confidence responses. Generated curricula may omit rare but important safety cases, and exploitable judges may favor persuasive behavior over correct or safe behavior~\citep{zhou2026more,yuan2024self}.

Accordingly, safety evaluation must examine the same three objects as capability evaluation. Policy tests measure harmful behavior and over-refusal, feedback audits test evaluator reliability under adversarial optimization, and experience audits assess whether generated curricula cover risk-relevant states instead of selecting only convenient cases. Sparse human review and constitutional audits can serve as external vetoes, but only if they remain independent of the policy, reward model, and curriculum generator. The broader aim is not to eliminate human involvement. It is to reserve human judgment for setting objectives, detecting drift, and intervening when self-generated evidence becomes unreliable.

\subsection{Toward Broader Training Paradigms and Superintelligence}
\label{subsec:future_pretrain}

Most methods considered in this paper apply RL after pretraining. A longer-term question is whether sparse external anchors, generated curricula, and closed experience loops can also support pretraining or continual mid-training, where direct annotation cannot cover the volume of data and interaction~\citep{silver2025welcome}. Extending these mechanisms will require reward signals that remain stable under large-scale optimization, curricula that expand semantic coverage, and evaluation procedures that detect drift before it compounds. Iterative self-distillation, emergent curricula, and agent-environment co-evolution provide pieces of this process, but none yet demonstrates safe, open-ended capability growth.

The five-level ladder describes a possible transfer of responsibility for sustaining learning. Humans may continue to specify goals, constraints, and external anchors even when they cannot demonstrate optimal behavior or inspect every reasoning step. Progress toward more autonomous learning is meaningful only if execution, interaction, frozen audits, and value constraints remain sufficiently independent to distinguish genuine improvement from an increasingly persuasive proxy. Whether this independence can be preserved at scale will determine if learning beyond direct human supervision supports reliable progress toward superintelligence or amplifies ungrounded self-reinforcement.

\section{Conclusion and Limitations}
\label{sec:conclusion}

Our paper examines how large reasoning models can continue to learn when direct human supervision no longer scales with the volume or difficulty of their experience. We organize this problem along two connected axes. The reward axis concerns the evidence used to evaluate a rollout, while the experience axis concerns the tasks and environments from which learning proceeds. The five-level ladder connects these axes by tracking which components still require continued human provision.

Along the reward axis, reusable evaluators and model-derived, reference-anchored, or environment-grounded signals reduce the need for direct judgment. Their scalability, however, is inseparable from the independence and validity of the evidence they use. Along the experience axis, task generation and environment construction allow the training distribution to adapt with the policy, but introduce difficulty miscalibration, curriculum narrowing, environment errors, and generator-verifier collusion. Continued improvement consequently requires rewards that remain faithful and experience streams that remain informative as both become more adaptive.

Together, these findings motivate a three-part evaluation framework that separates policy outcomes from the validity of the learning loop. Policy capability measures performance on frozen external tasks. Feedback fidelity tests whether the reward preserves its intended meaning under optimization. Experience quality assesses the diversity, validity, and transferability of generated tasks and environments. Capability gains alone are insufficient when feedback or experience is produced within the same learning loop. Scaling beyond human supervision also does not remove human intent. It relocates that intent into reusable evaluators, objectives, task seeds, environments, and sparse external audits. Such learning loops may contribute to progress toward superintelligence, but supervisory autonomy neither constitutes general intelligence nor guarantees reliable capability growth.

\paragraph{Limitations.}
The ladder is an analytical framework, not a set of mutually exclusive method classes. Methods often occupy transitional positions because reward acquisition, task generation, environment construction, and policy optimization do not become autonomous simultaneously. The framework also tracks the operational source of supervision during learning, not the historical amount of human knowledge embedded in pretrained models, references, corpora, tools, or environments.

Direct comparison across methods remains constrained by differences in models, compute, data, decoding, verifiers, and evaluation protocols. Our coverage of safety, multilingual reasoning, multimodal interaction, and long-context learning is necessarily selective, and the discussion of superintelligence remains conceptual. The framework does not imply that current systems are superintelligent or that movement along the ladder will inevitably produce superintelligent behavior.

Despite these limitations, the dual-axis framework clarifies the evidence required for credible progress beyond direct human supervision. Reliable capability growth depends on rewards that withstand optimization pressure, experience that continues to support learning, and external anchors that preserve contact with intended objectives.

\bibliography{ref}
\bibliographystyle{icml2026}  

\end{document}